\documentclass[PDFLaTex]{article}
\usepackage[english]{babel}
\usepackage{graphicx}
\usepackage{color}
\usepackage{hyperref}
\usepackage{todonotes}

\begin{document}

\title{Some hypotheses on how chatbots work in problem-solution-driven conversations:\\Large Language Models as confirmation of the Innovation Illusion}
\author{S.F.M.\ van Vlijmen and H.D.\ Lethe jr.}

\maketitle

\begin{abstract}
\noindent We discuss the nature of chatbots as conversation partners in problem-solving conversations. What can chatbots do and what can't they do? Our analysis draws on insights from Aggregation Dynamics, Cognitive Linguistics, Neuropsychology and Psychology.

We establish that chatbots are multifaceted and composite systems. Our argument focuses on basic chatbots in the hope of thereby making statements about the core functionality of more advanced chatbots. Basic chatbots are assumed to consist of a Large Language Model (LLM) with a simple interface.

The main results of our analysis are: a description of human imagination, understanding and thinking based on so-called metaphorical problem propagations; that the texts in text datasets used for training LLMs have specific characteristics and that these texts only partially imitate human thinking and understanding; that the LLM training process encodes artificial metaphorical problem propagations into an LLM from these text datasets.

Our conclusions are that a basic chatbot cannot be a thinking partner capable of matching the cognitive flexibility of humans, and that further development of LLMs will not lead to this either.

But chatbots exist, they are being used on a massive scale, by both individuals and organisations. It is therefore socially and politically important to understand them. Our article aims to contribute to the discussion on the functioning, benefits and drawbacks of chatbots. 

Cognitive Linguistics shows how the use of metaphor is an expression of our thinking. Aggregation Dynamics, is an attempt at a comprehensive systems theory. We believe that the concept of metaphorical problem propagation could provide an interesting addition for both.

Chatbots a solution? For what?
\end{abstract}

\section{Introduction}
Chatbots are winning people over! Why, and how? To many users, chatbots come across as fully-fledged, useful and reliable conversation partners. That is undoubtedly an exceptional achievement on the part of their creators. But the success of chatbots also raises many questions.

The seemingly creative and intelligent behaviour of chatbots is, for the time being, difficult to define, verify and explain. What are their characteristics, what exactly can chatbots do, and what can’t they do? They are certainly robust. Yet they are also unreliable in various ways. What tasks can be safely entrusted to them, and what absolutely cannot? These issues have attracted the attention of scientists, legislators and philosophers \cite{ebac00943-002404.14082v3, ebac00969-2407.15017v4, AI-snake-oil, ebac01025-NIST-FourPrinciplesExplainableArtificialIntelligence, ebac01044-TheNooscopeManifested, AI-ACT-EU, daEmpoli-Het-uur-vd-wolven,Brockman-WhatToThinkAboutMachenesThatThink}. 

In our article, we aim to provide an insight into how Large Language Model-based chatbots work and into the degree of freedom in problem-driven conversations with these chatbots, by combining insights from four different fields of research. The four fields are: Aggregation Dynamics, Cognitive Linguistics, Neuropsychology and Psychology. We emphasise it's about problem-driven conversations because, we suspect, it is in this type of human-chatbot conversation that what a chatbot can and cannot do becomes most clearly apparent.\\

\noindent \textbf{Aggregation Dynamics (AD): Innovation Illusion}\\
AD is an extension of ecology. Ecology studies the distribution and concentration of organisms based on their characteristics and interactions, both amongst themselves and with their environment \cite{ecology-begon-townsend5e-EN}. AD regards organisms and the environment as impractical abstractions when seeking to understand systems comprising people, products, organisations and cultures. The AD strives for an ecology without exceptions; it strives for totality \cite{transmathematica-about-EN-21-mei-2026}. The AD believes that this requires a much broader range of interacting players. AD refers to this broader category of players as aggregations. An aggregation is any \emph{thing} that is experienced by a player as a player. In the AD, for example, the following are all treated as equal players: projects, traffic jams, a combination of a cake plate with crumbs from a biscuit and a used coffee cup from a different set, bacteria in their agar-agar dish, the history of political dossiers, the jealousy of the gods and, of course, chatbots. To understand what concerns people and what they do, the AD looks, amongst other things, at what people perceive as opportunities or obstacles, what they identify as problems, and what solutions they devise and implement. The AD argues that people are less innovative in identifying problems and devising solutions than they themselves believe. The AD refers to this as the Innovation Illusion.\\

\noindent \textbf{Cognitive Linguistics: Metaphor Paradigm}\\
\noindent Cognitive Linguistics is fascinated by the relationship between language and thought. Metaphors are said to play a key role. Analyses in Cognitive Linguistics show that metaphors are much more than just exceptional poetic devices. In fact, our thoughts and our texts are teeming with them. Cognitive Linguistics argues that metaphors are the driving force behind human thought.\\ 

\noindent \textbf{Neuropsychology: Predictive Processing Theory}\\
\noindent A long-accepted theory of perception, thought and action posits that the eyes, and other familiar senses, form the basis of a process that generates a mental and detailed picture of the world. With that picture, and after reflecting on it, one can act. In Predictive Processing Theory, by contrast, the brain already forms hypotheses about the situation in which the thinking body finds itself. Perception, via many more sensory sources than just the familiar senses, serves to subsequently test or adjust those hypotheses. In this theory, the distinction between thinking and acting is fluid. Acting is a form of thinking that serves to refine perception, for example by looking more closely, or to intervene in the world and thereby bring it into line with the hypotheses about it.\\

\noindent \textbf{Psychology: Thinking Styles}\\
Psychology provides us with definitions of thinking, methods of thinking, and the classification of human thinking into a fast-satisfaction thinking style (System~1) and a systematic analytical thinking style (System~2).\\

\noindent \textbf{An outline of the argument}\\
\noindent The British cognitive scientist Margaret Boden described the field of Artificial Intelligence (AI) as follows: ``[it] seeks to make computers do the sorts of things minds can do'' \cite{MargaretBoden-AI-2018}. This brief characterisation serves as the starting point for our argument. If we are to say anything about chatbots, we must first have some understanding of those \emph{minds}. The speculative argument consists of four parts, which we summarise briefly below.

\textbf{1.} We formulate a model of human beings' more or less conscious mental spaces concerning the current situations in which they find themselves, their memories and their plans. These mental spaces are associative and flexible, and the mind can ‘wander’ through them, tinker with them, zoom in and out, move from the concrete to the abstract and back again, and these mental spaces can also be expanded or restricted at will. We call such mental spaces: \emph{metaphorical problem propagations}. Our model arises from the integration of the conceptual system of Cognitive Linguistics (concept, experience) with the problem and solution dynamics of Aggregation Dynamics (motivation, direction, interaction) and the prediction hypothesis of Predictive Processing Theory (perception, thinking, action).

\textbf{2.} Large Language Models are primarily based on written texts available on the internet. These texts are the result of a process of reduction. This is because texts, whether spoken or written, are always the result of a process of selection and combination. Furthermore, written texts lack context and physicality. We therefore argue that most written texts from datasets only partially correlate with human mental capabilities.

\textbf{3.} We propose that a substantial proportion of the texts on which LLMs are trained concern the relationship between problems and solutions. Texts have always been reduced material, and these texts also have a limited repertoire of structural patterns related to the relationship between problems and solutions.

Our hypothesis here is that the training process in a Large Language Model constructs fragments of artificial metaphorical problem propagations. After training, in a conversation, these propagations are activated to a greater or lesser extent and thus, as it were, act as riverbeds, exerting a pull on the course of word generation.

A problem-motivated conversation can now be defined as a path through metaphorical problem propagations, both on the chatbot’s side and on the user’s side. But these propagations differ from one another. The final step of the argument concerns these differences.

\textbf{4.} We argue that a basic chatbot cannot simulate analytical thinking, because it `tends' to seek out the `riverbeds' of the artificial metaphorical problem propagations, and these represent a reduced form of thinking. Since chatbots do not possess a body and have no experience, they are unable to assess and correct their own propagations in terms of effectiveness. That does not mean that a chatbot is incapable of providing certain forms of cognitive support, but this must be guided by an alert and creative prompt writer.\\

\noindent \textbf{Structure of the article}\\
\noindent Sections \ref{AIbots} through to \ref{denkstijlen} provide introductions tailored to this article on: AI, chatbots and neural networks; the insights drawn from Aggregation Dynamics, Cognitive Linguistics, Neuropsychology and Psychology. Anyone familiar with this material may skip it and start in Section~\ref{kernbetoog}. 

The main argument, in the four steps described earlier, follows in Section~\ref{kernbetoog} through to Section~\ref{watvoorp}. In the conclusion, Section \ref{conclusie}, we summarise everything.

\section{AI, chatbots and LLM modelling}\label{AIbots}
Artificial intelligence (AI) is a multidisciplinary field. And, relative to the current hype, it is also an old field. The development of AI went hand in hand with the development of automatic calculators (what we now call computers) from the 1930s onwards \cite{Mitchell-A-guide-AI-2019,AlbertsVanVlijmen-Computerpioniers2016-EN}. The question of what computers are capable of was already being asked at that time and has occupied generations of researchers and engineers. A prime example of this is the work of Alan Turing. For an overview, see \emph{Mind As Machine: A History Of Cognitive Science} by the British philosopher and cognitive scientist Margaret Boden \cite{epub00913-Margaret-Boden-Mind-as-Machine}. The term \emph{artificial intelligence} dates from 1956 \cite[p.~331]{epub00913-Margaret-Boden-Mind-as-Machine}; previously, the term ‘computer simulation’ was used. 
 
According to Boden, there are five main schools of thought in AI theory \cite{MargaretBoden-AI-2018}. These relate to the way in which information is represented and processed. One of them is inspired by data processing in living organisms: fault-tolerant, parallel and distributed. Artificial neural networks stem from this school of thought, also known as connectionism.

Artificial neural networks are modelled on the neural network structures in the brain and the way in which neurons within them receive and process chemical signals from one another, and then transmit the result to organs or other neurons. Artificial neural networks work in a similar way. An artificial neuron reads numerical signals from the input ports through which it is connected to other neurons, performs a calculation using these, and outputs the result via its output ports. 

How do the neurons in a neural network ‘know’ which calculation to perform? Initially, they know only a very small part of this. It requires a training process to teach them the details. In this process, each neuron and the network as a whole learn, step by step, the desired relationship between input and output. For example, that an image with a cat prominently in the centre corresponds to the word ‘cat’, and an image with a dog corresponds to the word ‘dog’.

To put it very simply, you could say that a neural network implements a lookup table. That sounds cumbersome, and it often is, for example, for a telephone directory. But neural networks also have major advantages. It is a one-size-fits-all computational model: in principle, you can teach the same neural network to function as a telephone directory, but also to answer questions as a helpdesk.
Furthermore, neural networks can withstand some damage. Imagine a large neural network distributed across 100 computers. If one of those computers fails, the network is not necessarily rendered unusable straight away. Moreover, neural networks are not fussy about their input. An example is offered by the well-known app Pl@ntNet. This app has been trained to identify a plant based on a photo of it. If you present it with a very blurry photo of a plant, the app will still try to say something meaningful about it. Both of these latter properties are referred to as robustness. 

However, neural networks also have disadvantages: they are unreliable. Pl@ntNet could, so to speak, look at a photo of a bicycle and claim with 100\% certainty that it is an oak tree. For further details, we refer to the extensive literature on this subject, for example \cite{Mitchell-A-guide-AI-2019}.

A Large Language Model is a specific type of neural network. These networks are not trained to link images to the words ‘cat’ or ‘dog’, but are trained on text and text that plausibly connects to it. For example, if you provide an LLM with the word ‘I’, there is a good chance it will return ‘am’. Because ‘I am’ is a very common sentence or the beginning of one. A Large Language Model forms the heart of a chatbot.

\subsection{Chatbots and basic chatbots}\label{chatbots}
By the term ‘chatbot’, we refer to systems accessible to a wide audience, such as ChatGPT, Claude, Grok, Lumo and so on. These are not designed to support very specific tasks; they are trained on bulk text from the internet.
 
Nevertheless, chatbots are difficult to categorise under a single heading. They are, to a large extent, multifaceted and complex or hybrid systems, which are also still very much under development. They are often combinations of different software components based on different approaches and different architectures. The most important part of a chatbot is the Large Language Model (LLM). Other components in a chatbot act as an interface, managing the interaction between the LLM and the user, providing additional functions and moderating the process. We could call that a software wrapper around the LLM. This ‘wrapper’ ensures that the LLM does not output text deemed undesirable, such as racist language or advice on suicide. Other components control the LLM when summarising text or adjusting the style of a text \cite{ebac01027-Basyal2023TextSU,ebac01028-info14060303,ebac01029-3419106, ebac01030-coli_a_00426}.

It is likely that LLMs will give rise to hybrid forms, because a natural language interface to all manner of tools is, of course, highly appealing. The American mathematician, author and software developer Stephen Wolfram describes the combination of the Wolfram|Alpha computational system and a chatbot \cite{Wolfram-ChatGPT-2023}. There are many more examples \cite{ebac01017_ICCC-2024-Proceedings,ebac01018_ICCC-2025-Proceedings}. Exactly how the current public chatbots are constructed is impossible to say for outsiders, who are far removed from the closed-off development process.

The hybrid nature of chatbots as assumed by us, has implications for our argument. For which functionality of which component should we be discussing? Let us define a basic chatbot as consisting of a Large Language Model with a simple interface. When we refer to chatbots and conversations hereafter, we mean chatbots limited to the functionality of these hypothetical basic chatbots. We assume that this abstraction is meaningful. That is to say, statements about basic chatbots are also relevant to their forms expanded with additional functionality. What we discuss in this article essentially concerns what an LLM is capable of and what that means for its current use and future potential.

The Large Language Model is the text-generating engine of a chatbot. Given any input text, the LLM is able to predict, on statistical grounds, what text might follow. The process of generating that text takes place in word fragments, known as tokens, or by words, or in even longer sequences. The LLMs of the aforementioned current chatbots generate text per token. We shall continue with the previous example, in which, for the sake of simplicity, we consider the generation process to be ‘per word’. If ‘I’ is the input, then ‘am’ will score highly in the output, but ‘have’ is also a good candidate. In contrast, ‘traffic sign' or ‘elephant' will score low. Because ‘I traffic sign' is not a typical sentence, except perhaps in absurdist theatre. 
So there is not just one possible continuation, but rather a long series of possibilities, each with its own probability. A choice is made from that series. 

Of course, it may also turn out that there is no plausible continuation, according to the chatbot. In that case, the text generation process stops or the shell surrounding the LLM intervenes. On what grounds? Stephen Wolfram states regarding the selection rules and configuration rules of an LLM \cite{Wolfram-ChatGPT-2023}: 
\begin{quote}Particularly over the past decade, there’ve been many advances in the art of training neural nets. And, yes, it is basically an art. Sometimes—especially in retrospect—one can see at least a glimmer of a ``scientific explanation'' for something that’s being done. But mostly things have been discovered by trial and error, adding ideas and tricks that have progressively built up a significant body of knowledge about how to work with neural networks.\end{quote} An LLM is a product whose functioning remains largely unexplained; see Section~\ref{interp}. Nevertheless, it performs very convincingly in conversations. But the question remains: how?

LLMs are trained on text fragments and their plausible textual continuations. That works perfectly well. Chatbots effortlessly produce fluent sentences. That was exactly what researchers and developers had hoped for, but they are capable of more.

To many people’s surprise, it turned out that LLMs are also capable of what appears to be creative reasoning. British researcher Geoffrey Hinton, a leading figure in AI, cites as an example the chatbot’s response to his question: ‘What is the similarity between a haystack and an atomic bomb?’ \cite{Hinton-interview-DOAC-Steven-Bartlett}. The chatbot replied that both are self-sustaining combustion processes. That is a striking analogy. A form of analogy-making ability had previously been observed in the semantic networks that encode words, which are components of LLMs \cite[p.~248]{Mitchell-A-guide-AI-2019}. Research into the capacity for analogy-finding is a topical subject within AI research \cite{ebac00978-lee2025curiouscaseanalogiesinvestigating,ebac01009-opielka2025analogicalreasoninginsidelarge,ebac01010-qin2025relevantrandomllmstruly}. 
We assume that LLMs possess some `sensitivity' in this area; this observation is sufficient for our argument.

Neural networks also have other desirable properties, such as robustness; and less desirable ones, such as unreliability. These properties play a role at the conclusion of the argument, namely in Section~\ref{watvoorp}. We will explain what we mean by unreliability there.

\subsection{\emph{Interpretability}: neural network models}\label{interp}
The question of the ‘how’ of neural networks is the subject of the research field \emph{Interpretability} \cite{ebac00969-2407.15017v4}. Within this field, researchers attempt to explain the emergent behaviour of neural networks in terms of the behaviour of individual neurons by creating models of neural networks and by design modifications.  Leonard Bereska and Efstratios Gavves describe the state of the art for LLMs in \emph{Mechanistic Interpretability for AI Safety -- A Review} \cite{ebac00943-002404.14082v3}. The authors note that developments in interpretability have parallels with those in Psychology.
 
Psychology began by observing behaviour and developing theories, as the techniques for measuring or simulating brain processes did not yet exist. Step by step, these techniques were developed within new disciplines such as Cognitive Science, Neuropsychology and Neurology. Research into AI interpretability exhibits a similar spectrum of approaches. Bereska en Gavves distinguish four of them, which they describe as follows (our italics): ``\emph{Behavioral} analyzes input-output relations; \emph{Attributional} quantifies individual input feature influences;\emph{ Concept-based} identifies high-level representations governing behavior; \emph{Mechanistic} uncovers precise causal mechanisms from inputs to outputs.''

Our argument focuses largely on concepts and the relationships between them. See Section~\ref{patinprop} and the discussion of the conceptual system in Cognitive Linguistics. 
To us, the results from Concept-based and Mechanistic interpretability seem to align most closely with our argument. By ‘measuring’ within the network, interpretability researchers attempt to isolate network fragments that correspond to specific concepts and activation patterns that can be traced back to behaviour of the artificial neural network \cite{ebac00967-s42256-025-01084-w}.

There are clear indications that concepts are not distributed randomly across the network, but can be localised as clusters. Furthermore, related concepts may overlap or lie in close proximity to one another. This does not mean that all concepts are stored in this way. However, we assume that overlap means that if, of related concepts A and B, A is activated, the likelihood increases that B will also be activated. The occurrence of this overlap is understandable: it saves storage space and thereby increases precision, which is precisely what the training of a neural network aims to achieve.

\section{Aggregation Dynamics}\label{ADInno} The philosopher Karl Popper titled a collection of essays \emph{All Life Is Problem Solving} \cite[p.~99]{Popper-All-life-problem-solving}. We see it that way too. The interactions between aggregations are about problems. Rather than tackling ‘life' -- that is, all the problems of all aggregations -- head-on, Aggregation Dynamics makes a start by investigating problems and solutions as experienced by people.
 
\subsection{Problem and problem position}\label{probpos}
Imagine you experience something that doesn’t align with your wishes, needs or ambitions. We call that a problem. But the problem only becomes concrete once you articulate or state it: you adopt a problem position. A problem position sets out in detail what we see as the relevant aspects when thinking about and/or communicating on the matter. Typical elements of a problem position are: pressing issues, classifications into main and subproblems, potential avenues for solutions, phasing, background knowledge, facts, rumours, opinions or views of other stakeholders, experiences with certain solutions, priorities, the current status of the implementation of certain subproblems, explanatory histories. Note that this involves weighing up potential solutions, but not the implementation of the solution step itself. A problem situation is a snapshot in time where one pauses for a moment and, for example, asks: ‘Where do we stand now?’
 
A problem position does not necessarily present a consistent picture. For example, one adviser might say ‘do it’, whilst another says ‘don’t do it’. In fact, the position is probably usually inconsistent, except in the case of extremely trivial problems. It is often precisely this lack of consistency that contributes to the sense that there is a problem.
 
The AD also claims that it makes sense to distinguish between three types of problem position: the natural, the internal and the external \cite{ebac00899-LosAngelesSmog-probleemposities}. The natural problem position is one that extends deep into the physical state of a person reflecting on a problem position. This is only partially accessible to consciousness. The internal one is the conscious mental representation. The external one is still a mental representation, but tailored to a recipient. There may be several of these: ‘if someone asks how the project is going, I say \emph{this} to A and \emph{that} to B.’

The AD claims that internal and external problem positions are manageable in terms of scope and complexity \cite{ebac00899-LosAngelesSmog-probleemposities}. This applies both to the mental representation and to a problem position expressed in words. It is assumed that a simple problem-position jargon, a kind of formulaic language, is sufficient for the description. A problem statement described in that jargon covers the essentials in a concise, matter-of-fact manner. Thus, problem statements concern a limited number of observations, assertions, facts, events from the past deemed relevant, potential next steps, actors and a limited number of relationships between all of these.
 
There are no rigid boundaries to this mental representation, nor are there any formal requirements. It is what an individual, entirely on a personal level, experiences as a whole and what they consider to be of significance. The words of the American philosopher John Dewey (1859-1952), as quoted by Mark Johnson, seem apt here. The problem position contains that which intuitively aligns with the ``pervasive unifying quality''  of the moment \cite[p.~52]{Johnson-embodied-2017}.

Finally, the AD assumes that, provided one works carefully, verbalised versions of problem positions can provide a picture of the complexity of a person’s internal problem position \cite{ebac00899-LosAngelesSmog-probleemposities}. That is to say, through careful interviewing, or by asking someone to work out an internal problem position honestly and fully, or through careful archival research, one can obtain a reasonably clear picture that correlates in terms of complexity with the complexity of an internal problem position. But the natural problem position is assumed to remain largely out of reach.

Although language is a tool for gaining an understanding of the complexity of an internal problem position, this does not mean that just any text already offers the necessary depth. In Section~\ref{metaforische-probprop-tekst}, we explore this further and will argue that much of the text produced by humans lacks the complexity of problem positions. Our claim is: most text represents a situation as stable and understood, and then directs the reader towards a goal that takes into account only a limited number of variables.

\subsection{Problem propagation}\label{probprog}
One could describe human behaviour by stating that people move from one problem state to another via solutions or steps in that direction. Leaving aside the question of whether the measures observed are well thought out or not, one could make the following observations. For trivial problems, a single solution step may be all that is needed. For more complex problems, a series of steps often needs to be taken. These are usually not simple linear series of steps. The approach to one problem may clash with the approach to another problem. Or, the approach may seem more complicated than first thought and necessitate splitting the problem into several subproblems. Problem positions are therefore linked through various relationships, through mutual dependencies and through solution steps.
 
Aggregation Dynamics refers to such a network of problem positions that merge into one another through solution steps as a ‘problem propagation’. Propagation should be understood as proliferation, expansion or growth: both the approach to a problem and its solution can lead to new problems. Simply because the world does not stand still.

When one maps out a problem propagation, one can look back (history) and look ahead (plans, scenarios). The ‘present’ in a network is referred to by AD as the problem front. At the front, problem positions can collide and choices are made. 

\begin{figure}
\centering
\includegraphics[height=9cm]{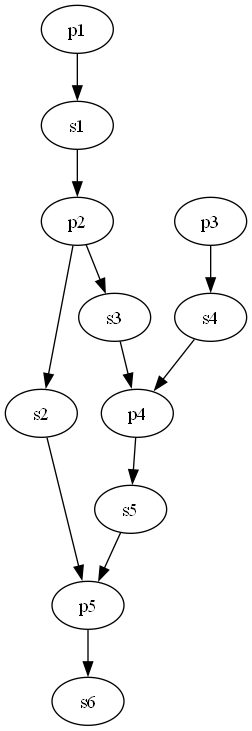}
\caption{\label{probleempropagatie}An example of abstract problem propagation. Here, the p’s represent problems and the s’s represent solutions.}
\end{figure}

Problem propagations can be represented graphically and/or in text. Figure~\ref{probleempropagatie} illustrates an abstract problem propagation, with the p’s as problems and the s’s as solutions. 
  
\begin{figure}
\centering
\includegraphics[width=11cm]{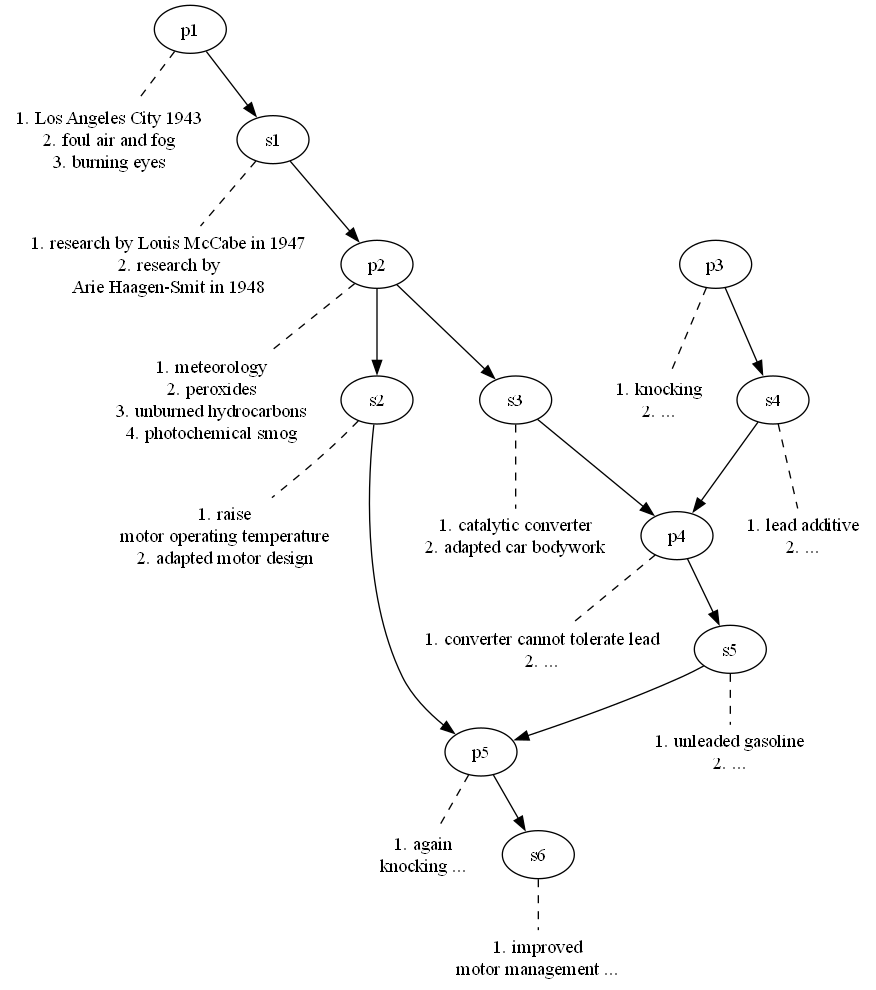}
\caption{\label{contextprobleempropagatie}An example of problem propagation and a suggestion of some context from the Los Angeles Smog case, as seen from the perspective of a fictional observer.}
\end{figure}

It doesn’t have to be so abstract. The representation of a problem propagation is not subject to any formal requirements \cite[p.~7]{ebac00898-LosAngelesSmog-probleemtypen}. A problem propagation can be elaborated to a greater or lesser extent with the details of a specific solution process. In Figure~\ref{contextprobleempropagatie}, the earlier propagation is supplemented with concrete information drawn from the Los Angeles Smog case \cite{ebac00899-LosAngelesSmog-probleemposities, Jacobs-Kelly-SmogTown2008, Dewey-Dont-breathe-the-air}. To help the reader understand the annotations in Figure~\ref{contextprobleempropagatie}, we will briefly explain what the Los Angeles Smog case is about.

Around 1943, air pollution in the industrial city of Los Angeles and the surrounding towns reached alarming levels. It became an issue that would dominate local political debate for decades, and which had a major influence on technical developments and environmental thinking both nationally and internationally. Initially, it was thought that the smog was caused by the use of coal for heating, power generation and industrial processes. Such smog was familiar from places such as St.\ Louis, Pittsburgh and London. But after years of research, the main cause in Los Angeles turned out to be of an entirely different nature: the incomplete combustion of petroleum products in car engines. In summer, the volatile hydrocarbons did not blow away but lingered over the cities due to meteorological conditions and the mountains surrounding Los Angeles County. The abundant sunshine now had free rein to cause the hydrocarbons to react into all sorts of nasty substances: photochemical smog. It took many years to bring this problem under control, for a great many reasons. For example, the car-addicted and vociferously complaining public refused to believe that they themselves were partly to blame. Reluctantly, the oil and automotive industries set about adapting their products to achieve better combustion. One example of this is the use of a catalytic converter in the exhaust (the precursor to our current three-way catalytic converter). However, this turned out to require unleaded petrol. But that lead was in petrol to prevent premature ignition in the cylinders, known as knocking. Without lead, knocking had to be prevented in some other way, and so on. The Los Angeles Smog case is rife with problems multiplying.

Just as with problem positions, problem propagations can also be described with varying degrees of accuracy or in varying degrees of detail. That said, according to the AD, it is never complete. Language always falls short. For example, because internal problem positions flow in an inimitable way into unattainable, deeper natural problem positions. That does not detract from the fact that problem propagations can be worked out in great detail by incorporating complete problem positions and unravelling the solution steps much further.
 
\subsection{The Innovation Illusion}
People seek stability, value predictability and like to work according to established patterns. Standardisation and working to procedures are the result of this. The Innovation Illusion argues that people regularly tend towards predictability when solving problems, whilst believing they are actually being innovative. This tendency towards predictability and repetition only becomes apparent when one looks more closely at the pattern of interaction and abstracts from details that have no causal significance, from the domain, the players, and from scales of time and space.

The Innovation Illusion posits problem types as a finite set of player, \mbox{scale-,} time- and domain-independent abstract undesirable situations and abstract solutions for them. The set of solutions is also assumed to be finite. The set is called the repertoire of the problem type; we sometimes also refer to solution types.

\pagebreak

An example is the problem type congestion. In the Netherlands, there is both a problem with traffic congestion on motorways and an overload on the national electricity grid.  Both problems are instances of the congestion problem type. The solutions to these problems show many parallels. Two examples: additional lanes in one domain correspond to thicker cables in the other; avoiding rush hour is comparable to ‘a request for limited electricity consumption by households between 9 am and 4 pm’. These are analogous situations and interventions. In terms of the type of problem, we are dealing with an aggregation (road/cable) with a function (facilitating transport) that, compared to other aggregations (road users/electricity users), is unable to fulfil its role sufficiently.
 
The Innovation Illusion is likely exaggerating, but where the line between repetition and innovation lies remains unclear for the time being. For our argument, it is important to assume that in most problem formulations and the subsequent solution processes, the tried and tested path is followed.

\subsection{Problem propagation and its patterns}\label{patinprop}
An important question is: are there patterns in problem propagations? The hypothesis, from a previous section, that descriptions of problem positions consist of a finite number of statements seems to suggest as much. The Innovation Illusion, explained above, significantly reinforces that suspicion.

The Innovation Illusion asserts that the number of problem types is finite and that the corresponding repertoires are also of finite size. Evidently, a problem propagation abstracted down to the problem-type level is then full of repetition and patterns. For, suppose there are $n$ problem types; then the next problem in a problem propagation that goes $n$ problem steps deep must be of a type that has already occurred. For example, the maintenance of a complex device leads to the deployment of a particular expert, and that expert must also maintain her own equipment.

It follows from the above that, when considering problem positions and problem propagations relating to concrete problems, one can also envisage more abstract variants. Step by step, one can work towards representations at the type level, or even more abstractly, types of types. And from there, one can of course also go back, namely by detailing more concrete cases. 

\section{Cognitive Linguistics}\label{taalcogni}
Here we outline the key concepts from Cognitive Linguistics that are relevant to us, as developed in particular by the linguist George Lakoff and the philosopher Mark Johnson.

Everyone is familiar with metaphors and analogies as figures of speech. Both refer to the way in which we understand and experience a thing, behaviour or situation in terms of another thing, behaviour or situation. Cognitive Linguistics favours the term ‘metaphor'. The metaphor perhaps has a more poetic connotation, in which the choice of words carries somewhat more weight than in the analogy. The latter is perhaps more descriptive, and the chosen words are instrumental; in other words, it can be put differently. Be that as it may, the metaphor or analogy sometimes comes in very handy for making something clear.

Sometimes? No, the opposite turns out to be true. Virtually every sentence contains a metaphor because we cannot understand or express ourselves without making comparisons. This insight gained widespread recognition with the publication of Lakoff and Johnson’s \emph{Metaphors We Live By} in 1979. This theory is of great importance to our argument. Texts are rich in layered meaning and give an insight into how we think. The field of research that Lakoff and Johnson helped shape is now referred to as Cognitive Linguistics \cite{Thewaywethink,Feldman-From-Molecule-Metaphor,ebac01035-s11097-025-10055-w}. We will discuss the central elements, which are the most important to us.

Cognitive Linguistics holds that human mental activity -- thinking, assigning meaning, and motivated action -- is profoundly shaped and constituted by bodily experiences, movements, sensory stimulation, emotions, feelings and actions. There is a constant interaction between the environment and the body-brain.

The concepts, their properties and relationships, with which we are so familiar in our daily lives, are not objective facts, but are learned step by step. This begins as early as infancy with our physical experiences, known as perceived correlations: if this happens, then that happens too. Direct physical experience leads to what are known as primary concepts or metaphors. Some examples are: \emph{Knowing Is Seeing}, \emph{More Is Up} and \emph{Speech Act is Physical Force}. We shall explain them.

\emph{Knowing} is understood in terms of \emph{Seeing}. As a child, you learn that looking more closely helps you understand better. You learn this even before you can put it into words. Once you have mastered language, you understand expressions such as: ‘I see what you mean’, ‘Could you elucidate this aspect further’ or ‘I don’t find the conclusion clear yet’. \emph{More Is Up}, for example, is something you learn when playing with water. You see that the level in your cup rises when you add more water. You learn the fundamental concept that \emph{Speech Act Is Physical Force} when your parents grab hold of your body and move it, for example your hand, whilst giving a verbal instruction: ‘Move your hand! It’s hot!’

The metaphor is the mental tool for giving meaning to new experiences and situations based on concepts, possibly primary concepts, that you already know. Anyone who conceives a new concept for the first time or learns it from another person will initially be aware of the metaphor, but as soon as the concept becomes more familiar, it takes on a life of its own and you are no longer aware of the metaphor \cite{ebac00966-NeuralTheoryofMetaphor}. Higher-order concepts, i.e. those above the primary level, are called conceptual metaphors. Metaphors elevated to the status of concepts. The more concepts one has at one’s disposal, the more situations can acquire meaning and the greater the scope for mental manoeuvre.

Concept formation and conceptual metaphors do not provide a complete explanation of abstract thinking, but are an important mechanism and key elements within it \cite{Johnson-embodied-2017,ebac01008-Holyoak02072024}. The metaphor \emph{Categories Are Containers} illustrates the relationship between primary and more abstract concepts. Hundreds of times a day we experience the concept \emph{container}; it is a recurring property of many things we deal with daily, from an early age: bags, bottles of apple juice, shoes, houses, trouser pockets. You can put something inside something. But sometimes it fits and sometimes it doesn’t. This is how we learn the spatial logic of cases within the category of containers. If I have my keys in my hand, and I have my hand in my trouser pocket, then my keys are in my trouser pocket. If one now elevates the container metaphorically to a category or concept, then abstract concepts can also contain one another, and this becomes set theory. If a blackbird is a bird and birds are animals, then a blackbird is also an animal \cite[p.~52-53]{Johnson-embodied-2017}. 

Cognitive Linguistics suggests that all of our concepts take the form of metaphors. Even the most abstract conceptualisations in the exact sciences acquire meaning through the engagement of very ordinary ``sensory-motor parts of the brain''. The meaning of concepts is therefore grounded in our ‘sensory-motor and affective experience’ \cite[p.~52]{Johnson-embodied-2017,Hofstadter-Sander-Surfaces-Essences}.

\section{Neuropsychology}\label{neuro-psych-en-psych}
The work of Lakoff and Johnson and others, dating from the late 1970s, predates by several decades a major leap forward in empirical research in the neurosciences. The results of this research confirm the picture outlined by Cognitive Linguistics and further flesh it out \cite{Feldman-From-Molecule-Metaphor}. To illustrate this, Johnson refers, amongst other things, to the work of the American psychologist D.M.\ Tucker, who stated: ``there are no brain parts for disembodied cognition'' \cite[p.~51]{Johnson-embodied-2017}. Here, we draw in particular on the work of the British researchers Andy Clark and Anil Seth and their perspective on Predictive Processing Theory \cite{AndyClark-ExperienceMachine,AnilSeth-BeingYou}.

Predictive Processing Theory posits that what we experience as perception is the brain’s best \emph{prediction} of the situation we find ourselves in. This theory assumes that our thought processes constantly generate hypotheses about the situations in which the body and brain find themselves, and then test these against sensory data. This includes data about the external world, but also about the body’s internal state and posture. So it concerns sensory data from familiar senses such as sight, hearing, smell, taste and touch, but also body temperature, heart rate and the orientation of body parts.
 
Integrating all this hypothesising yields a complete and coherent picture. At least, that is how we experience it. We think we are truly perceiving the world. But it is merely a projection. Following in the footsteps of others, Anil Seth refers to the perception as we consciously experience it as a \emph{controlled hallucination}.
 
A long-accepted, and intuitively understandable, theory of our perception, thought and action posits that the eyes, and other familiar senses, form the basis of a process that generates and updates our mentally experienced, clear picture of the world. Only then can thinking take place and action be taken. Predictive Processing Theory reverses the order of the process of perception, thinking and action. Here, perception begins with prediction. Conflicts between the prediction and sensory information lead to an adjustment of the prediction, thinking about it, and/or a physical action to adapt the situation to the prediction, and/or to investigate the situation further. Cognitive Linguistics provides insight into the structure of our prediction process by emphasising the metaphorical aspect of our thinking.

Both theories can be seen as part of a naturalistic strand of cognitive research. The reason for briefly exploring this side path is that work on artificial neural networks such as LLMs is typically part of it. The precursors of Predictive Processing Theory have a long history, which we shall not discuss further here \cite{NewellSimon-HumanProblemSolving,MaturanaVarela1998,epub00913-Margaret-Boden-Mind-as-Machine,
ebac01035-s11097-025-10055-w, ebac01034-Nikolova2025-NIKPPT-2}.

\section{Thinking styles and conversation partners}\label{denkstijlen}
The American Psychological Association defines thinking as follows \cite{dictionary-APA-org}:
\begin{quote}Thinking. Cognitive behavior in which ideas, images, mental representations, or other hypothetical elements of thought are experienced or manipulated. In this sense, thinking includes imagining, remembering, problem solving, daydreaming, free association, concept formation, and many other processes. Thinking may be said to have two defining characteristics: (a) It is covert -- that is, it is not directly observable but must be inferred from actions or self-reports; and (b) it is symbolic -- that is, it seems to involve operations on mental symbols or representations, the nature of which remains obscure and controversial.\end{quote}

\noindent This is a very general statement. Psychological research into the supposed rational behaviour of Homo economicus, with which the names of the American-Israeli psychologists Daniel Kahneman and Amos Tversky are closely associated, has characterised thinking styles in greater detail \cite{Kahneman-Fast-slow-2012,DeRegtDooremalen2015-EN}. They argue that thinking activity can be broadly divided into two styles. These styles are referred to as System~1 and System~2. System~1 ``operates automatically and quickly, with little or no effort and no sense of voluntary control.'' System~2 ``allocates attention to the effortful mental activities that demand it, including complex computation. The operations of System~2 are often associated with the subjective experience of agency, choice and concentration.'' System~2 takes much more time and effort than System~1. Kahneman et al.\ note that System~1 is by far the most commonly used thinking style in our daily actions. The division into System~1 and System~2 is useful for modelling purposes; in practice, people often use both styles interchangeably and they overlap.

Although the American Psychological Association (APA) does not refer to System~1 and System~2, we believe that the APA’s supplementary definitions provide a useful further elaboration of System~2.
 
\begin{quote}Critical thinking. A form of directed, problem-focused thinking in which the individual tests ideas or possible solutions for errors or drawbacks. It is essential to such activities as examining the validity of a hypothesis or interpreting the meaning of research results.\end{quote}

\begin{quote}Creative thinking. The mental processes leading to a new invention, solution, or synthesis in any area. A creative solution may use preexisting elements (e.g., objects, ideas) but creates a new relationship between them. Products of creative thinking include, for example, new machines, social ideas, scientific theories, and artistic works.\end{quote}

\noindent Furthermore, the APA also distinguishes between \emph{divergent thinking}, \emph{lateral thinking} and \emph{convergent thinking} as variants of critical or creative thinking, in which the emphasis lies respectively on new thinking strategies, the systematic reconsideration of assumptions, and the systematic verification of the logic of thought processes, that is, reflecting on one’s own thinking. We take the liberty of interpreting all these modes of thinking, from \emph{critical} to \emph{convergent}, as characteristic of System~2 thinking. We shall refer to this as analytical thinking hereafter.
 
In the literature on LLMs, we came across the distinction between \emph{reasoning behaviour} and \emph{reasoning performance} \cite{ebac00968-2404.01869v2}. This is worth noting. Our classification clearly focuses on behaviour and not on performance, \emph{accuracy} or \emph{efficiency}. Those aspects are relevant in data centres seeking to process queries quickly, not for a discussion of principles. Another aspect we wish to mention here is the evolutionary basis of human thought \cite{ebac01045-rstb20120111}. We believe that humans represent merely a form of cognition determined by the terrestrial situation \cite[p.~12]{Sebastiaan-Mathot-wereld-vol-denkers-EN}.
 
\section{Metaphorical problem propagations}\label{kernbetoog}
The preceding Sections~\ref{AIbots} through to \ref{denkstijlen} form the basis for the main argument that begins below. As indicated in the introduction, the argument consists of four parts. The first part follows in this section and the other three in Section~\ref{metaforische-probprop-tekst}, \ref{denkruimte-chatbots} and \ref{watvoorp}.

In brief, the four parts deal with: 1. a model of the human thought space; 2. the verbalisation of fragments from that thought space into text, and the properties of that text; 3. the training of LLMs on text and the artificial thought space that this produces; 4. what a chatbot can and cannot do within its artificial cognitive framework, and what implications this might have.

In this argument, the central question is: what kind of conversation partner is a chatbot? Following on from the previous section on thinking styles, we can now refine that question to: is the chatbot more of a System~1 or a System~2 conversation partner? Let us regard conversations as problem-motivated, as stated earlier. The conversation partners want to solve a problem or develop a perspective on it.

\subsection{Problem types and repertoires are metaphors}\label{ptenprmeta}
\noindent Consider the conceptual system from Cognitive Linguistics. Lakoff and Johnson state \cite[p.~3]{Lakoff-Johnson-By-Metaphores}:
\begin{quote} In most of the little things we do every day, we simply think and act more or less automatically along certain lines. Just what these lines are is by no means obvious. One way to find out is by looking at language. Since communication is based on the same conceptual system that we use in thinking and acting, language is an important source of evidence for what that system is like.\end{quote}
Here, we understand language as language-in-action, as concrete linguistic expressions about concrete matters.
What is the purpose of that ``thinking and acting''?

Mark Johnson elaborates on this by, among other things, referring to the philosophy of the American pragmatists William James and John Dewey. Johnson summarises this in five key points; points 3 and 4 are the cherries we would like to pick here \cite[p.~69]{Johnson-embodied-2017}: \begin{quote} \begin{itemize}\item Embodied cognition is problem-centred, and it operates relative to the needs, interests, and values of organisms. \item Embodied cognition is not concerned with finding some allegedly perfect solution to a problem, but one that works well enough relative to the current situation.\end{itemize}\end{quote}

\noindent The American philosopher Donald A.\ Sch\"on uses a few key terms relevant to our argument when he writes about social policy and generative metaphors in a collection \emph{Metaphor and Thought} \cite [p.~138]{Metaphor-and-thought-2ndedition}: \begin{quote}When we examine the problem-setting stories told by the analysts and practitioners of social policy, it becomes apparent that the framing of problems often depends upon metaphors underlying the stories which generate problem setting and set the directions of problem solving.\end{quote}
 
\noindent We interpret \emph{framing of problems} and \emph{metaphors underlying the stories} in this quotation as problem types and problem positions, respectively. We therefore believe that the process described by Sch\"on also applies outside the domain of social policy. There is a further indication of this correspondence with Cognitive Linguistics.
 
In the AD, problem types and solutions are understood as patterns of roles and interactions between parties. In our view, this conception aligns with Cognitive Linguistics, which does not seek the nature of an object's concept in terms of objective aspects and properties. Having a concept of an object is ``to be able to simulate the kinds of perceptual, motor, and affective interactions you typically have with that kind of object'' \cite[p.~24]{Johnson-embodied-2017}. And these interactions yield primary and conceptual metaphors via experiential correlation (see Section~\ref{taalcogni}). So the similarity lies in interactive experiences. Think back for a moment to the metaphor \emph{Knowing Is Seeing} (example from Section~\ref{taalcogni}). Thinking is not seeing. But looking closely at something often goes hand in hand with better understanding. The AD discusses such relationships in terms of roles and role patterns.

\subsection{The notion metaphorical problem propagation}\label{defmpp}
The notion \emph{Metaphorical problem propagation} is a model of the human space of thought and experience that consists of a combination of the problem concepts from Aggregation Dynamics, the conceptual system of Cognitive Linguistics, and the predictive dynamics from Predictive Processing Theory.

This human space of thought and experience is not new; the reader experiences it every moment. What is new, perhaps, is that the combination of these three theoretical perspectives gives us the opportunity to reflect on problem-solution-driven conversations with chatbots. We believe this combination is best explained, to begin with, using an example.

Problem position 1: `A driver is driving on the motorway. The cars in front suddenly brake hard. Problem: how to react. One solution is to brake.' Problem position 2: `The driver is braking. The sub-problem is now: braking adequately. The solution: to brake in such a way that the driver does not collide with the vehicle in front, and the cars behind also have the opportunity to come to a halt.' Problem position 3: `The driver has come to a standstill. The braking manoeuvre was successful. Problem: the driver’s journey and daily schedule are no longer on track.'

The above is a simple example of problem propagation. Let’s make the events a bit more personal.

Problem position 1: `I'm driving on the A9 highway at 8 a.m.\ when I suddenly see the brake lights of the car in front of me flashing on and off. I~want to stay alive. One solution is to brake.' Problem position 2: `I brake. I~still want to stay alive, but the subproblem now is: braking in such a way that I don’t crash into the car in front of me, and the car behind me doesn’t crash into me. The solution is to continue braking appropriately.’ Problem position 3: `I’m at a standstill. I realize I’m still alive. But I’m definitely not going to make it to my 10 a.m. appointment, with the possible consequence that the others in the meeting will decide something I oppose!'
 
The above is a simple example of metaphorical problem propagation. In this example, we see problem-solution relationships, predictions, and metaphorical relationships. Before I had to brake, I was under the assumption: `Everyone drives carefully and keeps their distance; business as usual; I’ll be at my appointment in 15 minutes, right on time.' The brake lights of the car in front of me, which flash on, disrupt that image. Several predictions arise. We’ll mention two: `Someone accidentally taps the brake pedal, if the lights go out immediately and the car keeps moving, I can ignore this red light signal'; `this is actual braking, but how hard are they braking?’ The last prediction calls for closer observation to better understand what is happening. These predictions are part of our conceptual system, which we usually take for granted. In this situation, our conceptual system says \emph{Pumping braking Is Immediate danger}, and that calls for immediate action: braking. This example illustrates how the various elements of metaphorical problem propagation interact, generate meaning, and prepare and activate actions. If we take all these emotions, predictions and metaphorical relationships into account – which we are only doing to a limited extent here – the result is a holistic experience: an example of metaphorical problem propagation.

Metaphorical problem propagation is undoubtedly a complex concept. We find it helpful to visualize the mental architecture of metaphorical problem propagation as follows. The conceptual layering, from primary concepts to conceptual metaphors, constitutes a `vertical' dimension here. In a `horizontal' dimension, the propagations unfold. It is not appropriate to speak of a time axis. That suggests a precision that does not exist, and it would also limit mental freedom, for example -- for other, possibly contradictory -- orderings of events, or for projections into the future or the past. Finally, Predictive Processing Theory adds a third dimension, a prediction dimension.  In this dimension, various predictions operate in parallel, at different levels of abstraction, with different emphases on the role of the conceptual system, and using different words.

Reflecting on current situations, formulating and testing hypotheses, and reflecting on and remembering experiences can thus be understood as the processing of malleable and flexible metaphorical problem propagations. A metaphorical problem propagation is not objective; it is partly unconscious, and chance also plays a role. Some observations stand out, others are missed. Some are perceived as problems, others are not.

Johnson describes a ``representation theory'', referring to the work of J.~Fodor, as ``the idea that human thought consists of a series of functional computational operations on language-like symbols ``in the mind'' that can be used to represent external states.'' Although we speak of it in words and words are often a consequence, our notion of metaphorical problem propagation is not a symbolic representation in this sense. Metaphorical problem propagation is our interpretation of what Johnson, with reference to Lakoff and Gallese, describes as an ``interactionist, multimodal, simulation theory of meaning and thought'' \cite[p.~145]{Johnson-embodied-2017}. In this, we emphasise problem-solution relationships.\\

\noindent We will explore the intersections between the three theoretical perspectives in greater depth. As mentioned earlier, one can envision the conceptual system as a hierarchy with primary metaphors at the bottom and, moving upward, more abstract concepts and sometimes even complex and inconsistent combinations \cite[p.~53]{Johnson-embodied-2017}. The concepts derived from AD simply fit into this hierarchy, with, for example, practical problems corresponding to the primary concepts at the bottom and abstract problems higher up, all the way to the most abstract category, namely that of the `unspecified problem'; ``Houston, we have a problem'' \cite{houston-we-have-a-problem}.

The AD also posits a similar hierarchy, from concrete to abstract, for problem situations, for problems, and for solutions (see the last paragraph of Section ~\ref{patinprop}). The relationships that one mentally brings to life when contemplating a problem position, are experienced through metaphors. We therefore imagine that it is possible to initiate and continue the thought process at different levels of abstraction. And also that it can then proceed in parallel at other levels. So thinking about `the neighbours John and Jack who are arguing over the hedge between their gardens' proceeds, so to speak, also at a more abstract level as $P_1$, $P_2$, and $h$ in relation $r$, and perhaps also at more concrete or even more abstract levels. Or, for example, it proceeds in parallel in another remembered propagation, concerning a different argument, via associations with certain aspects of the context or Jack’s character traits. The freedom to switch between these levels
and to introduce new levels are part of our conceptual system. It is also easy to imagine that it is possible to think backward and forward in the metaphorical problem propagation. Forward and backward exist by virtue of memory and experience; there is no fixed timeline.
 
The problem concepts from the AD and the conceptual system not only align well, but their combination also expands the range of possible applications. The conceptual system of Cognitive Linguistics already provides a dimension for describing action elements. \emph{Image schemas} are ``recurring sensory-motor patterns'‘ that ``afford us possibilities for meaningful interaction with our surroundings, both physical and social’' \cite[p.~21 \& 100]{Johnson-embodied-2017}. An extension of these are \emph{X-schemas} \cite{ebac01054-ebacBailey,ebac00966-NeuralTheoryofMetaphor}, which can be understood as composite processes of sensory-motor patterns. Sensory-motor patterns remain a current topic \cite{ebac01053-role-emotion-in-acquisition-of-verb-meaning}.

But these links between perception and action involve relatively basic interactions and short chains. Aggregation Dynamics operate on a different scale: memories, histories, plans, scenarios, interests, and hypotheses about what opponents are thinking (Theory of Mind). On that scale, it involves long, interconnected sequences of events or projected actions. This vastly expands the action dimension of the conceptual system and makes it relevant in many more situations, which, we believe, is entirely conceivable through metaphors as tools of thought. The sensory-motor patterns are, so to speak, metaphorically stretched to encompass problem situations between, for example, superpowers. This larger scale is not entirely foreign to Cognitive Linguistics, as is clearly evident from the earlier quote from Schön concerning ``social policy.'' There is quite a lot involved there. But it does not seem to be the scale at which one prefers to analyse.
 
But how do these mental images come to life? In his 2017 book, Johnson focuses specifically on the predictive brain \cite[p.~142]{Johnson-embodied-2017}, but Predictive Processing Theory is all about that \cite{AndyClark-ExperienceMachine,AnilSeth-BeingYou}. This theory describes how the brain constantly generates predictions about the situation we find ourselves in and how it will unfold. But these predictions can just as easily pertain to the past, for example, regarding the question, `How could this have gone wrong?' Predictive Processing Theory also explains how, from these competing and often mutually inconsistent predictions, a comprehensive and emotionally coherent picture, the Controlled Hallucination, is nevertheless generated.
 
We have the impression that these predictions are not fleshed out in very concrete terms by Predictive Processing Theory. This seems understandable given the highly technical, empirical setting in which this theory is developed, where the focus is on correlations between sensory stimulation, neural effects, perceptions, and actions. 
But we take the liberty here of stretching the concept of prediction and postulating the metaphorical problem propagations as its structure. 

Predictive Processing Theory deepens the notions of problem and solution.
It is obvious that we identify problems with the mismatch between the predicted and the perceived world. A problem is then a \emph{prediction error}. We naturally identify a solution with adjustments to the prediction or an action. A solution or step towards a solution brings the world into line with the prediction, or brings the prediction and the world closer together.
 
We believe that the concept of metaphorical problem propagation could be an interesting addition to Cognitive Linguistics and Predictive Processing Theory. For Aggregation Dynamics, as an attempt at a total system theory, metaphorical problem propagation represents a significant expansion of its potential applications. Furthermore, the connection with Cognitive Linguistics and Predictive Processing Theory strengthens the theoretical foundation.

Just like the problem propagation from Section~\ref{probprog}, metaphorical problem propagations form a category of mental structures. A metaphorical problem propagation can be on a very large scale, encompassing everything a person knows and has experienced, but it can also concern a fleeting thought fragment that is quickly forgotten, or a fragment centred on a specific problem. In our argument, the context makes it clear what scope we mean, whether it concerns a total mental space, a cluster of related issues, a fragment, or a single thought.

\section{Metaphorical problem propagations and texts}\label{metaforische-probprop-tekst}
In Sections~\ref{probpos} and \ref{probprog}, we anticipated the conversion of the mental world of problem positions and problem propagations into textual or graphical form. Below, we propose a process that describes this transition and ask ourselves what the properties of the resulting text are.

Now suppose someone has a thought and wishes to share it with another person, or tell a story to another. What happens when that thought or story is told? Based on what we postulated above, we imagine that thought or story as a memorised or spontaneously unfolding metaphorical problem propagation. When we convert the metaphorical problem propagation from Figure~\ref{contextprobleempropagatie}, which represents memorized events, into text; this could, for example, yield something as illustrated in Figure~\ref{Verbaliseren-CPP}.

\begin{figure}
\begin{tabular}{l}
\textbf{The situation and problem was}\\
`Los Angeles City 1943; polluted air and fog; stinging eyes'\\
\textbf{The first steps towards a solution were}\\
`Louis McCabe’s research in 1947; research by Arie Haagen-Smit in 1948'\\
\textbf{This identified the causes} \\
`meteorology; peroxides; unburned hydrocarbons; photochemical smog'\\ 
\textbf{The solutions for these problem were sought in two ways}\\
`increasing engine temperature; modified engine'\\
 \textbf{and}\\
`catalytic converter; modified car body'\\
\textbf{But it turned out that}\\
`catalytic converters cannot withstand lead; ...'\\ 
\textbf{which called for}\\
`unleaded petrol; ...'\\
\textbf{This, however, brought back an old problem} \\
`knocking; ...'\\
\textbf{The}\\
`lead additive'\\
\textbf{prevented that, now it was back}\\
`knocking again; ...'\\
\textbf{A solution was found in}\\
`improved engine management; ...'\\
\textbf{this solved knocking in a different way}
\end{tabular}
\caption{\label{Verbaliseren-CPP}A linearisation of the problem propagation shown in Figure~\ref{contextprobleempropagatie}, in which a few words have been added in \textbf{bold} to link the statements together.}
\end{figure}

But this is rather clunky language still. One can imagine that, with a little more editing, a smooth-flowing text could be produced from the list of sequentially arranged statements in Figure~\ref{contextprobleempropagatie}. We envisage the entire process of verbalisation as comprising three transformations. Firstly, the problem positions and the solution steps must be elaborated into text, for instance `Los Angeles City 1943; polluted air and fog; stinging eyes'. The problem positions break down into sequences that already suggest sentences or paragraphs. Secondly, the logic of the network structure must be represented in supplementary text with introductory and concluding sentences, references, headings, layout, summaries and examples. Here these are texts as `\textbf{The situation and problem was}', and `\textbf{The first steps towards a solution were}.' Thirdly, an editorial process turns this into a smoothly flowing text. That final step is not shown here. We call the result of this three-step process a \emph{verbalised metaphorical problem propagation}. In the same way, we can of course also speak of a \emph{verbalised metaphorical problem position}.

Verbalisation is the result of choices: the text ‘1. increased engine temperature 2. modified engine’ could also have been placed after the catalytic converter complications. The ‘mental space’ becomes smaller, but stability increases. Using a physics metaphor, one could say that the metaphorical problem propagation is in a state of superposition; in the text, a choice has been made.

That effect is stronger in written text than in spoken texts, where interaction is still possible. Then someone can still ask a question and the speaker can respond with a nuance and then continue the story. 
Since, our ``body cannot not communicate'' \cite{Watzlawick-nietniet-communiceren}. But our argument concerns written text, as that is the bulk raw material for training LLMs.

\subsection{Structure and contents of text: static text}\label{structuur-inhoud-tekst} 
\noindent If ``all life is problem-solving'', then problem types are important concepts in our thinking \cite[p.~99]{Popper-All-life-problem-solving}. Anyone can observe that the word ‘problem’ crops up frequently in any conversation. We assume this is also true of texts. That is already one characteristic.

Communication requires effort and must yield a benefit for the sender. This is the view put forward by evolutionary biologist Arik Kerschenbaum in \emph {Why Animals Talk.\ The New Science of Animal Communication} \cite{Animal-Talk-Kershenbaum}. The message must therefore be meaningful for the sender’s survival. This in turn suggests that the message must contribute to solving the sender’s current problems. This is most successful when the receiver can understand the message. It is therefore unlikely that anyone, when asked about her problem position, would really bring up absolutely everything under the sun or structure her message as a multidimensional ``stream of consciousness'' as James Joyce did in \emph{Finnegans Wake} \cite{Finnegans-wake,litseller-finwake-EN}. That won’t work for a manual, a report or a leaflet. If the reader loses interest, the text misses its mark. The text must be to the point.

As mentioned, the person producing the text wants to achieve something with it. It is therefore to be expected that there is a certain degree of direction in it and that it is engaging for the user. This is consistent with the distinction made in Section~\ref{probpos} between internal and external problem positions. The external problem position coincides with a sender’s text strategy.

We can further clarify these `text strategies' based on the work of Kahneman et al.\ They argue that System~1 dominates our daily actions. The direct, fast and uncomplicated System~1 suffices in most situations. This seems entirely in line with ``we simply think and act more or less automatically along certain lines'' from an earlier quotation of Lakoff and Johnson (Section~\ref{ptenprmeta}). We suspect that a significant proportion of the text that people produce also has a System~1 character.

In summary, we believe we can hypothesise that most texts are written from the perspective of a strategically oriented sender, deal with relevant issues that this sender wishes to address, are coherent and consistent, point towards solutions, are to the point, are non-experimental, are not difficult to understand, and are tailored to an audience seeking reassurance. Furthermore, these texts, like all written texts, are fixed, static and closed. They cannot be compared to the fluid and associative process of human thought or to people discussing a particular topic. The latter is also a form of text. Yet it is precisely this `limited' type of text that is used for training LLMs.

We argue that examples of these texts can be found in a wide range of categories: newspaper articles; brochures; manuals; website posts; tweets; instruction guides; reports; Wikipedia articles; historical narratives; novels; (popular science) articles. In short, very everyday categories. Of course, within these categories there will be examples that do not quite meet our criteria, such as in-depth background articles or debate reports that record a great deal of interaction. But these remain static nonetheless.

The set of characteristics outlined above is difficult to summarise in a single word. Somewhat useful terms include: professional text, functional text, purposeful text, goal-oriented text, or solution-oriented text. But these are not quite right either and may already have other meanings. As we believe that the lack of flexibility is the most important factor, we refer to the whole as \emph{static} text. This is a nice short term. But keep in mind this notion covers a range of text characteristics.

If we combine the findings of the previous section with those of this one, we believe we can conclude that the text people produce does not represent the intellectual richness from which it stems. Firstly, the verbalisation of any metaphorical problem-propagation already results in a fixation and simplification. Secondly, people do not simply produce text for the sake of it. In most cases, it is simple, purposeful text. This text lists a few points, offers some considerations, draws a conclusion or notes a line of thought or an action, and then moves on to the next point that needs to be made.\\

\noindent We have the impression that researchers analysing text datasets for LLM training focus primarily on the content-related characteristics of texts, such as unacceptable statements \cite{ebac01042-34421883445922,ebac01049-s10462-024-10888-y,ebac01050-naous-etal-2024-beer}. The thought processes and storylines in text datasets receive less attention. The authors of \emph{Narrative Theory-Driven LLM Methods for Automatic Story Generation and Understanding: A Survey} state \cite{ebac01047-liu2026narrativetheorydrivenllmmethods}: 
``Overall, [narrative] researchers rarely conduct unsupervised learning and pre-train LLMs on large volumes of texts themselves'' and ``(...) it cannot be ignored that narrative texts make up large portions of LLM pre-training.'' This assertion supports our hypothesis, formulated above, regarding the structure and content of the texts in the text datasets, namely that they are of a specific type.

They then raise two highly relevant questions: ``What effect does pre-training on more or less narrative texts have on an LLM?'' and ``(...) what kind of narrative artefact is an LLM?'' These questions are also on our minds, and we offer some preliminary answers to them.

These questions are the subject of Section \ref{denkruimte-chatbots} and Section \ref{watvoorp}. We believe the second question is comparable to our own, namely, `what sort of conversation partner is a chatbot?' We will argue that the static nature of the text dataset significantly determines the nature of the LLM as a `narrative artefact'. Assuming, therefore, that we may equate `narrative artefact' with `conversation partner'.

The application of the various narrative theories discussed in the survey compels us to refine our arguments. For example, the quantification of narrativity, which ``denotes the extent to which a text or discourse exhibits the qualities of narrative, including temporal progression, causal linkage, and the presence of agents and events.'' We find this quantification interesting in its own right in relation to text datasets \cite{ebac01048-piper-etal-2021-narrative}. This metric could perhaps also be extended to include a measure of reflexivity or other characteristics we have identified.

\section{The thinking space of chatbots}\label{denkruimte-chatbots}
The raw data used to train LLMs consists of texts that are available online in one form or another \cite{commoncrawl-site,ebac01041-s10462-025-11403-7}: websites, brochures, wikis, novels, history books, e-mails, reports, manuals, accounts, and laws. This material is not immediately suitable for use. There are various reasons for making a selection -- for example, ethical, content-related or technical grounds -- and methods -- for example, automated or partially manual -- to make the texts suitable for training purposes \cite{c4-dataset,ebac01040-3691620.3695061}. We now assume that most of the texts in the text dataset for training publicly accessible systems such as ChatGPT, Claude and Grok are of the, in the previous section, postulated static type.

These static texts share structural similarities because verbalisation is a process of strategic selection and fixation. However, we believe there are further similarities. There are three further reasons why texts within a dataset, however large, are likely to resemble one another: 1.\ Verbalisation creates connections between problems, problem positions and solutions. The Innovation Illusion posits that the majority of combinations of these are familiar patterns and that a limited set of jargon suffices to describe them; 2.\ Problems, problem positions and solutions correlate with conceptual metaphors from the conceptual system. This system is rich in patterns because conceptual metaphors build upon one another; 3.\ Even statements that are not directly related to problems, problem positions and solutions are based on metaphors, because we verbalise through comparisons. These, too, are patterns.
 
It is well known that neural networks are good at recognising patterns. It is well known that LLMs can store human concepts and features. And it is also known that LLMs can encode and utilise analogies. Suppose that this characteristic is not merely incidental but a relatively systematic effect of the training process.\\

\begin{sloppypar}\noindent Our hypothesis on this is now that the LLM training process reconstructs the metaphorical problem propagations inherent in the texts of a text dataset. The resulting artificial metaphorical problem propagation, with varying degrees of coherence, together with word prediction, forms the basis for a chatbot’s ability to conduct a conversation that appears convincing to humans.\\\end{sloppypar}

\noindent Is there any evidence to support this hypothesis? We see a connection between our hypothesis and the empirical results described in \emph{Evidence of a predictive coding hierarchy in the human brain listening to speech} \cite{ebac01036-caucheteux}. In this study, participants listen to stories. Whilst they hear a story, brain measurements track their predictions regarding the story’s continuation. The stories were also fed to LLMs. It appears that: i) the activation patterns in the LLMs ``linearly map onto the brain responses to speech''; ii) the LLM operates on multiple timescales; iii) predictions in humans ``are organised hierarchically: frontoparietal cortices predict higher-level, longer-range and more contextual representations than temporal cortices.'' These three findings appear to align with the parallel we propose between metaphorical problem propagation and its artificial reconstruction and expression in chatbots; metaphorical problem propagation involves projection across time frames of varying degrees of detail; and, the parallel conceptualisation across multiple levels of abstraction.

We find another indication in \emph{Implicit Representations of Meaning in Neural Language Models} \cite{ebac01043-li-etal-2021-implicit}. The authors ``identify contextual word representations that function as models of entities and situations as they evolve throughout a discourse.'' In the authors’ words, the key difference from other studies that empirically attempt to uncover what an LLM encodes, is that they ``aim to recover a representation of the situation described by a discourse rather than representations of the sentences that make up the discourse.’' Based on examples, they succeed in demonstrating that these situations carry meaning because the actions described by the discourses have consequences for the statistical prominence of subsequent situations. We see a clear connection between our problem positions and these \emph{situations}.\\

\noindent Leaving aside the question of whether even larger datasets are a desirable development \cite{ebac01042-34421883445922}. With the current text datasets and LLMs, which are enormous compared to the past, the training process is quite successful in reconstructing metaphorical problem propagations -- assuming our hypothesis is correct -- as evidenced by the public’s positive reception of chatbots. But the reconstruction is not perfect; for example, because a chatbot can make things up and often has no answer to simple questions. Are these minor glitches that just need to be ironed out, or are there fundamental limitations? The question we wish to answer in Section~\ref{watvoorp} is: what kind of metaphorical problem propagation does an LLM simulate, and what can the chatbot do with it?

\subsection{Conversations}
Our hypothesis regarding conversations is as follows. A prompt activates a spectrum of artificial problem positions within the LLM. These are loosely connected groups of loosely connected clusters of neurons that encode concepts.

The activation of an artificial problem position is more or less strong. The degree of activation is the statistical degree of recognition of the prompt as a conceptual pattern. The prompt also activates a spectrum of solution steps (likewise groups of clusters of concepts) and problem positions that result from them, and so on. These are all activated to a lesser degree.
 
The tokens generated by the LLM lie within the ‘contexts’ of the statistically most appropriate problem positions and the statistically most appropriate solution steps. The text generation process has immediate effect: the spectrum of activated groups shifts as a result of the text generation, their composition changes, and the statistical relevance of the groups changes. Some problem positions become more prominent and others less relevant. When text generation stops, the neural network has reached a new state. There are ‘new’ problem positions that now take centre stage in processing the user’s next prompt.

We suspect that the ‘software shell’ surrounding the LLM does not simply follow the LLM blindly, but that it attempts to determine which text generation options yield the `best' conversation. The shell will make choices based on certain parameters, but exactly how is irrelevant here. It may be that, according to the shell, there is no significant activation of problem positions. In that case, the chatbot responds with some seemingly appropriate phrase, such as a request for further explanation.

Depending on the outcome of the text generation process, with or without adjustment or selection by the shell, the chatbot responds with a counter-question, a suggestion, an answer or a solution to the problem posed. The conversation unfolds like a path through the chatbot’s simulated metaphorical problem propagation and, simultaneously yet fundamentally differently, like a path through the user’s metaphorical problem propagation.\\

\noindent In our view, the distinction between a conversation and a prompt is not clear-cut. On both the chatbot’s side and the user’s side, elements from the conversation continue to have an impact, or rather, influence the effect of the most recent prompt. This is essential, because otherwise a user’s response to a chatbot’s question would have no effect: the range of problem positions activated must change. The interaction between conversation and prompt can also be explicitly controlled using \emph{in-context learning} as described in the literature; this should elevate the chatbot to a higher cognitive level \cite{ebac00972-3774896}.

It is tempting to imagine the successive artificial problem positions through solution steps as steps through the neural network, from one group of neurons to a completely different group of neurons. We do not believe that is what usually happens. If a prompt activates a spectrum of artificial problem positions, these will overlap, and they will also overlap with solution steps and the associated problem positions. This is not surprising because choosing a solution step generally only partially changes the problem position. This is certainly the case with complex problems. If someone formulates a prompt $ABC$, and the chatbot suggests $D$ as a solution step, then $ABCD$ is the new problem position. The training process is a form of data compression, and this means that these problem positions become encoded in an overlapping manner.

\section{What kind of conversational partner is the chatbot?}\label{watvoorp}
Suppose an LLM encodes metaphorical problem propagation, as argued above.  What kind of cognitive aid is a chatbot, and what kind of cognitive tasks is a chatbot capable of performing?
 
\subsection{What kind of thing is a chatbot?}
First of all, we need to free ourselves from the thought that chatbots are some kind of magical entity that defies human understanding.
 
Cognitive Linguistics posits that meaning arises from a sentient and experiential body that seeks to navigate a physical environment and a socio-cultural situation. We agree with this. A chatbot does, in a sense, have a ‘body’, but it is not human. The data centre is not a human socio-cultural environment. The chatbot therefore lacks the foundation of human primary subconscious concepts. This is a well-known argument \cite{Mitchell-A-guide-AI-2019}. So whatever form of aggregation a chatbot may be, at best it can simulate human interactions that it finds in text. The words it produces do not have the meaning they have for us.

That does not, however, mean that an intelligent system requires a biological body or that it cannot possess its own form of understanding \cite{ebac01051-Meincke2018}. But that is a different discussion \cite{ebac01045-rstb20120111}. Our argument concerns the basic chatbots to be analysed and, by extension, chatbots in general. These therefore exist within a limited human world of words, and how that relates to the world we experience lies beyond the capacity of the chatbots.

Is it still conceivable, then, that a chatbot could fully simulate our cognitive world, even if the chatbot does not understand what we are talking about?
We find this idea reflected in a hypothesis by blogger janus: ``a model [as an LLM] whose objective is text prediction will simulate the causal processes underlying text creation if optimised sufficiently strongly'' \cite{ebac00943-002404.14082v3}. This statement suggests that if you continue to optimise the model – for example, by using larger datasets, more intricate and extensive neural networks, and a more intensive training regime – you will end up with a model capable of simulating the causal functioning of our body and brain. In our view, this is a vain hope. As concluded at the end of Section~\ref{structuur-inhoud-tekst}, dimensions and associations are lost in the transition from the mental world of human thought to text. That is a fundamental problem. A practical problem associated with this is that people do not express everything that goes on inside them in text. The available text we have to work with for LLM training is predominantly of the static type, as we argued. This text certainly provides an incomplete picture of human cognitive capabilities.

The chatbot is therefore limited in two ways. It is a simulation because the chatbot lacks a human, experiencing body; the simulation is flawed because it is trained on fundamentally incomplete data. It remains an artefact which, in our view, is far removed from all claims regarding Artificial General Intelligence or even stronger claims \cite{Kurzweil-Singularity-Nearer}. In Section~\ref{opvoerenredeneer}, we argue that it is not plausible that ‘more complete’ data would lead to significantly ‘better’ results.

\subsection{Static-thinking, no reflection}
Our hypothesis is that the training text for public chatbots has an static character (Section~\ref{structuur-inhoud-tekst}). This has an effect on the artificial metaphorical problem propagation distilled from it. It seems highly plausible that this, too, therefore has an static character. By this we mean that a chatbot simulates a form of thinking that is more akin to System~1 thinking than to analytical thinking, and is therefore not very critical or analytical \cite{Kahneman-Fast-slow-2012}. This thinking moves swiftly from problems and circumstances to conclusions, or solution steps, or actions, without much reflection. Consequently, whilst the circumstances -- an important part of problem situations -- may well be extensive, the assertions are not treated critically; they are simply accepted. The chatbot generally does not stray from the problem front.

It is difficult to determine precisely to what extent this `thinking' mirrors human thinking, as System~1 thinking is not precisely defined, and there is, of course, a `grey area' where System~1 thinking transitions into more analytical thinking. However, given the success of chatbots, it appears to align well with users’ everyday needs. It seems plausible, in line with Kahneman et al., that if many people rely on System~1-like thinking in many activities, those everyday needs are also characteristic of conversations with chatbots. Let us assume that. Users are then looking for quick answers and accept them if they seem plausible. We suspect that the average user would quickly tire of a chatbot that nitpicks over every little detail or starts nitpicking about supposed assumptions that might not actually be true. No, the chatbot fits neatly into the dominant System~1 part of the thinking spectrum. The chatbot provides what many people are looking for in their daily conversations: confirmation of their world view. They avoid difficult questions if they believe they can achieve their goals with a quick fix.

Our conclusion comes as no surprise to those who critically study the behaviour of chatbots: \begin{quote}\cite{ebac00968-2404.01869v2}: ``It is likely that the apparent success of LLMs in reasoning tasks predominantly reflects their ability to memorise the extensive data they have been trained on.''

\cite{ebac00984-Webster31122025}: ``In terms of the Dual Process Theory of human cognition, LLMs are analogous, at best, to System~1 cognitive processing, and it remains unknown how anything approaching System~2 cognition could be incorporated into a machine.'' \end{quote}

\noindent We, of course, agree with these statements. The difference between our argument and that of these authors is that our model explicitly structures `LLM reasoning' and `cognitive processing' respectively by asking about the motivational basis of knowledge and concepts. Our perspective focuses on the characteristics of the texts on which LLMs are trained and emphasises problem-solving relationships. These two aspects appear to receive little attention in the literature.

\subsection{A thinking aid after all}\label{tocheendenkhulp}
Is the chatbot limited to problem scenarios present in the training data, or is it capable of more? Geoffrey Hinton’s example from Section~\ref{chatbots} shows that chatbots can find apt analogies. According to Hinton, this was an analogy that was not present in the training data. We suspect that this is a fundamental characteristic. So, we suspect that a chatbot could serve as a thinking aid, but how does it go about this, and how far does this thinking aid reach?

Various studies show that people can come up with ideas when working with chatbots. People faced with a creative task perform better when assisted by a chatbot than when left to their own devices or when they only have web searches at their disposal \cite{ebac01079-ChatGPT-on-creativity}.

However, follow-up research prompted by these results shows that when the creative performance of the group being studied is aggregated, creativity decreases. In short, the chatbot steers the group towards the familiar path, the statistically obvious path \cite{ebac01057-Rebecca-Winthrop-NY-27mei2026, ebac01078-ChatGPT-decreases-idea-diversity-in-brainstorming, ebac01060-1-s2.0-S294988212500091X-main, ebac01061-sciadv.adn5290}. In the words of the title of one of these articles: `Generative AI enhances individual creativity but reduces the collective diversity of novel content.'

We therefore note that Hinton puts the word `similarity' into the chatbot’s mouth. The careful design and optimisation of prompts is a topical subject, see, for example, the Google page on prompt engineering \cite{ebac01080-what-is-prompt-engineering}. Work is also underway to gain a better understanding of the ‘thought processes’ a chatbot follows, thereby enabling greater control over them; for example, \emph{chain of thought prompting} \cite{ebac01038-NEURIPS2022_8bb0d291,ebac01039-9798331314385, ebac00972-3774896}. However, `chain of thought prompting' does not appear to be without its problems \cite{ebac01081-kambhampati2026positionstopanthropomorphizingintermediate}. We refer to all these methods and tools for guiding the chatbot as ‘targeted prompting’. Targeted prompting allows the chatbot to search for analogies, metaphors or other relationships between concepts. These concepts or relationships may be new or useful to the user. In the literature, this is referred to as \emph{dark knowledge} \cite{ebac00969-2407.15017v4}. Such a metaphor suddenly frames a phenomenon that is normally described in certain terms using different terminology, and this can lead to new, useful insights or breakthroughs. Will the chatbot successfully get off the mark as a thinking partner through carefully targeted prompts?

You can ask a chatbot what it thinks of a particular assumption. However, you would expect a critical, analytical thinking partner to introduce new elements of its own, to recap – ‘what do we have so far’ or ‘where do we stand now’ – or to breathe new life into a stalled analysis. In the latter case, this might involve uncovering a hidden and incorrect assumption, adjusting a definition, proposing a new hypothesis, going back in the line of reasoning and starting afresh, or re-examining all the steps in a dubious sub-argument. These are all mental strategies which the American Psychological Association categorises as critical, creative, lateral, divergent and convergent thinking. We believe that, with only the conversation transcript to rely on, the chatbot lacks the mental tools to carry out these kinds of mental operations itself. Therefore, the chatbot needs to be guided in order to be of any use in the analytical part of the spectrum. The researchers Taylor Webb, Keith J. Holyoak and Hongjing Lu confirm this view when they state \cite{ebac01008-Holyoak02072024}: ``But there is no reason to suppose that the same system, absent human-generated inputs, would spontaneously develop a disposition to think analogically, as apparently happened at some point in human evolution. Thus, to the extent that LLMs capture the analogical abilities of adult human reasoners, their capacity to do so is fundamentally parasitic on natural human intelligence.'' The chatbot is therefore not a fully-fledged thinking partner.

That is not to say that a chatbot cannot play a role in a thought process, as already indicated above \cite{ebac01079-ChatGPT-on-creativity}. However, in our view, this does require that the results produced by the chatbot in its role as a respondent are reliable and demonstrate systematic execution. Unfortunately, this is also where it falls short.

In the introductory sections, we have already described chatbots as unreliable. It is time to explain what we mean by ‘unreliability’. We use this term to describe the following characteristics: 1. unreliable in the sense of misleading the user; 2. unreliable in the sense that the chatbot harbours biases; 3. unreliable in the sense of being inconsistent, i.e. first saying ‘A’ and later, without giving any reason, saying ‘not A’; 4. unreliable in the sense of not exhibiting the same behaviour in situations that are identical or comparable for the user; 5. unreliable in the sense of fabricating and hallucinating, making things up out of thin air.

In our view, the lack of a physical body largely explains aspects of unreliability 3–5, and contributes to 1–2. We believe that problems of type 4 can be observed in work on analogy puzzles and mathematical problems \cite{ebac01007-lewis2024evaluatingrobustnessanalogicalreasoning, ebac01009-opielka2025analogicalreasoninginsidelarge, ebac01011-glazer2025frontiermathbenchmarkevaluatingadvanced}.
However, points 1–2 are an effect of the training data; people are not always equally reliable, and much text is more or less strategic, and in some cases even manipulative. The concepts employed by the chatbot lack a foundation; the chatbot cannot assess whether a sentence makes sense, or whether it goes too far in manipulating the user; the chatbot lacks \emph{common sense}.

Aspects 3–5 may offer advantages within the analytical spectrum of thought, namely in thinking outside the box. However, the user must be very alert. We believe, however, that these characteristics pose a risk to users who operate in the System~1 part of the spectrum of thought. We will return to this in Section~\ref{gevaarlijkemix}.

\subsection{Enhancing reasoning skills?}\label{opvoerenredeneer}
Could reflective ability be strengthened with an ‘analytical training set’? What would constitute a typical analytical text? We envisage in-depth philosophical inquiry into a subject that is far from mainstream science, which systematically unpicks and weighs up all manner of arguments and which, ultimately, draws no definitive conclusions, only conditional ones. An example might be Jan Verplaetse’s overview and critique of the free will debate \cite{JanVerplaetse-zvw}. We suspect it will not be easy to compile this analytical training set, but that is beside the point.
 
Nevertheless, even if an analytical training set could be compiled, we do not believe that a chatbot could thereby elevate itself to the status of an analytical thinking partner. We suspect that if the simulated metaphorical problem propagation partially loses aspects of its static nature by replacing certainties in the statements with reflective feedback, chatbots might start to engage in more elaborate speculation. You might then be able to use them more effectively as thinking partners, but the need for vigilant supervision of the chatbot also increases.

\begin{sloppypar}Analytical text does not solve the problem of the chatbot lacking an experience-based conceptual system. Even between people, text often falls short when it comes to conveying meaning. But we do possess an experience-based conceptual system. This helps us to reconstruct what another person means. Sometimes this is only possible through repeated reading and rereading (\emph{close reading}) in order to construct a mental structure that is consistent with what you are reading. Even then, you may still be mistaken; you do not really understand the text. Lakoff and Johnson state, as previously quoted, \cite[p.~3]{Lakoff-Johnson-By-Metaphores}:
``language is an important source of evidence for what [the conceptual system] is like.'' Text is, in this respect, a widely available medium. But it is a problematic medium.\end{sloppypar}

Our conclusion, therefore, is that a chatbot, as a conversational partner, is not an analytical thinking partner, nor can it become one with its current architecture and through text-only training. Larger and larger models will not make chatbots more analytical as thinking partners; moreover, there are many other objections to moving in that direction \cite{ebac01042-34421883445922}. Yann LeCun, a leading AI researcher and Meta’s former Chief AI Scientist, states: ``Animals and humans exhibit learning abilities and understandings of the world that are far beyond the capabilities of current AI and machine learning (ML) systems.'' And he does not believe that current AI technology is capable of bridging this gap \cite{ebac01004-LeCun-path_towards_autonomous_mach}. We agree with him on this point.

\subsection{Concept-based interpretability}\label{gevaarlijkemix}
Metaphorical problem propagation
links human thought, human text and the functioning of LLMs. Research
in the field of interpretability shows that it is possible to identify concepts and
properties. If our model is of any use, we suspect it applies to concept-based interpretability (Section~\ref{interp}). After all, our model suggests a meta-structure over concepts. We should point out, however, that we are too far removed from this highly technical field of research to assess the value of our model.
 
In any case, if concept-based interpretability succeeds in unravelling the conceptual structures of LLMs, this might open up the possibility of controlling certain properties, such as the various aspects of reliability. That sounds appealing because the combination of unreliability and robustness -- the chatbot always has a smooth-running response -- has proven dangerous for users who use chatbots too frivolously or are too trusting \cite{ebac00914-nrc-krant-20250920-5725713-je-gevoelens-zijn-echt-chatgpt-adam-zelfmoordplan-EN}. Does concept-based interpretability offer hope that the risk of accidents can be mitigated?
 
As we have already argued above, the reasoning ability cannot be improved and robustness is an inherent characteristic. This leaves only the option of increasing reliability to reduce the risk of accidents. We can identify a number of approaches to achieve this: safety barriers, proofs of correctness and redesign.
 
By ‘safety barriers’, we mean software surrounding an LLM that monitors the conversation and that intervenes when things go wrong. It is comparable to an automatic fire extinguisher in a kitchen with a gas hob. Correctness proofs are mathematically precise methods for demonstrating the properties of software. Redesign involves configuring an LLM in such a way that it always behaves correctly.
 
We believe that all these approaches face a fundamental problem: it is rather difficult to specify precisely which behaviour is acceptable and which is not. After all, we only understand ourselves to a very limited extent. Prejudices, tendencies, and bias present a separate problem within this context \cite{ebac01044-TheNooscopeManifested,ebac01042-34421883445922}. Suppose one were to succeed in specifying an `acceptable' set of desired properties; in that case, we expect that demonstrating those properties in an implementation would be very difficult \cite{Modelling-Analysis-GrooteMousavi}. So, there are two major obstacles, then.

If, however, both obstacles are overcome convincingly enough – that is, if one succeeds reasonably well in specifying the desired behaviour and demonstrating that a system behaves accordingly – then, in accordance with the precautionary principle, we suspect there will be a strong tendency to maximise reliability \cite{RoelPieterman2008-EN}. If that choice is made, we suspect it will be at the expense of the playfulness and associative ability of LLMs. And, we believe, that the usability of chatbots based on these LLMs would then decline significantly. We disagree with a claim in the literature that \emph{hallucination} and \emph{knowledge conflict} are challenges that ``mainly derive from improper learning data'' \cite{ebac00969-2407.15017v4}.

It seems that, if chatbots are to be of any use as creative thinking aids, one must be prepared to settle for a loosely defined and domain-specific ``comfort zone'' \cite{Kleinberg-AI-comfortzone}. And users must themselves take care to remain within that zone and not be carried away by the illusion that the chatbot is an all-knowing innovation. It offers some hope that researchers Eric Bigelow and Tomer Ullman, in \emph{People Evaluate Agents Based on the Algorithms That Drive Their Behaviour}, found that, given the chance, people take into account how a system works before assessing its results \cite{ebac00974-opmia26}.

\section{Conclusion}\label{conclusie}
Chatbots are computer programs that serve as conversation partners or sources of information for humans in written or spoken conversations. Examples include ChatGPT, Grok, Claude, Deepseek, Gemini, and Lumo. 
The purpose of this article is to offer a perspective on the nature of chatbots as genuine conversation partners regarding problems in relation to their solutions. What can chatbots do and what can’t they do in that context, and how can this be explained?\\ 

\noindent Our argument focuses on basic chatbots in the hope of making statements about the core functionality of chatbots in a general sense. Basic chatbots are assumed to consist of a Large Language Model (LLM) with a simple interface.

To better understand the behaviour of these chatbots in problem-solving-driven conversations, we examine the relationship between: human thinking, human text production, the text dataset used for LLM training, and the functioning of LLMs. We introduce the concept of \emph{metaphorical problem propagation} to reflect on this relationship.
 
A metaphorical problem propagation is a model of the human thinking space, or a part of it related to a specific issue, consisting of the combination of the problem concepts from Aggregation Dynamics, the conceptual system of Cognitive Linguistics, and the prediction dynamics from Predictive Processing Theory.

People produce texts. We wonder what the characteristics of those texts are. We argue that the texts used to train chatbots only partially represent human thinking. We argue that this is because these texts are purpose-driven, intended for specific users and closed. We call this \emph{static} text. Furthermore, these static texts share similarities because people tend to repeat themselves when identifying and solving problems; they are less innovative than they think (the Innovation Illusion); and because of the central role played by metaphors – which are closely intertwined – in human conceptualisation, the articulation of thoughts and the composition of texts.
  
It is well known that neural networks can identify patterns. Our hypothesis is that, during the learning process, an LLM reconstructs the implicit, reduced and static metaphorical problem propagations contained within the texts it is presented with. A prompt makes multiple problem positions and solution steps statistically more or less plausible. Whether the LLM works out one of the solution steps (which may also be a request for more information) or whether the chatbot intervenes with a response is not particularly important. The response alters the landscape of activated problem positions and the subsequent development. In that state, the user’s next prompt alters the landscape once more. A conversation can now be defined as a path through the problem positions in the LLM’s artificial metaphorical problem propagations and a path through the user’s metaphorical problem propagations.

A chatbot can think along familiar patterns, look up facts, and generate text based on known lines of reasoning. That is a far cry from analytical thinking. Nevertheless, a chatbot can be coaxed into taking creative `steps in its thinking.' But to do so, the chatbot must be guided. If properly directed, it can find analogies in the enormous database at its disposal, and in its unrestrained nature, it can thus establish relationships between systems, situations, or domains that are new and potentially meaningful. We argue that, for various reasons, chatbot technology cannot be developed into a reliable, independent, analytical thinking partner.\\

\noindent Our conclusions align with Yann LeCun’s scepticism and are at odds with the optimism of Big Tech. That does not change the fact that chatbots exist, that they
are being used on a massive scale, by both individuals and organizations, and that
it is therefore socially and politically important to understand them. A key question for the discussion in this regard is: for which problem is a chatbot a solution? Our article aims to help answer that question.

\subsection*{Acknowledgements}
We would like to thank Atze Bosma and Jos van Wamel for their comments on earlier versions of this article.

\end{document}


\title{Enkele hypothesen over de werking van chatbots in probleem-oplossing-gedreven conversaties: Large Language Model als een bevestiging van de Innovatie-illusie}
\author{S.F.M.\ van Vlijmen en H.D.\ Lethe jr.\\\email{bas@basvanvlijmen.com}}
\maketitle

\begin{abstract}
\noindent We bespreken de aard van chatbots als gesprekspartners bij het oplossen van problemen. Wat kunnen chatbots wel en wat kunnen ze niet? Onze analyse is gebaseerd op inzichten uit de Aggregatie-dynamica, de Cognitieve taalkunde, de Neuropsychologie en de Psychologie.

We stellen vast dat chatbots veelzijdige en samengestelde systemen zijn. Onze redenering richt zich op basale chatbots, in de hoop daarmee uitspraken te kunnen doen over de kernfunctionaliteit van meer geavanceerde chatbots. We gaan ervan uit dat basale chatbots bestaan uit een Large Language Model (LLM) met een eenvoudige interface.

De belangrijkste resultaten van onze analyse zijn: een beschrijving van menselijk voorstellingsvermogen, begrip en denken op basis van zogenaamde metaforische probleempropagaties; dat de teksten in de tekstdatasets die voor het trainen van LLMs worden gebruikt, specifieke kenmerken hebben en dat deze teksten het menselijk denken en begrip slechts gedeeltelijk nabootsen; dat het LLM-trainingsproces kunstmatige metaforische probleempropagaties uit deze tekstdatasets in een LLM codeert.

Onze conclusies zijn dat een basale chatbot geen denkpartner kan zijn die de cognitieve flexibiliteit van mensen kan evenaren, en dat verdere ontwikkeling van LLMs hier evenmin toe zal leiden.

Maar chatbots bestaan, ze worden op grote schaal gebruikt, zowel door individuen als door organisaties. Daarom is het maatschappelijk en politiek belangrijk om ze te begrijpen. Ons artikel wil een bijdrage leveren aan de discussie over de werking, voordelen en nadelen van chatbots. 

De cognitieve taalkunde laat zien hoe het gebruik van metaforen een uitdrukking is van ons denken. Aggregatie-dynamica is een poging tot een alomvattende systeemtheorie. Wij zijn van mening dat het concept van metaforische probleempropagatie een interessante aanvulling zou kunnen vormen voor beide.

Chatbots een oplossing? Waarvoor?
\end{abstract}

\section{Inleiding}
Chatbots overtuigen! Waarom en waardoor? Chatbots komen voor een groot aantal gebruikers als een volwaardige, bruikbare en betrouwbare gesprekspartner over. Dat is zondermeer een uitzonderlijke prestatie van hun makers. Maar het succes van chatbots roept ook veel vragen op.

Het kennelijk creatieve en intelligente gedrag van chatbots is, vooralsnog, moeilijk te kwalificeren, te controleren en te verklaren. Wat zijn hun eigenschappen, wat kunnen chatbots nu wel en wat niet? Robuust zijn ze wel. Onbetrouwbaar, op verschillende manieren, zijn ze ook. Wat kan je veilig aan ze overlaten en wat zeker niet? Deze kwesties hebben de aandacht van wetenschappers, wetgevers en filosofen \cite{ebac00943-002404.14082v3, ebac00969-2407.15017v4, AI-snake-oil, ebac01025-NIST-FourPrinciplesExplainableArtificialIntelligence,ebac01044-TheNooscopeManifested, AI-ACT-EU, daEmpoli-Het-uur-vd-wolven,Brockman-WhatToThinkAboutMachenesThatThink}.

\sloppypar{In ons artikel proberen we door de combinatie van inzichten uit vier verschillende onderzoeksdomeinen een perspectief te bieden op de werking van Large Language Models-gebaseerde chatbots en op de vrijheidsgraden van probleem-gemotiveerde conversaties met deze chatbots. De vier domeinen zijn: de Aggregatie-dynamica, de Cognitieve taalkunde, de Neuropsychologie en de Psychologie. We benadrukken dat het om probleem-gedreven conversaties gaat, want in dat soort mens-chatbot-conversaties komt, vermoeden we, het duidelijkst bovendrijven wat een chatbot kan en niet kan.}\\

\noindent \textbf{Aggregatie-dynamica (AD): de Innovatie-illusie}\\
De AD is een uitbreiding van de ecologie. De ecologie bestudeert de verspreiding en concentratie van organismen uit de eigenschappen en interacties van organismen, onderling en met hun omgeving \cite{ecology-begon-townsend5e}. De AD vindt organismen en omgeving te beperkte abstracties wanneer men systemen van mensen, producten, organisaties en culturen wil begrijpen. De AD streeft naar een ecologie zonder uitzonderingen, streeft naar totaliteit \cite{transmathematica-about-EN-21-mei-2026}. De AD meent dat er dan een veel ruimere categorie van interacterende spelers nodig is. Deze ruimere categorie van spelers noemt de AD aggregaties. Een aggregatie is elk \emph{iets} dat door een speler als speler ervaren wordt. In de AD tellen, bijvoorbeeld, als gelijkgerechtigde spelers mee: projecten, autofiles, een combinatie van een taartbordje met kruimels van een koekje en een gebruikt koffiekopje van een ander servies, bacteri\"en in hun agaragar-bakje, de geschiedenis van politieke dossiers, de jaloezie der goden en natuurlijk ook chatbots. Om te begrijpen wat mensen bezighoudt en wat zij ondernemen, kijkt de AD onder andere naar datgene wat mensen als kans of hindernis ervaren, wat zij als probleem benoemen en welke oplossingen zij hiervoor bedenken en toepassen. De AD beweert dat mensen minder innovatief zijn in het stellen van problemen en bedenken van oplossingen dan ze zelf denken. De AD noemt dit de Innovatie-illusie.\\

\noindent \textbf{Cognitieve taalkunde: het metafoorparadigma}\\
\noindent De Cognitieve taalkunde is gefascineerd door de relatie tussen taal en denken. Een belangrijke rol wordt toegedicht aan metaforen. Metaforen zijn veel meer dan alleen uitzonderlijke dichterlijke stijlfiguren, blijkt uit de analyses van de Cognitieve taalkunde. Ons denken en onze teksten ritselen er juist van. De Cognitieve taalkunde betoogt dat metaforen de motor zijn van het menselijk denken.\\ 

\noindent \textbf{Neuropsychologie: de Predictive Processing Theory}\\
\noindent Een lang geaccepteerde theorie over waarneming, denken en handelen stelt dat de ogen, en andere bekende zintuigen, aan de basis staan van een proces dat een gedetailleerd  mentaal beeld van de wereld genereert. Met dat beeld, en na denken erover, kan gehandeld worden. In de Predictive Processing Theory, daarentegen, vormt het brein al hypothesen over de situatie waarin het denkende lichaam verkeert. De waarneming, via veel meer sensorische bronnen dan alleen de bekende zintuigen, dient om die hypothesen vervolgens te toetsen of bij te stellen. In deze theorie is het onderscheid tussen denken en handelen flu\"ide. Handelen is een vorm van denken die dient om de waarneming te verfijnen, door bijvoorbeeld beter te kijken,  of om in te grijpen in de wereld en deze daardoor in overeenstemming te brengen met de hypothesen erover.\\

\noindent \textbf{Psychologie: denkstijlen}\\
Aan de Psychologie ontlenen we definities van denken, denkmethoden en de indeling van het menselijk denken in een snel-tevreden denkstijl (Systeem~1) en een systematische analytische denkstijl (Systeem~2).\\

\noindent \textbf{Het betoog op hoofdlijnen}\\
\noindent De Britse cognitiewetenschapper Margaret Boden beschreef het onderzoeksgebied van de Artifici\"ele Intelligentie (AI) met: ``[it] seeks to make computers do the sorts of things minds can do'' \cite{MargaretBoden-AI-2018}. Deze korte karakterisering is de start voor ons betoog. Als wij iets over chatbots willen zeggen, moeten we eerst idee hebben over die \emph{minds}. Het betoog bestaat uit vier onderdelen die we hieronder kort weergeven.

\textbf{1.} We formuleren een model van min of meer bewuste denkruimten van mensen over de actuele situaties waarin ze verkeren, hun herinneringen en hun plannen. Deze denkruimten zijn associatief en flexibel en de geest kan er doorheen `dwalen', er aan sleutelen, in- een uitzoomen, van concreet naar abstract gaan en terug, en deze denkruimten kunnen ook naar believen worden uitbreid of beperkt. We noemen dergelijke denkruimten: \emph{metaforische probleempropagaties}. Ons model ontstaat uit de integratie van het conceptuele systeem van de Cognitieve taalkunde (begrip, ervaring) met de probleem- en oplossingsdynamiek van de Aggregatie-dynamica (motivatie, richting, interactie) en de voorspellingshypothese van de Predictive Processing Theory (waarneming, denken, handelen). 

\textbf{2.} Large Language Models zijn vooral gebaseerd op schriftelijke teksten die ontsloten zijn op het internet. Deze teksten zijn het resultaat van een reductieproces. Want teksten, zowel mondeling als schriftelijk, zijn altijd het resultaat van een selectie- en combinatieproces. Bovendien, ontberen schriftelijke teksten situatie en lichamelijkheid. We beweren dus dat de meeste schriftelijke teksten uit datasets slechts ten dele correleren met de menselijk mentale mogelijkheden.

\textbf{3.} Wij stellen dat een substantieel deel van de teksten waarop LLMs getraind worden handelt over de relatie tussen problemen en oplossingen. Teksten zijn altijd al gereduceerd materiaal en deze teksten kennen bovendien een beperkt repertoire aan structuurpatronen die met de relatie tussen problemen en oplossingen samenhangen.

Onze hypothese hier luidt nu dat het trainingsproces in een Large Language Model fragmenten van kunstmatige metaforische probleempropagaties construeert. Na training, in een conversatie, raken deze propagaties minder of meer geactiveerd en hebben daarmee, als waren het rivierbeddingen, een aanzuigende werking op de loop van de woordgeneratie.

Een probleem-gemotiveerde conversatie kan nu gedefinieerd worden als een pad door metaforische probleempropagaties, zowel aan de chatbotkant als aan de kant van de gebruiker. Maar die propagaties  verschillen van elkaar. Over die verschillen gaat het in de laatste stap van het betoog.

\textbf{4.} We betogen dat een basale chatbot geen analytisch denkvermogen kan simuleren, omdat deze `de neiging' heeft de `rivierbeddingen' van de kunstmatige metaforische probleempropagaties op te zoeken en die vertegenwoordigen een reduceerde vorm van denken. Aangezien chatbots niet beschikken over een lichaam en geen ervaring kennen, zijn ze niet instaat om zijn eigen propagaties te beoordelen en te corrigeren qua doeltreffendheid. Dat betekent niet dat een chatbot niet in staat is tot bepaalde denkondersteuning, maar deze moet dan wel bij de hand worden genomen door een alerte en creatieve promptmaker.\\

\noindent \textbf{Opbouw van het artikel}\\
\noindent In Sectie~\ref{AIbots} tot en met \ref{denkstijlen} volgen respectievelijk op dit artikel toegesneden inleidingen tot: AI, chatbots en neurale netwerken; de gebruikte inzichten uit de Aggregatie-dynamica, de Cognitieve taalkunde, Neuropsychologie en Psychologie. Iemand die met dit materiaal vertrouwd is kan het ook overslaan en beginnen in Sectie~\ref{kernbetoog}.

De kern van het betoog, in de eerder beschreven vier stappen, volgt in Sectie~\ref{kernbetoog} tot en met Sectie~\ref{watvoorp}. In de conclusie, Sectie \ref{conclusie}, vatten we alles samen.

\section{AI, chatbots en modellering van LLMs}\label{AIbots}
De Artifici\"ele intelligentie (AI) is een multidisciplinair vakgebied. En is, relatief t.o.v. de actuele hype ook een oud vakgebied. De ontwikkeling van AI ging gelijk op met de ontwikkeling van rekenautomaten (wat we nu computers noemen) vanaf de jaren 1930 \cite{Mitchell-A-guide-AI-2019,AlbertsVanVlijmen-Computerpioniers2016-EN}. De vraag waartoe computers in staat zijn, werd destijds al gesteld en heeft generaties onderzoekers en ingenieurs beziggehouden. Exemplarisch is hier het werk van Alan Turing. Zie voor een overzicht \emph{Mind As Machine: A History Of Cognitive Science} van de Britse filosoof en cognitiewetenschapper Margaret Boden \cite{epub00913-Margaret-Boden-Mind-as-Machine}. De term \emph{artificial intelligence} stamt uit 1956 \cite[p.~331]{epub00913-Margaret-Boden-Mind-as-Machine}, eerder sprak men van computersimulatie. 

Volgens Boden kent de theorievorming over AI vijf hoofdstromingen \cite{MargaretBoden-AI-2018}. Die hangen samen met de wijze waarop informatie gerepresenteerd en verwerkt wordt. Eén ervan is ge\"inspireerd op gegevensverwerking in organismen: fout-tolerant, parallel en gedistribueerd. Kunstmatige neurale netwerken komen voort uit deze stroming, ook wel bekend als connectionisme. 

Kunstmatige neurale netwerken zijn ge\"inspireerd op de neurale netwerkstructuren in de hersenen en de wijze waarop neuronen daarin chemische signalen van elkaar ontvangen, verwerken en het resultaat daarvan weer doorgeven aan organen of weer andere neuronen. Kunstmatige neurale netwerken werken op een vergelijkbare manier. Een kunstmatig neuron leest numeriek signalen van de invoerpoorten waarmee het met andere neuronen verbonden is, maakt een rekensom hiermee en zet het resultaat op zijn uitvoerpoorten. 

Hoe `weten' de neuronen in een neuraal netwerk welke rekensom ze moeten maken? Dat weten ze aanvankelijk maar zeer ten dele. Het vergt een trainingsproces om ze de details te leren. Hierin leert elk neuron en het netwerk als geheel stap voor stap de gewenste relatie tussen invoer en uitvoer. Bijvoorbeeld, dat bij een afbeelding met een kat prominent in het midden het woord 'kat' hoort en bij een afbeelding met een hond het woord 'hond'. 

Om de werking heel eenvoudig te verwoorden, zou je kunnen zeggen dat een neuraal netwerk een tabel implementeert. Dat klinkt omslachtig en dat is het vaak ook, voor bijvoorbeeld een telefoonboek. Maar neurale netwerken hebben ook grote voordelen. Het is een one-size-fits-all rekenmodel: in principe kan je hetzelfde neurale netwerk een telefoonboek leren, maar ook hoe vragen te beantwoorden als helpdesk. Verder, neurale netwerken kunnen tegen enige beschadiging. Stel je een groot neuraal netwerk voor dat gedistribueerd is over 100 computers. Valt een van die computer uit dan is het netwerk niet noodzakelijk meteen onbruikbaar. Bovendien zijn neurale netwerken niet kieskeurig met betrekking tot hun invoer. Een voorbeeld biedt de bekende app Pl@ntNet. Deze app is getraind op het op naam brengen van een plant op basis van een foto ervan. Biedt je het een heel vage foto van een plant aan dan probeert de app daar toch iets zinnigs over te melden. Beide laatste eigenschappen worden robuustheid genoemd. 

Maar, neurale netwerken hebben ook nadelen: ze zijn onbetrouwbaar. Pl@ntNet zou, bij wijze van spreken, kunnen melden bij een foto van een fiets dat het met 100\% zekerheid een eik is. Voor meer details verwijzen we naar de overvloedige literatuur over dit onderwerp, bijvoorbeeld \cite{Mitchell-A-guide-AI-2019}.

Een Large Language Model is een bepaald type neuraal netwerk. Deze netwerken zijn niet getraind op afbeeldingen koppelen aan de woorden kat of hond, maar getraind op tekst en tekst die daar plausibel op aansluit. Bijvoorbeeld, bied je een LLM het woord 'ik' aan dan is de kans groot dat het 'ben' teruggeeft. Want, 'ik ben' is een heel gewoon zinnetje of een begin ervan. Een Large Language Model vormt het hart van een chatbot.

\subsection{Chatbots en basale chatbots}\label{chatbots}
Met de term chatbot doelen wij op voor een groot publiek toegankelijke systemen als ChatGPT, Claude, Grok, Lumo enzovoort. Deze zijn niet ingericht voor het ondersteunen van heel specifieke taken, ze zijn getraind op bulktekst van het Internet. 

Toch zijn chatbots moeilijk onder \'e\'en noemer te vangen. Het zijn in belangrijke mate pluriforme en complexe of hybride systemen, die ook nog eens volop in ontwikkeling zijn. Het zijn vaak combinaties van verschillende softwarecomponenten die stoelen op verschillende inzichten en verschillende bouwwijzen. Het belangrijkste onderdeel van een chatbot is het Large Language Model (LLM). Andere componenten in een chatbot regelen, als interface, de interactie tussen het LLM en de gebruiker, leveren aanvullende functies en modereren het proces. We zouden dat een softwareschil om het LLM kunnen noemen. Zo zorgt deze  `schil' dat het LLM geen tekst uitslaat die onwenselijk geacht wordt, bijvoorbeeld racistische taal of een advies over su\"icide. Andere componenten voeren de regie over het LLM bij het samenvatten van tekst of het aanpassen van de stijl van een tekst \cite{ebac01027-Basyal2023TextSU, ebac01028-info14060303,ebac01029-3419106, ebac01030-coli_a_00426}.

Vermoedelijk nodigen LLMs uit tot hybride vormen, want een natuurlijke taal interface tot allerlei gereedschap is natuurlijk heel aantrekkelijk. De Amerikaanse wiskundige, auteur en softwarebouwer Stephen Wolfram beschrijft de combinatie van rekensysteem Wolfram|Alpha en een chatbot \cite{Wolfram-ChatGPT-2023}. Er zijn veel meer voorbeelden \cite{ebac01017_ICCC-2024-Proceedings,ebac01018_ICCC-2025-Proceedings}. Hoe de actuele publiekschatbots precies in elkaar steken, is voor buitenstaanders, die ver afstaan van de afgeschermde bouwpraktijk, niet te zeggen.

De hybride vorm van chatbots, die wij veronderstellen, heeft consequenties voor ons betoog. Want over welke functionaliteit van welk onderdeel moeten we het hebben? Laten we een basale chatbot defini\"eren als bestaande uit een Large Language Model met een eenvoudig interface. Als we in het vervolg spreken over chatbots en conversaties dan bedoelen we chatbots beperkt tot de functionaliteit van deze hypothetische basale chatbots. Wij gaan ervan uit dat deze abstractie zinvol is. Dat wil zeggen, dat uitspraken over basale chatbots ook relevantie hebben voor de met extra functionaliteit uitgebreide vormen ervan. Wat wij betogen in dit artikel gaat namelijk in hoofdzaak over wat een LLM vermag en wat dat betekent voor het actuele gebruik en de toekomstige mogelijkheden.

Het Large Language Model is de tekst-genererende motor van een chatbot. Het LLM is in staat om gegeven willekeurig welke invoertekst, op statistische gronden, te voorspellen welke tekst daar als vervolg op aan zou kunnen knopen. Het generatieproces van die tekst gaat per woordfragment, een token genaamd, of per woord, of nog langere reeksen. De LLMs van de eerder genoemde actuele chatbots genereren tekst per token. We vervolgen het eerdere voorbeeld, waarbij we voor het gemak het generatieproces opvatten als `per woord'. Als 'ik' de invoer is dan zal 'ben' in de uitvoer hoog scoren, maar `heb' is ook een goede kandidaat. Daarentegen zullen `verkeersbord' of `olifant' laag scoren. Want, 'ik verkeersbord' is niet een doorsnee zin, behalve wellicht in absurdistisch theater. Er is dus niet \'e\'en vervolg mogelijk, maar er is sprake van een lange reeks van mogelijkheden, elke met een eigen waarschijnlijkheid. Uit die reeks wordt een keuze gemaakt. 

Natuurlijk, er kan ook geen plausibel vervolg blijken te zijn, volgens de chatbot. Dan stopt het tekstgeneratieproces of grijpt de schil rond het LLM in. Op welke gronden? Stephen Wolfram stelt over de keuzeregels en inrichtingsregels van een LLM \cite{Wolfram-ChatGPT-2023}: \begin{quote}Particularly over the past decade, there’ve been many advances in the art of training neural nets. And, yes, it is basically an art. Sometimes -- especially in retrospect—one can see at least a glimmer of a ``scientific explanation'' for something that’s being done. But mostly things have been discovered by trial and error, adding ideas and tricks that have progressively built a significant lore about how to work with neural nets.\end{quote} Een LLM is een product waarvan de werking grotendeels onverklaard is, zie Sectie~\ref{interp}. Het werkt in conversaties desalniettemin zeer overtuigend. Maar hoe blijft dus de vraag.

LLMs zijn getraind op tekstfragmenten en de plausibele tekstuele opvolging ervan. Dat gaat prima. Chatbots produceren moeiteloos goed lopende zinnen. Dat was wat onderzoekers en ontwikkelaars ook hoopten, maar ze kunnen meer.

Het bleek tot veler verbazing dat LLMs ook in staat zijn tot wat lijkt creatieve redeneringen. De Britse onderzoeker Geoffrey Hinton, een grote naam uit de AI, geeft als voorbeeld de chatbot-reactie op zijn vraag: `Wat is de overeenkomst tussen een hooiberg en een atoombom?' \cite{Hinton-interview-DOAC-Steven-Bartlett}. De chatbot antwoordde dat het allebei zichzelf aanjagende verbrandingsprocessen zijn. Dat is een rake analogie. Een vorm van analogievaardigheid werd al eerder vastgesteld op de semantische netwerken die woorden coderen, wat onderdelen van LLMs zijn \cite[p.~248]{Mitchell-A-guide-AI-2019}. Onderzoek naar de capaciteit tot analogievinding is een actueel onderwerp binnen het AI-onderzoek  \cite{ebac00978-lee2025curiouscaseanalogiesinvestigating,ebac01009-opielka2025analogicalreasoninginsidelarge,ebac01010-qin2025relevantrandomllmstruly}. We nemen aan dat LLM enige `gevoeligheid' hebben op dit vlak, deze vaststelling is voor ons betoog voldoende.

Neurale netwerken hebben ook andere mooie eigenschappen, zoals robuustheid; en minder mooie, zoals onbetrouwbaarheid. Deze eigenschappen spelen een rol aan het slot van het betoog, namelijk in Sectie~\ref{watvoorp}. Wat we onder onbetrouwbaarheid verstaan zullen we daar uitleggen.

\subsection{\emph{Interpretability}: modellen van neurale netwerken}\label{interp}
De vraag naar het `hoe' van neurale netwerken is het onderwerp van het onderzoeksterrein \emph{Interpretability} \cite{ebac00969-2407.15017v4}. Hierbinnen proberen onderzoekers het emergente gedrag van neurale netwerken te verklaren uit het gedrag van individuele neuronen door modellen van neurale netwerken te maken en door ontwerpaanpassingen. Leonard Bereska en Efstratios Gavves beschrijven in \emph{Mechanistic Interpretability for AI Safety -- A Review} de stand van zaken voor LLMs \cite{ebac00943-002404.14082v3}. De auteurs merken op dat de ontwikkeling rond interpreteerbaarheid parallellen heeft met die in de Psychologie. 

De Psychologie begon met het observeren van gedrag en met theoretiseren, want de technieken om te meten aan hersenprocessen of deze te simuleren bestonden nog niet. Stap voor stap werden die technieken ontwikkeld binnen nieuwe disciplines als Cognitiewetenschap, Neuropsychologie en Neurologie. Het onderzoek aan AI-interpreteerbaarheid vertoont een vergelijkbaar spectrum aan benaderingen. Bereska en Gavves onderscheiden er vier die zij alsvolgt beschrijven (schuinschrift van ons): ``\emph{Behavioral} analyzes input-output relations; \emph{Attributional} quantifies individual input feature influences; \emph{Concept-based} identifies high-level representations governing behavior; \emph{Mechanistic} uncovers precise causal mechanisms from inputs to outputs.''

Ons betoog gaat voor een belangrijk deel over concepten en de samenhang van concepten. Zie Sectie~\ref{patinprop} en de bespreking van het conceptuele systeem van de Cognitieve taalkunde. Voor ons lijken de resultaten uit de Concept-gebaseerde en de Mechanistische interpreteerbaarheid het meest aan te sluiten bij ons betoog. Door te `meten' in het netwerk proberen interpretability-onderzoekers netwerkfragmenten te isoleren die corresponderen met bepaalde concepten en activatiepatronen die herleidbaar zijn tot gedrag van het kunstmatige neurale netwerk \cite{ebac00967-s42256-025-01084-w}. 

Er zijn duidelijke aanwijzingen dat concepten niet willekeurig over het netwerk gedistribueerd zijn, maar als clusters te lokaliseren zijn. En ook dat verwante concepten overlappend of in elkaars buurt kunnen liggen. Dat betekent niet dat alle concepten zo zijn opgeslagen. Echter, wij nemen aan dat overlap betekent dat als van verwante concepten A en B, A geactiveerd wordt de kans toeneemt dat ook B geactiveerd raakt. Dat deze overlap optreedt is verklaarbaar: het bespaart opslagruimte en verhoogt daarmee de precisie, en dat is nou precies wat de training van een neuraal netwerk beoogt.

\section{Aggregatie-dynamica}\label{ADInno} De filosoof Karl Popper gaf een bundel essays de titel \emph{All life is problem solving} \cite[p.~99]{Popper-All-life-problem-solving}. We zien dat ook zo. De interacties tussen aggregaties gaan over problemen. Om niet meteen `life', dus alle problemen van alle aggregaties aan te pakken, maakt de Aggregatie-dynamica een begin door problemen en oplossingen zoals ervaren door mensen te onderzoeken. 

\subsection{Probleem en probleempositie}\label{probpos}
Stel je ervaart iets dat niet klopt met je wensen, behoeften of ambities. We noemen dat een probleem. Maar het probleem wordt pas concreet als je het stelt of concretiseert: je neemt een probleempositie in. Een probleempositie detailleert wat we zien als de relevante aspecten bij het denken erover en/of communiceren erover. Typische elementen van een probleempositie zijn: prangende kwesties, indelingen in hoofd- en subproblemen, potenti\"ele oplossingsrichtingen, faseringen, achtergrondkennis, feiten, geruchten, meningen of visies van andere betrokkenen, ervaringen met bepaalde oplossingen, prioriteiten, de stand van zaken met de uitvoering van bepaalde subproblemen, verklarende geschiedenissen. Merk op dat er sprake is van afwegingen over kandidaat oplossingen, maar niet van de uitvoering van de zich aandienende oplossingsstap zelf. Een probleempositie is een momentopname waarbij even de adem wordt ingehouden en,  bijvoorbeeld, de vraag wordt gesteld: `waar staan we nu?' 

Een probleempositie hoeft niet een consistent beeld te geven. De ene adviseur zegt bijvoorbeeld `doen', de andere zegt `niet doen'. Sterker nog, de positie is vermoedelijk meestal niet consistent, behalve in het geval van uiterst triviale problemen. Het is vaak juist het gebrek aan consistentie dat bijdraagt aan de ervaring dat er een probleem is. 

De AD beweert ook dat het zinvol is om onderscheid te maken tussen drie typen probleempositie: de natuurlijke, de interne en de externe \cite{ebac00899-LosAngelesSmog-probleemposities}. De natuurlijke is de probleempositie die zich uitstrekt tot diep in de lichaamstoestand van iemand die zijn probleempositie overdenkt. Deze is slechts deels toegankelijk voor het bewustzijn. De interne is de bewuste mentale voorstelling. De externe is nog steeds een mentale voorstelling, maar op een ontvanger toegesneden. Dat kunnen er meerdere zijn: `als iemand vraagt hoe het project loopt dan zeg ik \emph{zus} tegen A en \emph{zo} tegen B.'

De AD beweert dat interne en externe probleemposities hanteerbaar zijn qua omvang en complexiteit \cite{ebac00899-LosAngelesSmog-probleemposities}. Dit geldt zowel voor de mentale voorstelling als voor een onder woorden gebrachte probleempositie. Er wordt verondersteld dat een eenvoudig probleempositiejargon, een soort formule-taal, reikt voor de beschrijving. Een probleempositie beschreven in dat jargon dekt op compacte, droge manier de lading. Dus in probleemposities gaat het over een beperkt aantal observaties, beweringen, feiten, relevant geachte gebeurtenissen uit het verleden, potenti\"ele vervolgstappen, spelers en een beperkt aantal relaties tussen dit alles. 

Er zijn geen harde grenzen aan deze mentale voorstelling en er zijn geen vormeisen. Het is wat iemand, geheel persoonlijk, als geheel ervaart en wat deze van belang acht. De woorden van Amerikaanse filosoof John Dewey (1859-1952) zoals geciteerd door Mark Johnson lijken hier passend. De probleempositie bevat dat wat gevoelsmatig aansluit op de ``pervasive unifying quality'' van het moment \cite[p.~52]{Johnson-embodied-2017}. 

Tot slot neemt de AD aan dat, mits men zorgvuldig werkt, onder woorden gebrachte versies van probleemposities een beeld kunnen geven van de complexiteit van de interne probleempositie van iemand \cite{ebac00899-LosAngelesSmog-probleemposities}. Dat wil zeggen, je kan door zorgvuldig interviewen, of door iemand te vragen zijn interne probleempositie eerlijk en volledig uit te werken, of door zorgvuldig archiefonderzoek een redelijk scherp beeld krijgen dat qua complexiteit correleert met de complexiteit van een interne probleempositie. Maar de natuurlijke probleem positie wordt verondersteld grotendeels buiten bereik te blijven.

Alhoewel taal een instrument is om een beeld te krijgen van de complexiteit van een interne probleempositie wil dat niet zeggen dat willekeurig welke tekst de vereiste diepgang al biedt. In Sectie~\ref{metaforische-probprop-tekst} werken we dat verder uit en zullen we betogen dat veel door mensen geproduceerde tekst de complexiteit van probleemposities niet heeft. Onze bewering is: de meeste tekst representeert een situatie als stabiel en begrepen en stuurt vervolgens naar een doel waarbij maar met een beperkt aantal variabelen rekening wordt gehouden.

\subsection{Probleempropagatie}\label{probprog}
Men zou het handelen van mensen in kaart kunnen brengen door te stellen dat ze van probleempositie naar probleempositie gaan via oplossingen of stappen in die richting. Even los van de vraag of de geobserveerde stappen nu goed doordacht zijn of niet, zou je dan het volgende kunnen constateren. Voor triviale problemen kan de kous af zijn na \'e\'en oplossingsstap. Voor complexere problemen moeten vaak series van stappen ondernomen worden. Simpele lineaire series van stappen zijn dat meestal niet. De aanpak van een probleem kan botsen met de aanpak van een ander probleem. Of, de aanpak kan ingewikkelder lijken dan eerst gedacht en nopen tot het splitsen van het probleem in meerdere subproblemen. Probleemposities hangen dus via verschillende relaties samen, door onderlinge afhankelijkheden en door oplossingsstappen. 

De Aggregatie-dynamica noemt zo'n netwerk van probleemposities die via oplossingstappen in elkaar overgaan een probleempropagatie. Propagatie moet men lezen als voortplanting, uitbreiding of groei: zowel de aanpak van het ene probleem en de oplossing ervan kunnen tot nieuwe problemen leiden. Simpelweg omdat de wereld niet stilstaat.

Als men een probleempropagatie in kaart brengt, kan men daarin terugblikken (geschiedenis) en vooruit kijken (plannen, scenario's). Het `heden' in een netwerk noemt de AD het probleemfront. Aan het front kunnen probleemposities botsen en worden keuzes gemaakt. 

\begin{figure}
\centering
\includegraphics[height=9cm]{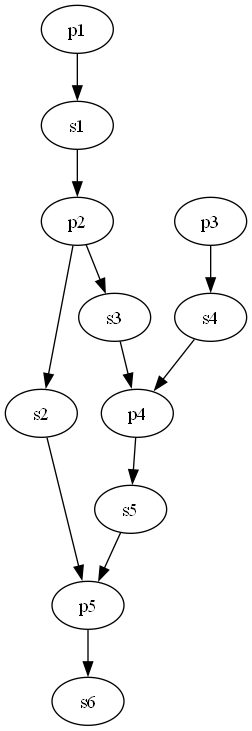}
\caption{\label{probleempropagatie}Een voorbeeld van een abstracte probleempropagatie. Hierin stellen de p's problemen voor en de s-en oplossingen (\emph{solutions}).}
\end{figure}

Probleempropagaties kan men grafisch en/of in tekst representeren. In Figuur~\ref{probleempropagatie} is een abstracte probleempropagatie ge\"illustreerd, met de p's als problemen en de s-en als oplossingen (\emph{solutions}).
 
\begin{figure}
\centering
\includegraphics[width=11cm]{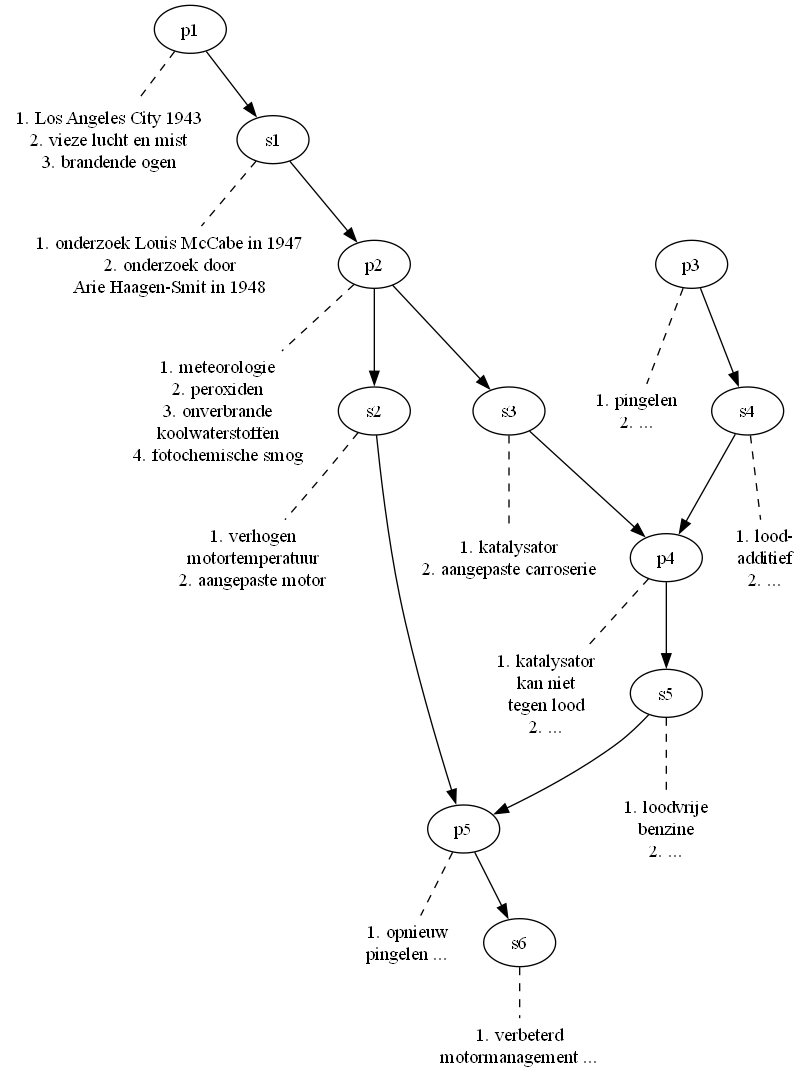}
\caption{\label{contextprobleempropagatie}Een voorbeeld van een probleempropagatie en een suggestie van enige context uit de Los Angeles Smog casus vanuit het gezichtspunt van een fictieve observator.}
\end{figure}

Het hoeft niet zo abstract. De representatie van een probleempropagatie kent geen formele eisen \cite[p.~7]{ebac00898-LosAngelesSmog-probleemtypen}. Een probleempropagatie kan meer of minder worden uitgewerkt met de details van een concreet oplossingsproces. In Figuur~\ref{contextprobleempropagatie} is de eerdere propagatie voorzien van concrete informatie ontleend aan de Los Angeles Smog-casus \cite{ebac00899-LosAngelesSmog-probleemposities, Jacobs-Kelly-SmogTown2008, Dewey-Dont-breathe-the-air}. Om de lezer te helpen de annotaties in Figuur~\ref{contextprobleempropagatie} te plaatsen vertellen we kort waar de Los Angeles Smog-casus over gaat.

Rond 1943 nam de luchtvervuiling in de industriestad Los Angeles City en in de steden eromheen ernstige vormen aan. Het werd tot een thema dat decennia het lokale politieke debat zou beheersen, en dat nationaal en internationaal grote invloed had op technische ontwikkelingen en het milieudenken. Aanvankelijk dacht men te maken te hebben met smog veroorzaakt door het gebruik van steenkool voor verwarming, energieopwekking en industrí\"ele processen. Dergelijke smog kende men bijvoorbeeld uit St.\ Louis, Pittsburgh, London. Maar na jaren onderzoek bleek de hoofdoorzaak in Los Angeles van geheel andere aard: onvolledige verbranding van olieproducten in automotoren.  De vluchtige koolwaterstoffen waaiden zomers niet weg maar bleven boven de steden hangen door meteorologische omstandigheden en door de bergen die Los Angeles County omringen. De overvloedig schijnende zon had nu vrij spel om de koolwaterstoffen te laten reageren tot allerlei nare stofjes: fotochemische smog. Het kostte vele jaren dit probleem onder controle te krijgen om heel veel redenen. Bijvoorbeeld, het autoverslaafde en luid klagende publiek wilde niet geloven dat ze zelf deels de oorzaak waren. Schoorvoetend gingen de olie- en de auto-industrie werken aan aanpassing van hun producten om tot betere verbranding te komen. Een voorbeeld hiervan is de toepassing van een katalysator in de uitlaat (de voorloper van onze huidige driewegkatalysator). Deze bleek echter loodvrije benzine te vereisen. Maar dat lood zat in benzine om vroegtijdige ontbranding in de cilinders te voorkomen, het zogenaamde pingelen. Zonder lood moest pingelen op een andere manier voorkomen worden, et cetera. De Los Angeles Smog-casus is rijk aan probleempropagaties.

Net als voor probleemposities kunnen ook probleempropagaties meer of minder getrouw of met meer of minder detail verwoord worden. Zij het, volgens de AD, nooit compleet. Taal schiet altijd te kort. Bijvoorbeeld omdat interne probleemposities op een onnavolgbare manier overlopen in onbereikbare diepere natuurlijke probleemposities. Dat neemt niet weg dat men probleempropagaties in groot detail kan uitwerken door complete probleemposities op te nemen en de oplossingsstappen veel verder uiteen te rafelen.

\subsection{De Innovatie-illusie}
Mensen zoeken naar stabiliteit, houden van voorspelbaarheid en werken graag volgens vaste patronen. Standaardisering en het werken volgens procedures zijn daar het resultaat van. De Innovatie-illusie stelt dat mensen regelmatig neigen naar voorspelbaarheid bij het oplossen van problemen, terwijl ze denken juist innovatief bezig te zijn. Die neiging tot voorspelbaarheid en herhaling valt pas op als men meer naar het interactiepatroon kijkt en abstraheert van details die geen oorzakelijke betekenis hebben, van het domein, de spelers, en van tijd- en ruimteschalen.

De Innovatie-illusie postuleert probleemtypen als een eindige verzameling van speler, schaal-, tijd- en domeinonafhankelijke abstracte onwenselijke situaties en abstracte oplossingen ervoor. De set van oplossingen wordt ook verondersteld eindig te zijn. De set wordt het repertoire van het probleemtype genoemd, we spreken soms ook van oplossingstypen. 

Een voorbeeld is het probleemtype congestie. In Nederland is zowel sprake van een fileprobleem op autosnelwegen en van overbelasting van het landelijke elektriciteitsnetwerk. Beide problemen zijn gevallen van het probleemtype congestie. De oplossingen hierbij laten veel parallellen zien. Twee voorbeelden: extra rijstroken in het ene domein komt overeen met dikkere kabels in het andere domein; de spits vermijden is vergelijkbaar met `een verzoek tot beperkte stroomgebruik tussen 9 en 16 uur door huishoudens'. Het zijn analoge situaties en ingrepen. In termen van het probleemtype is sprake van een aggregatie (weg/kabel) met een functie (de rol transport faciliteren) die ten opzichte van andere aggregaties (weggebruikers/stroomgebruikers) haar rol niet voldoende waar kan maken.
 
De Innovatie-illusie chargeert vermoedelijk, maar waar de grens tussen herhaling en innovatie zit, is vooralsnog onduidelijk. Voor ons betoog is de aanname van belang dat in de meeste probleemstellingen en de erop volgende oplossingsprocessen de bekende weg gevolgd wordt.

\subsection{Probleempropagaties en hun patronen}\label{patinprop}
Een belangrijke vraag is: zitten er patronen in probleempropagaties? De hypothese, uit een vorige sectie, dat beschrijvingen van probleemposities uit een overzichtelijk aantal beweringen bestaan lijkt dat al te suggereren. De Innovatie-illusie die hierboven werd toegelicht versterkt dat vermoeden aanzienlijk. 

De Innovatie-illusie beweert dat het aantal probleemtypen eindig is en ook bijbehorende repertoires een eindige omvang hebben. Evident zit een probleempropagatie die geabstraheerd is tot op probleemtype-niveau dan vol met herhaling en patronen. Want, stel dat er $n$ probleemtypen zijn dan moet het volgende probleem in een probleempropagatie die $n$ probleemstappen diep gaat van een al eerder voorgekomen type zijn. Bijvoorbeeld, het onderhoud van een complex apparaat leidt tot de inzet van een bepaalde expert en die moet ook haar apparatuur weer onderhouden.

Uit het voorgaande volgt dat men zich bij probleemposities en probleempropagaties rond concrete problemen ook abstractere varianten kan voorstellen. Stapsgewijs kan men naar voorstellingen op typeniveau toewerken, of nog abstracter, typen van typen. En vandaar kan men natuurlijk ook weer terug, namelijk detailleren naar concretere gevallen. 

\section{Cognitieve taalkunde}\label{taalcogni}
We schetsen hier de voor ons relevante hoofdlijnen uit de Cognitieve taalkunde zoals met name ontwikkeld door de taalkundige George Lakoff en de filosoof Mark Johnson.

Iedereen kent metaforen en analogie\"en als stijlfiguren. Beide duiden op het geval waarin we een ding, gedrag of situatie begrijpen en ervaren in termen van een \'ander ding, gedrag of situatie. Cognitieve taalkunde geeft de voorkeur aan de term metafoor. De metafoor heeft wellicht een meer dichterlijke connotatie waarin de woordvondst wat zwaarder telt dan in de analogie. Die is dan wellicht meer beschrijvend en de gekozen woorden zijn instrumenteel, het kan ook anders gezegd. Hoe ook, de metafoor of analogie komt soms mooi van pas om iets duidelijk te maken.

Soms? Nee, het tegendeel blijkt waar. Vrijwel elke zin bevat een metafoor omdat wij niet kunnen begrijpen en verwoorden zonder te vergelijken. Dit inzicht brak door met de publicatie van \emph{Metaphors we live by} van Lakoff en Johnson in 1979. Deze theorie is voor ons betoog van groot belang. Teksten zijn rijk aan gelaagde betekenis en geven een indruk hoe we denken. Het onderzoeksgebied dat Lakoff en Johnson mede vormgaven wordt tegenwoordig wel Cognitieve taalkunde genoemd \cite{Thewaywethink,Feldman-From-Molecule-Metaphor,ebac01035-s11097-025-10055-w}. We bespreken de centrale en voor ons belangrijkste elementen.

De Cognitieve taalkunde meent dat menselijke mentale activiteit -- denken, betekenis geven, gemotiveerd handelen -- diepgaand gevormd en samengesteld wordt door lichamelijke ervaringen, bewegingen, sensorische prikkeling, emoties, gevoelens en acties. Er is een permanente wisselwerking tussen de omgeving en het lichaam-brein.

De concepten, hun eigenschappen en relaties, waar we zo vertrouwd mee zijn in ons dagelijks leven, zijn geen objectieve gegevens, maar leren we stap voor stap. Dat begint al als baby met onze lichamelijke ervaringen, ervaren correlaties genoemd: als dit gebeurt dan gebeurt ook dat. De directe lichamelijke ervaring leidt tot de zogenaamde primaire concepten of metaforen. Enkele voorbeelden zijn: \emph{Knowing Is Seeing}, \emph{More Is Up} en \emph{Speech Act Is Physical Force}. We lichten ze toe.

\emph{Knowing} wordt begrepen in termen van \emph{Seeing}. Je leert als kind dat beter kijken helpt bij beter begrijpen. Je leert dat al voor je er woorden aan kan geven. Ben je eenmaal taal machtig dan begrijp je uitdrukkingen als: `Ik zie wat je bedoelt', `Kan je dit aspect nog verder toelichten' of `De conclusie vind ik nog niet helder'. \emph{More Is Up}, leer je bijvoorbeeld door te spelen met water. Je ziet dat het peil in je beker stijgt als je meer water erin doet. Het primaire concept \emph{Speech Act Is Physical Force} leer je als je ouders jouw lichaam vastpakken en besturen, bijvoorbeeld je hand, en tegelijk een verbale instructie geven: `Hand weg! Heet!'

De metafoor is het mentale gereedschap om betekenis te geven aan nieuwe ervaringen en situaties op basis van concepten, eventueel primaire concepten, die je al kent. Degene die een nieuw concept voor het eerst bedenkt of leert van een ander zal zich eerst bewust zijn van de metafoor, maar zodra het concept meer vertrouwd raakt gaat het zijn eigen mentale leven leiden en ben je je niet meer bewust van de metafoor \cite{ebac00966-NeuralTheoryofMetaphor}. Hogere orde concepten, dus boven het primaire niveau, worden conceptuele metaforen genoemd. Tot concept verheven metaforen. Hoe meer concepten men tot zijn beschikking heeft, hoe meer situaties betekenis kunnen krijgen en hoe meer mentale speelruimte er is.

Conceptvorming en conceptuele metaforen zijn niet de totale verklaring voor abstract denken, maar een belangrijke mechanisme ervan en belangrijke elementen erin \cite{Johnson-embodied-2017,ebac01008-Holyoak02072024}. De metafoor \emph{Categories Are Containers} illustreert de relatie tussen primaire en abstractere concepten. Honderden keren per dag ervaren we het concept \emph{container}, het is een terugkerende eigenschap van veel dingen waar we dagelijks, van jongs af, mee omgaan: tassen, flessen met appelsap, schoenen, huizen, broekzakken. Je kan ergens iets in stoppen. Maar soms past dat wel en soms niet. Zo leren we de ruimtelijke logica van gevallen binnen de categorie container. Als ik mijn sleutels in mijn hand heb, en ik heb mijn hand in mijn broekzak, dan zijn mijn sleutels in mijn broekzak. Als men container nu metaforisch verheft tot categorie of concept dan kunnen abstracte concepten elkaar ook bevatten, dan wordt het verzamelingenleer. Als een merel een vogel is en vogels zijn dieren dan is een merel ook een dier \cite[p.~52-53]{Johnson-embodied-2017}. 

De Cognitieve taalkunde maakt aannemelijk dat al onze concepten de vorm hebben van metaforen. Ook de meest abstracte conceptualisaties in de exacte wetenschappen krijgen betekenis via het inschakelen van heel gewone ``sensory-motor parts of the brain''. De betekenis van concepten is dus gefundeerd in onze ``sensory-motor and affective experience'' \cite[p.~52]{Johnson-embodied-2017,Hofstadter-Sander-Surfaces-Essences}.

\section{Neuropsychologie}\label{neuro-psych-en-psych}
Het werk van Lakoff en Johnson en anderen, van eind jaren 70 van de vorige eeuw, gaat tientallen jaren vooraf aan een grote sprong in het empirisch onderzoek in de neurowetenschappen. De resultaten daarvan bevestigen het door het Cognitieve taalkunde geschetste beeld en vullen het verder in \cite{Feldman-From-Molecule-Metaphor}. Om dat te illustreren verwijst Johnson onder andere naar werk van de Amerikaanse psycholoog D.M.\ Tucker, die stelde: ``there are no brain parts for disembodied cognition'' \cite[p.~51]{Johnson-embodied-2017}. We baseren ons hier met name op werk van de Britse onderzoekers Andy Clark en Anil Seth en hun visie op de Predictive processing theory \cite{AndyClark-ExperienceMachine,AnilSeth-BeingYou}.

Predictive processing theory stelt: wat wij ervaren als waarneming is de beste \emph{voorspelling} van het brein van de situatie waarin we verkeren. Deze theorie gaat er vanuit dat ons denkproces voortdurend hypothesen genereert over de situaties waarin het lichaam en brein zich bevinden en deze vervolgens toetst aan de sensorische data. Dat is data over de buitenwereld, maar ook over de interne toestand en houding van het lichaam. Dus het betreft sensorische data van bekende zintuigen zoals ogen, oren, reuk, smaak, tast, maar ook lichaamstemperatuur, hartslag, de ori\"entatie van lichaamsdelen. 

Integratie van al dit gehypothetiseer levert een totaal en sluitend beeld op. Althans, zo ervaren we dat. We denken dat we de wereld echt waarnemen. Maar het is slechts een projectie. In navolging van anderen spreekt Anil Seth over het voorspellingsresultaat zoals we die bewust ervaren als een \emph{controlled hallucination}. 

Een lang geaccepteerde, en intu\"itief invoelbare, theorie over waarneming, denken en handelen stelt dat de ogen, en andere bekende zintuigen, aan de basis staan van een proces dat onze mentaal ervaren scherpe beeld van de wereld genereert en actualiseert. Daarna kan het denken aan de slag en kan gehandeld worden. De Predictive processing theory draait de volgorde in het proces van waarnemen, denken en handelen om. Hier begint de waarneming bij de voorspelling. Conflicten tussen de voorspelling en de sensorische informatie leidt tot aanpassing van de voorspelling, denken erover, en/of een lichamelijk actie om de situatie aan te passen aan de voorspelling, en/of om de situatie nader te onderzoeken. De Cognitieve taalkunde maakt inzichtelijk hoe de structuur van ons voorspellingsproces is door de nadruk op het metaforische aspect van ons denken.

Beide theorie\"en kunnen gezien worden als onderdeel van een naturalistische stroming van cognitieonderzoek. De reden om dit zijpad even te bewandelen is dat ook het werk aan kunstmatige neurale netwerken als LLMs typisch deel ervan uitmaakt. De voorlopers van de Predictive processing theory hebben een lange geschiedenis, die laten we hier verder onbesproken \cite{NewellSimon-HumanProblemSolving,MaturanaVarela1998,epub00913-Margaret-Boden-Mind-as-Machine,ebac01035-s11097-025-10055-w, ebac01034-Nikolova2025-NIKPPT-2}.

\section{Denken, denkstijlen en gesprekspartners}\label{denkstijlen}
De American Psychological Association definieert denken als volgt \cite{dictionary-APA-org}:
\begin{quote}Thinking. Cognitive behavior in which ideas, images, mental representations, or other hypothetical elements of thought are experienced or manipulated. In this sense, thinking includes imagining, remembering, problem solving, daydreaming, free association, concept formation, and many other processes. Thinking may be said to have two defining characteristics: (a) It is covert -- that is, it is not directly observable but must be inferred from actions or self-reports; and (b) it is symbolic -- that is, it seems to involve operations on mental symbols or representations, the nature of which remains obscure and controversial.\end{quote}

\noindent Dit is heel algemeen gesteld. Het psychologisch onderzoek naar het vermeende rationeel handelen van de homo economicus, waaraan de namen van de Amerikaans-Isra\"elische psychologen Daniel Kahneman en Amos Tversky nadrukkelijk verbonden zijn, heeft denkstijlen nader gekarakteriseerd \cite{Kahneman-Fast-slow-2012,DeRegtDooremalen2015-EN}. Zij stellen dat denkactiviteit onder te verdelen is in ruwweg twee stijlen. Die stijlen worden aangeduid met Systeem~1 en Systeem~2. Systeem~1 ``operates automatically and quickly, with little or no effort and no sense of voluntary control.'' Systeem~2 ``allocates attention to the effortful mental activities that demand it, including complex computation. The operations of System-2 are often associated with the subjective experience of agency, choice and concentration.'' Systeem~2 kost veel meer tijd en moeite dan Systeem~1. Kahneman c.s.\  stellen vast dat Systeem~1 veruit de meest gebruikte denkstijl is voor ons dagelijks handelen. De indeling in Systeem~1 en Systeem~2 is modelmatig praktisch, in de praktijk gebruiken mensen beide stijlen vaak door elkaar en lopen ze in elkaar over.

Alhoewel de American Psychological Association (APA) niet refereert aan Systeem~1 en Systeem~2 menen we dat aanvullende definities van de APA prima verdere invulling geven aan Systeem~2. 

\begin{quote}Critical thinking. A form of directed, problem-focused thinking in which the individual tests ideas or possible solutions for errors or drawbacks. It is essential to such activities as examining the validity of a hypothesis or interpreting the meaning of research results.\end{quote}

\begin{quote}Creative thinking. The mental processes leading to a new invention, solution, or synthesis in any area. A creative solution may use preexisting elements (e.g., objects, ideas) but creates a new relationship between them. Products of creative thinking include, for example, new machines, social ideas, scientific theories, and artistic works.\end{quote}

\noindent Verder onderscheid de APA ook nog \emph{divergent thinking}, \emph{lateral thinking} en \emph{convergent thinking} als varianten van kritisch of creatief denken waarin de nadruk ligt op respectievelijk nieuwe denkstrategie\"en, systematische heroverweging van aannamen en systematische controle van de logica van denkstappen, denken over het eigen denken. Wij nemen de vrijheid al deze denkwijzen, van \emph{critical} tot en met \emph{convergent} op te vatten als kenmerkend voor Systeem~2-denken. We noemen dat in het vervolg analytisch denken. 

In de literatuur over LLMs kwamen wij de indeling tegen in \emph{reasoning behaviour} en \emph{reasoning performance} \cite{ebac00968-2404.01869v2}. Dat is een opmerking waard. Onze indeling richt zich duidelijk op behaviour en niet op performance of \emph{accuracy} of \emph{efficiency}. Die aspecten zijn relevant in datacentra die queries snel willen afhandelen, niet voor een discussie over principes. Een ander aspect dat we hier willen benoemen is de evolutionaire basis van menselijk denken \cite{ebac01045-rstb20120111}. We menen dat mensen slechts een door de aardse situatie bepaalde vorm van cognitie vertegenwoordigen \cite[p.~12]{Sebastiaan-Mathot-wereld-vol-denkers-EN}. 

\section{Metaforische probleempropagaties}\label{kernbetoog}
De voorgaande Secties~\ref{AIbots} tot en met \ref{denkstijlen} vormen het basismateriaal voor het eigenlijke betoog dat hieronder begint. Zoals in de inleiding al aangegeven is, bestaat het betoog uit vier delen. Het eerste deel volgt in deze sectie en de andere drie in Sectie~\ref{metaforische-probprop-tekst}, \ref{denkruimte-chatbots} en \ref{watvoorp}.  

Kort samengevat gaan de vier delen over: 1. een model van de menselijke denkruimte; 2. de verbalisering van fragmenten uit die denkruimte in tekst, en de eigenschappen van die tekst; 3. de training van LLMs op tekst en wat dat voor kunstmatige denkruimte dat oplevert; 4.  wat een chatbot wel en niet kan in zijn kunstmatige denkruimte en wat dat verder zou kunnen betekenen.

In dit betoog is de centrale vraag: wat voor gesprekspartner is een chatbot? Naar aanleiding van de vorige sectie over denkstijlen kunnen we die vraag nu verfijnen tot de vraag: is de chatbot eerder een Systeem~1- of een Systeem~2-geprekspartner? Laten we conversaties opvatten als probleem-gemotiveerd, zoals al eerder gesteld. De gesprekspartners willen een probleem oplossen of er een visie op ontwikkelen.

\subsection{Probleemtype en repertoire zijn metaforen}\label{ptenprmeta}
\noindent Neem het conceptuele systeem uit de Cognitieve taalkunde in gedachten. Lakoff en Johnson stellen \cite[p.~3]{Lakoff-Johnson-By-Metaphores}:
\begin{quote} In most of the little things we do every day, we simply think and act more or less automatically along certain lines. Just what these lines are is by no means obvious. One way to find out is by looking at language. Since communication is based on the same conceptual system that we use in thinking and acting, language is an important source of evidence for what that system is like.\end{quote}
Wij vatten hier language op als taal-in-actie, als concrete taaluitingen over concrete zaken. 
Waar dient dat ``thinking and acting'' voor?

Mark Johnson vult dat onder andere in door te verwijzen naar de filosofie van de Amerikaanse pragmatici William James en John Dewey. Johnson vat dit samen in vijf hoofdpunten, nummer 3 en 4 zijn de kersen die we hier graag plukken \cite[p.~69]{Johnson-embodied-2017}: \begin{quote} \begin{itemize}\item Embodied cognition is problem centered, and it operates relative to the needs, interests, and values of organisms. \item Embodied cognition is not concerned with finding some alledgedly perfect solution to a problem, but one that works well enough relative to the current situation.\end{itemize}\end{quote}

\noindent De Amerikaanse filosoof Donald A.\ Sch\"on gebruikt voor ons betoog een paar cruciale termen als hij schrijft over sociaal beleid en generatieve metaforen in een bundel \emph{Metaphor and Thought} \cite[p.~138]{Metaphor-and-thought-2ndedition}: \begin{quote}When we examine the problem-setting stories told by the analysts and practitioners of social policy, it becomes apparent that the framing of problems often depends upon metaphors underlying the stories which generate problem setting and set the directions of problem solving.\end{quote} 

\noindent Wij interpreteren \emph{framing of problems} en \emph{metaphors underlying the stories} in Sch\"ons citaat respectievelijk als probleemtypen respectievelijk probleemposities. Wij menen dan ook dat het door Sch\"on beschreven proces ook buiten het domein van sociaal beleid geldt. Voor deze overeenkomst met de Cognitieve taalkunde is nog een aanwijzing. 

Probleemtype en oplossingen worden in de AD opgevat als rol- en interactiepatronen tussen partijen. Dat beeld sluit naar ons idee aan op de Cognitieve taalkunde. Die zoekt de aard van het concept bij een object niet in termen van objectieve aspecten en eigenschappen. Het hebben van een concept van een object is ``to be able to simulate the kinds of perceptual, motor, and affective interactions you typically have with that kind of object'' \cite[p.~24]{Johnson-embodied-2017}. En deze interacties leveren via ervaringscorrelaties   primaire en conceptuele metaforen op (zie Sectie~\ref{taalcogni}). Dus de overeenkomst zit in interactie-ervaringen. Denk even terug aan de metafoor \emph{Knowing Is Seeing} (voorbeeld uit Sectie~\ref{taalcogni}). Denken is geen zien. Maar iets goed bekijken gaat vaak samen met beter begrip. Over dergelijke relaties spreekt de AD in termen van rollen en rolpatronen.  

\subsection{De notie metaforische probleempropagatie}\label{defmpp}
De notie \emph{metaforische probleempropagatie} is een model van de menselijke denk- en ervaringsruimte die bestaat uit de combinatie van de probleembegrippen van de Aggregatie-dynamica, het conceptuele systeem van de Cognitieve taalkunde en de voorspellingsdynamiek uit de Predictive processing theory. 

Deze menselijke denk- en ervaringsruimte is niet nieuw, die ervaart de lezer zelf ook ieder moment. Wat nieuw is, wellicht, is dat de combinatie van deze drie theoretische invalshoeken ons de mogelijkheid geeft om te kunnen reflecteren op probleem-oplossing-gedreven conversaties met chatbots. Deze combinatie is denken wij, om te beginnen, het best uit te leggen aan de hand van een voorbeeld.

Probleempositie 1: `Een automobilist rijdt op de snelweg. De auto's die ervoor rijden remmen plots hard af. Probleem: hoe te reageren. Een oplossing is remmen.' Probleempositie 2: `De automobilist is aan het afremmen. Het subprobleem is nu: adequaat remmen. De oplossing: zo remmen dat de automobilist niet op de voorganger botst, en de auto's achterop komen ook gelegenheid hebben om tot stilstand te komen.' Probleempositie 3: `De automobilist staat stil. De remoperatie is geslaagd. Probleem: de reis- en de dagplanning van de automobilist klopt niet meer.'

Het voorafgaande is een eenvoudig voorbeeld van een probleempropagatie. Laten we de gebeurtenissen persoonlijker maken.

Probleempositie 1: `Ik rij over de snelweg A9 om 8 uur 's morgens, plotseling zie ik de remlichten van de auto voor mij pompend oplichten. Ik wil in leven blijven. Een oplossing is remmen.' Probleempositie 2: `Ik rem af. Ik wil nog steeds in leven blijven, maar het subprobleem is nu: adequaat remmen. De oplossing is zo af te remmen dat ik niet op mijn voorganger bots, en de auto die achter mij rijdt, niet op mij. ' Probleempositie 3: `Ik sta stil. Ik besef dat ik nog leef. Maar mijn afspraak van 10 uur ga ik zeker niet halen, met als mogelijk gevolg dat de anderen in de vergadering iets beslissen waar ik tegen ben!' 

We zien probleem-oplossingrelaties, we zien voorspellingen en metaforische relaties in dit voorbeeld. Voordat ik moest remmen leefde ik in de veronderstelling: `iedereen rijdt netjes en houdt afstand, business as usual, ik ben over 15 minuten op mijn afspraak, mooi op tijd.' De remlichten van de auto voor me, die oplichten, doorkruisen dat beeld. Meerdere voorspellingen dienen zich aan. We noemen er twee: `iemand tikt per ongeluk even op zijn rempedaal, als de lichten meteen doven en de auto doorrijdt kan ik dit rode lichtsignaal negeren'; `dit is echt remmen, maar hoe hard wordt er dan geremd?' De laatste voorspelling vraagt om beter kijken om beter te begrijpen wat er gebeurt. Deze voorspellingen zijn onderdeel van ons conceptuele systeem, die wij doorgaans als volkomen vanzelfsprekend ervaren. In deze situatie zegt ons conceptuele systeem \emph{Pompend remmen Is Direct gevaar} en dat vraagt om directe actie: remmen. Dit voorbeeld illustreert hoe de verschillende elementen van een metaforische probleempropagatie op elkaar inwerken, betekenis genereren en handelingen voorbereiden en activeren. Nemen we al deze emoties, voorspellingen, metaforische relaties mee -- wat we hier maar in beperkte mate doen -- dan ontstaat een totale beleving een voorbeeld van een metaforische probleempropagatie.

De metaforische probleempropagatie is zondermeer een complexe notie. Wij vinden het bruikbaar ons de mentale architectuur van de metaforische probleempropagatie als volgt voor te stellen. De conceptuele stapeling, van primaire concepten naar conceptuele metaforen, is hier een `verticale' dimensie. In een `horizontale' dimensie ontvouwen de propagaties zich. Om van een tijd-as te spreken is niet passend. Dat suggereert een precisie die er niet is, en het zou ook de mentale vrijheid beperken, bijvoorbeeld voor andere -- eventueel tegenstrijdige -- ordeningen van gebeurtenissen, of voor projecties in de toekomst of het verleden. Tot slot voegt de Predictive processing theory en derde dimensie toe, een voorspellings-dimensie.  In deze dimensie zijn verschillende voorspellingen parallel aan het werk, op verschillende abstractieniveaus, met andere prioritering van de rol van het conceptuele systeem, gebruikmakend van andere woorden. 

Denken over actuele situaties, hypothesen opstellen en toetsen, het overdenken van ervaringen en het onthouden ervan valt hiermee voor te stellen als het bewerken van de plastische en flexibele metaforische probleempropagaties. Een metaforische probleempropagatie is niet objectief, deels onbewust en bovendien speelt het toeval ook een rol. Sommige waarnemingen dringen door, anderen worden gemist. Sommige worden als probleem ervaren andere niet.

Johnson beschrijft een ``representatietheory'', onder verwijzing naar werk van J.~Fodor, als ``the idea that human thought consists of a series of functional computational operations on language-like symbols ``in the mind'' that can be used to represent external states.'' Alhoewel we er met woorden over spreken en woorden vaak een gevolg zijn, is onze notie van de metaforische probleempropagatie niet een symbolische representatie in deze betekenis. De metaforische probleempropagatie is onze invulling van wat Johnson met verwijzing naar Lakoff en Gallese beschrijft als een  ``interactionist, multimodal, simulation theory of meaning and thought'' \cite[p.~145]{Johnson-embodied-2017}. Wij leggen hierin de nadruk op probleem-oplossingrelaties.\\

\noindent We diepen de raakvlakken tussen de drie theoretische invalshoeken nog wat verder uit. Zoals al gezegd kan men zich het conceptuele systeem voorstellen als een hi\"erarchie met onderop primaire metaforen en verder naar boven meer abstracte concepten en soms ook complexe en zelfs inconsistente combinaties \cite[p.~53]{Johnson-embodied-2017}. De aan de AD ontleende concepten lopen gewoon mee in deze hi\"erarchie, met bijvoorbeeld praktische problemen die aansluiten op de primaire concepten onderop en abstracte problemen hogerop tot aan de meest abstracte categorie, namelijk die van het `ongespecificeerde probleem'; ``Houston, we have a problem'' \cite{houston-we-have-a-problem}.

Een vergelijkbare hi\"erarchie, van concreet naar abstract, postuleert de AD ook voor probleemposities, voor problemen en voor oplossingen (zie de laatste alinea van Sectie~\ref{patinprop}). De relaties die men mentaal tot leven wekt als men een probleempositie overdenkt, worden ervaren in metaforen. We stellen ons dan ook voor dat het mogelijk is om het denkproces op verschillende niveaus van abstractie te starten en verder te voeren. En ook dat het dan parallel op andere niveaus kan lopen. Dus het denken over `de buren Jan en Piet die ruzie hebben over de heg tussen hun tuinen' verloopt bij wijze van spreken ook op een abstracter niveau als $P_1$, $P_2$ en $h$ in relatie $r$ en wellicht ook op concretere of nog abstracte niveaus. Of, verloopt, bijvoorbeeld, parallel in een andere onthouden propagatie, over een andere ruzie, via associaties met bepaalde aspecten uit de context of de karaktertrekken van Piet. De vrijheid om tussen deze niveaus te schakelen en om nieuwe niveaus te introduceren zijn onderdeel van ons conceptuele systeem. Ook kan men zich goed voorstellen dat het mogelijk is terug te denken en vooruit te denken in de metaforische probleempropagatie. Vooruit en achteruit bestaan bij de gratie van geheugen en ervaring, er is geen harde tijdas. 

De probleembegrippen uit de AD en het conceptuele systeem sluiten niet alleen mooi aan, de combinatie doet de toepassingsmogelijkheden toenemen. Het conceptuele systeem van de Cognitieve taalkunde voorziet al in een dimensie om actie-elementen te beschrijven. \emph{Image schemas} zijn ``recurring sensory-motor patterns'' die ``afford us possibilities for meaningful interaction with our surroundings, both physical and social'' \cite[p.~21 \& 100]{Johnson-embodied-2017}. Een uitbreiding hiervan zijn \emph{X-schemas} \cite{ebac01054-ebacBailey,ebac00966-NeuralTheoryofMetaphor} die men als samengestelde processen van sensory-motor patterns zou kunnen opvatten. De sensory-motor patterns zijn nog steeds een actueel thema \cite{ebac01053-role-emotion-in-acquisition-of-verb-meaning}.

Maar deze koppelingen van gewaarwording en handeling gaan over relatief primaire interacties, en korte ketens. De Aggregatie-dynamica modelleert op een andere schaal: herinneringen, geschiedenissen, plannen, scenario's, belangen en hypothesen over wat tegenspelers denken (Theory of Mind). Op die schaal gaat het over lange verknoopte reeksen van gebeurtenissen of geprojecteerde acties. Hiermee wordt de actie-dimensie van het conceptuele systeem enorm vergroot en krijgt ook relevantie in veel meer situaties, wat via de metaforen als denkinstrumenten volkomen voorstelbaar is, menen wij. De sensory-motor patterns worden, zeg maar, metaforisch opgerekt tot probleemsituaties tussen, bijvoorbeeld, supermachten. Deze grotere schaal is niet direct wezensvreemd aan de Cognitieve taalkunde, dat blijkt duidelijk uit het eerdere citaat van Sch\"on dat handelt over ``social policy''. Daar komt nogal wat bij kijken. Maar het lijkt niet de schaal waarop men bij voorkeur analyseert.  

Maar hoe komen die mentale voorstellingen tot leven? In zijn boek uit 2017 heeft Johnson nadrukkelijk aandacht voor het voorspellende brein \cite[p.~142]{Johnson-embodied-2017}, maar de Predictive processing theory draait daar helemaal om \cite{AndyClark-ExperienceMachine,AnilSeth-BeingYou}. Deze theorie beschrijft hoe het brein voortdurend voorspellingen genereert over de situatie waarin we verkeren en hoe die zich zal ontwikkelingen. Maar net zo goed gaan die voorspellingen over het verleden, bijvoorbeeld over de vraag `hoe heeft dit mis kunnen gaan'. De Predictive processing theory verklaart ook hoe uit de concurrerende en onderling vaak inconsistente voorspellingen toch een totaal en gevoelsmatig kloppend beeld als Controlled Hallucination gegenereerd wordt. 

Wij hebben de indruk dat deze voorspellingen door de Predictive Processing theory niet heel concreet worden ingevuld. Dat lijkt begrijpelijk vanuit de hoog technische empirische setting waarbinnen deze theorie zich ontwikkelt waarbij men zich concentreert op correlaties tussen sensorische prikkeling, neurale effecten, gewaarwordingen en handelingen. Maar wij permitteren ons hier het concept `voorspelling' op te rekken en de metaforische probleempropagaties als structuur ervoor de postuleren. 

Door de Predictive processing theory worden de noties probleem en oplossing verdiept.
Het ligt voor de hand dat we de problemen identificeren met de mismatch tussen de voorspelde en de waargenomen wereld. Een probleem is dan een \emph{prediction error}. Een oplossing identificeren we dan vanzelfsprekend met de aanpassingen van de voorspelling of een actie. Een oplossing of oplossingsstap brengt de wereld in overeenstemming met de voorspelling of brengt voorspelling en wereld dichter bij elkaar. 

Wij denken dat de notie metaforische probleempropagatie een interessante aanvulling kan zijn voor de Cognitieve taalkunde en de Predictive processing theory. Voor de Aggregatie-dynamica, als poging tot een totale systeemtheorie, is de metaforische probleempropagatie een belangrijke uitbreiding van de toepassingsmogelijkheden. Bovendien is de connectie met Cognitieve taalkunde en de Predictive processing theory een versteviging van de theoretische basis.

Net zoals de probleempropagatie uit Sectie~\ref{probprog} vormen de metaforische probleempropagaties een categorie van mentale structuren. Een metaforische probleempropagatie kan van zeer grote omvang zijn, zoals alles wat iemand weet en ervaren heeft, maar ook een tijdelijke gedachtenflard betreffen die snel vergeten raakt, of een fragment rond een specifiek probleem. In ons betoog wordt uit de context duidelijk welke reikwijdte we bedoelen, of het dus een totale denkruimte betreft, een cluster van een aantal gerelateerde kwesties, een fragment, of een gedachte.

\section{Metaforische probleempropagaties en teksten}\label{metaforische-probprop-tekst}
In de Sectie~\ref{probpos} en \ref{probprog} werd vooruitgelopen op de omzetting van de mentale wereld van probleemposities en probleempropagaties in tekstuele vorm of grafische vorm. Hieronder postuleren we een proces dat die overgang beschrijft en we vragen ons af wat de eigenschappen zijn van resulterende tekst. 

Stel nu dat iemand een gedachte heeft en die met een ander wil delen, of een verhaal wil vertellen aan een ander. Wat gebeurt er als die gedachte of dat verhaal verteld wordt? Naar aanleiding van wat we hierboven postuleerden, stellen we ons die gedachte of dat verhaal voor als een onthouden of ter plekke ontrollende metaforische probleempropagatie. Als we de metaforische probleempropagatie uit Figuur~\ref{contextprobleempropagatie}, die onthouden gebeurtenissen representeert, in tekst omzetten dan zou dat bijvoorbeeld iets kunnen opleveren als ge\"illustreerd in Figuur~\ref{Verbaliseren-CPP}.

\begin{figure}
\textbf{De situatie en het probleem betroffen}\\
`Los Angeles City 1943; vieze lucht en mist; brandende ogen'\\
\textbf{De eerste stappen richting een oplossing waren}\\
`onderzoek Louis McCabe in 1947; onderzoek door Arie Haagen-Smit in 1948'\\ 
\textbf{Dit bracht de oorzaken in beeld}\\
`meteorologie; peroxiden; onverbrande koolwaterstoffen; fotochemische smog'\\
\textbf{Op twee manieren werd gezocht naar oplossingen voor deze problemen}\\  
`verhogen motortemperatuur; aangepaste motor'\\
\textbf{en}\\
`katalysator; aangepaste carrosserie'\\
\textbf{Maar het bleek dat}\\ 
`katalysator kan niet tegen lood; ...'\\ 
\textbf{deze vroeg om}\\
`loodvrije benzine; ...'\\ 
\textbf{Dat bracht echter een oud probleem terug}\\ 
`pingelen; ...'\\ 
\textbf{Het}\\
`loodadditief; ...'\\
\textbf{voorkwam dat, nu was dat probleem weer actueel}\\
`opnieuw pingelen; ...'\\ 
\textbf{een oplossing hiervoor werd gevonden in}\\
`verbeterd motormanagement; ...'\\
\textbf{Dat loste dat pingelen weer op, nu anders.}
\caption{\label{Verbaliseren-CPP}Linearisering van de probleempropagatie uit Figuur~\ref{contextprobleempropagatie} waarbij in \textbf{vette tekst} enige al wat woorden zijn toegevoegd om de beweringen aan elkaar te praten.}
\end{figure}

Maar dit is nog steeds nogal houterige taal. Men kan zich voorstellen dat met nog een redactieslag soepel lopende tekst te maken valt uit de lijst van in volgorde geplaatste mededelingen in Figuur~\ref{contextprobleempropagatie}. Het totale proces van verbaliseren stellen we ons voor als drie transformaties. Ten eerste moeten de probleemposities en de oplossingsstappen worden uitgewerkt in tekst. De probleemposities vallen uiteen in reeksen die al zinnen of alinea's suggereren, zoals `Los Angeles City 1943; vieze lucht en mist; brandende ogen'. Ten tweede dient de logica van de netwerkstructuur in aanvullende tekst te worden gerepresenteerd met inleidende en uitleidende zinnen, verwijzingen, kopjes, lay-out, samenvattingen, voorbeelden. Hier zijn dit teksten als `\textbf{De situatie en het probleem betrof}, \textbf{De eerste stappen richting een oplossing waren}.' Ten derde maakt een redactieslag er soepel lopende tekst van. Die laatste stap is hier niet ge\"illustreerd. Het resultaat van dit driestappenproces noemen we een \emph{geverbaliseerde metaforische probleempropagatie}. Zo kunnen we natuurlijk ook spreken van \emph{geverbaliseerde metaforische probleempositie}.

Verbalisering is het resultaat van keuzes: de tekst `1. verhogen motortemperatuur 2. aangepaste motor' had ook na de katalysatorcomplicaties kunnen worden geplaatst. De `denkruimte' wordt kleiner maar de stabiliteit neemt toe. In een natuurkundige metafoor zou je kunnen zeggen dat de metaforische probleempropagatie in superpositie verkeert, in de tekst is een keuze gemaakt.

Dat effect is bij geschreven tekst sterker dan bij mondelinge teksten waarbij nog interactie mogelijk is. Dan kan iemand nog een vraag stellen en de spreker kan reageren met een nuancering en daarna zijn verhaal vervolgen. Kortom, ons ``lichaam kan niet niet-communiceren'' \cite{Watzlawick-nietniet-communiceren}. Maar in ons betoog draait het om geschreven tekst, want dat is de bulkgrondstof voor de training van LLMs.

\subsection{Structuur en inhoud van tekst: statische tekst}\label{structuur-inhoud-tekst}
\noindent Als ``all life probleem solving'' is, dan zijn probleemtypen belangrijke concepten van ons denken \cite[p.~99]{Popper-All-life-problem-solving}. Iedereen kan vaststellen dat het woord probleem in welke conversatie dan ook al gauw valt. We nemen aan dat dit voor teksten ook zo is. Dat is al \'e\'en karakteristiek.

Communicatie kost moeite en moet voordeel opleveren voor de zender. Dat stelt de evolutionair bioloog Arik Kerschenbaum in \emph{Why Animals Talk.\ The New Science of Animal Communication} \cite{Animal-Talk-Kershenbaum}. De tekst moet dus betekenis hebben voor het overleven van de zender. Dit wijst er weer op dat de tekst moet bijdragen aan het oplossen van de actuele problemen van de zender. Dat slaagt het beste als de ontvanger de tekst kan begrijpen. Het is dan ook niet waarschijnlijk dat iemand, mits gevraagd naar zijn probleemposities, werkelijk alles wat los en vast zit erbij sleept of zijn mededeling vormgeeft met een multidimensionale ``stream of consciousness'' zoals James Joyce dat deed in \emph{Finnegans Wake} \cite{Finnegans-wake,litseller-finwake-EN}. Voor een handleiding, een rapport of een bijsluiter werkt dat niet. Als de lezer afhaakt mist de tekst zijn doel. De tekst zal \emph{to the point} moeten zijn.

Zoals gezegd wil degene die de tekst produceert er iets mee bereiken. Het is dan te verwachten dat er enige sturing vanuit gaat en dat het lezen aantrekkelijk is voor de gebruiker. Dit sluit aan bij het onderscheid dat in Sectie~\ref{probpos} gemaakt werd tussen interne en externe probleempositie. De externe probleempositie valt samen met de tekststrategie van een zender.

We kunnen deze `tekststrategie\"en' op basis van het werk van Kahneman c.s.\ nog verder kunnen verhelderen. Zij stellen dat Systeem~1 ons dagelijkse handelen domineert. Het directe, snelle en weinig omslachtige Systeem~1 voldoet in de meeste situaties. Dat lijkt geheel in lijn met ``we simply think and act more or less automatically along certain lines'' uit een eerder citaat van Lakoff en Johnson (Sectie~\ref{ptenprmeta}). Wij vermoeden dat een significant deel van de tekst die mensen produceren ook een Systeem~1-karakter heeft.

Samenvattend, wij menen te mogen hypothetiseren dat de meeste teksten zijn geschreven vanuit een strategisch geori\"enteerde zender, gaan over relevante problemen die deze zender wil adresseren, samenhangend en consistent zijn, naar oplossingen sturen, \emph{to the point} zijn, niet-experimenteel zijn, niet moeilijk te begrijpen zijn, en toegesneden zijn op een toehoorder met een behoefte aan comfort. Bovendien, deze teksten, zoals alle geschreven teksten, zijn gefixeerd, statisch en gesloten. Ze zijn niet te vergelijken met het vloeiende en associatieve proces van het menselijk denken of met mensen die over een bepaald onderwerp discussi\"eren. Dat laatste is ook een vorm van tekst. Toch is het juist het soort `beperkte' type tekst dat wordt gebruikt voor het trainen van LLMs.

Wij stellen dat voorbeelden van deze teksten in tal van categorieën te vinden zijn: krantenartikelen; brochures; handleidingen; berichten op websites; tweets; instructiegidsen; rapporten; Wikipedia-lemma’s; historische verhalen; romans; (populairwetenschappelijke) artikelen. Kortom, heel alledaagse categorieën. Natuurlijk zullen er binnen deze categorieën voorbeelden zijn die minder goed aan onze criteria voldoen, zoals bijvoorbeeld achtergrondartikelen die inhoudelijk diep gaan, of debatverslagen die juist veel interactie registreren. Maar statisch blijven ze toch.

De hierboven geschetste serie van karakteristieken is moeilijk in \'e\'en woord te vatten. Enigszins bruikbare benamingen lijken: professionele tekst, functionele tekst, doelmatige tekst, doelgerichte tekst, of oplossingsgerichte tekst. Maar deze zijn het ook weer net niet en hebben mogelijk al andere betekenissen. Omdat we menen dat het gebrek aan flexibiliteit het belangrijkste is spreken hier voor het geheel van \emph{statische} tekst. Dat is een handige korte term. Maar vergeet dus niet dat onder deze term dus meerdere karakteristieken van tekst schuilgaan.
 
Als we het resultaat van de vorige en deze sectie bundelen, dan menen we dus te mogen concluderen dat de tekst die mensen produceren niet de geestelijke rijkdom representeert waar de tekst het gevolg van is. Ten eerste levert de verbalisering van willekeurig welke metaforische probleempropagatie al een fixatie en vereenvoudiging op. Ten tweede maken mensen niet zo maar tekst. Het gaat in de meeste gevallen om eenvoudige doelmatige tekst. Deze tekst somt een paar punten op, geeft enige overwegingen, trekt een conclusie of noteert een denkstap of een actie en gaat dan door naar het volgende punt dat gemaakt dient te worden.\\

\noindent Wij hebben de indruk dat onderzoekers die tekst-datasets voor LLM-training analyseren, zich vooral richten op inhoudelijke kenmerken van teksten, zoals onaanvaardbare uitspraken \cite{ebac01042-34421883445922,ebac01049-s10462-024-10888-y,ebac01050-naous-etal-2024-beer}. De gedachtenprocessen en verhaallijnen in tekst-datasets krijgen minder aandacht. De auteurs van \emph{Narrative Theory-Driven LLM Methods for Automatic Story Generation and Understanding: A Survey} stellen \cite{ebac01047-liu2026narrativetheorydrivenllmmethods}: ``Overall, [narrative] researchers rarely conduct unsupervised learning and pre-train LLMs on large volumes of texts themselves'' en ``(...) it cannot be ignored that narrative texts make up large portions of LLM pre-training.''  Deze bewering geeft voeding aan onze hierboven geformuleerde hypothese over de structuur en inhoud van de teksten in de tekst-datasets, namelijk dat deze van een specifiek type zijn. 

Ze komen vervolgens met twee zeer relevante vragen: ``What effect does pre-training on more or less narrative texts have on an LLM?'' en ``(...) what kind of narrative artefact is an LLM?'' Deze vragen houden ons ook bezig en we geven er voorlopige antwoorden op. 

Deze vragen zijn het onderwerp van Sectie \ref{denkruimte-chatbots} en Sectie \ref{watvoorp}. De tweede vraag is, denken we, vergelijkbaar met onze vraag, namelijk, `wat voor gespekspartner is een chatbot?' We zullen betogen dat het statische karakter van de tekst-dataset in belangrijke mate de aard van de LLM als ``narrative artefact'' bepaalt. Aangenomen dat we dus ``narrative artefact'' en ``gesprekspartner'' gelijk mogen stellen.

De inzet van de verschillende verhaaltheorie\"en die het survey bespreekt nopen ons om onze argumenten aan te scherpen. Bijvoorbeeld, de kwantificering van narrativiteit, die ``denotes the extent to which a text or discourse exhibits the qualities of narrative, including temporal progression, causal linkage, and the presence of agents and events.'' Deze kwantificering lijkt ons op zichzelf al interessant in relatie tot tekst-datasets \cite{ebac01048-piper-etal-2021-narrative}. Deze metriek kan wellicht ook uitgebreid worden richting een maat van reflexiviteit of andere door ons benoemde kenmerken. 

\section{De denkruimte van chatbots}\label{denkruimte-chatbots}
De ruwe data voor de training van LLMs wordt gevormd door teksten die op een of andere manier on-line voorhanden zijn \cite{commoncrawl-site,ebac01041-s10462-025-11403-7}: websites, brochures, wiki's, romans, geschiedenissen, e-mails, rapporten, handleidingen, verslagen, wetten. Dit materiaal is niet meteen geschikt voor gebruik. Er zijn verschillende redenen om een selectie te maken -- bijvoorbeeld ethische, inhoudelijke of technische gronden -- en methoden, -- bijvoorbeeld automatisch of deels handmatig -- om de teksten geschikt te maken voor trainingsdoeleinden \cite{c4-dataset,ebac01040-3691620.3695061}.  We nemen nu aan de meeste teksten in de tekstdataset voor de training van publiek toegankelijke systemen als ChatGPT, Claude, Grok van het, in de vorige sectie, gepostuleerde statische type zijn. 

Deze statische teksten hebben overeenkomsten qua structuur omdat verbalisering een strategisch selectie en fixatieproces is. Maar er zijn meer overeenkomsten menen wij. Er zijn nog drie redenen die het aannemelijk maken dat teksten in een dataset, hoe groot dan ook, op elkaar lijken: 1.\ De verbalisering cre\"eert verbindingen tussen problemen, probleemposities en oplossingen. De Innovatie-illusie stelt dat het merendeel van de combinaties hiervan bekende werkvormen zijn en dat een beperkt jargon volstaat voor de beschrijving ervan; 2.\ Problemen, probleemposities en oplossingen correleren met conceptuele metaforen uit het conceptuele systeem. Dit systeem is rijk aan patronen omdat conceptuele metaforen op elkaar bouwen; 3.\ Ook de beweringen die niet direct samenhangen met problemen, probleemposities en oplossingen steunen op metaforen, want wij verbaliseren in vergelijkingen. Dat zijn wederom patronen.

Het is bekend dat neurale netwerken goed zijn in het herkennen van patronen. Het is bekend dat LLMs menselijke concepten en features kunnen opslaan. En ook dat LLMs analogie\"en kunnen coderen en benutten. Stel dat die eigenschap niet incidenteel is maar een relatief systematisch effect van het trainingsproces.\\

\noindent Onze hypothese op dit punt is dat het LLM-trainingsproces de metaforische probleempropagaties die in de teksten van een tekst-dataset besloten liggen reconstrueert. Deze resulterende kunstmatige metaforische probleempropagatie, met meer of minder samenhang, vormt tezamen met de woordvoorspelling de basis voor de vaardigheid van een chatbot om een conversatie te voeren die op mensen overtuigend overkomt.\\

\noindent Is er enig bewijs voor deze hypothese? Wij zien verwantschap tussen onze hypothese en de empirische resultaten beschreven in \emph{Evidence of a predictive coding hierarchy in the human brain listening to speech} \cite{ebac01036-caucheteux}. In dit onderzoek luisteren proefpersonen naar verhalen. Ze horen een verhaal terwijl metingen aan hun hersenen volgen wat hun voorspelling is voor het vervolg van het verhaal. Ook worden de verhalen gevoerd aan LLMs. Het blijkt dat: i) de activatiepatronen in de LLMs ``linearly map onto the brain responses to speech''; ii) het LLM werkt op meerdere tijdschalen; iii) voorspelling bij mensen ``are organized hierarchically: frontoparietal cortices predict higher-level, longer-range and more contextual representations than temporal cortices.'' Deze drie resultaten lijken aan te sluiten op de parallel die wij postuleren tussen en metaforische probleempropagatie en de kunstmatige reconstructie en expressie ervan in chatbots; metaforische probleempropagatie kent projectie in meer of minder gedetailleerde tijdruimtes; en, het parallelle conceptualiseren op meerdere niveaus van abstractie.

Een andere aanwijzing vinden wij in \emph{Implicit Representations of Meaning in Neural Language Models} \cite{ebac01043-li-etal-2021-implicit}. De auteurs ``identify contextual word representations that function as models of entities and situations as they evolve throughout a discourse.'' In de woorden van de auteurs, het sleutelverschil met andere studies die proefondervindelijk proberen bloot te leggen wat een LLM codeert, is dat zij ``aim to recover a representation of the situation described by a discourse rather than representations of the sentences that make up the discourse.'' Men slaagt erin, op basis van voorbeelden, aan te tonen dat deze \emph{situations} betekenis dragen want de handelingen die de discourses beschrijven hebben consequenties voor de statistische prominentie van vervolg-\emph{situations}. Wij zien een duidelijke verwantschap tussen onze probleemposities en deze \emph{situations}.\\

\noindent Even los van de vraag of nog grotere datasets een wenselijke ontwikkeling is \cite{ebac01042-34421883445922}. Met de relatief ten opzichte van het verleden enorme actuele tekst-datasets en enorme actuele LLMs slaagt het trainingsproces behoorlijk goed in de reconstructie van metaforische probleempropagaties -- aangenomen dat onze hypothese klopt -- getuige de succesvolle receptie van chatbots door het publiek. Maar de reconstructie is niet perfect, bijvoorbeeld omdat een chatbot kan fabuleren en vaak op eenvoudige vragen geen antwoord heeft. Is hier sprake van kleine foutjes die nog even moeten worden opgelost, of is er sprake van fundamentele beperkingen? De vraag die we in Sectie~\ref{watvoorp} willen beantwoorden is: wat voor metaforische probleempropagatie simuleert een LLM en wat kan de chatbot ermee?

\subsection{Conversaties}
Onze hypothese over conversaties is nu de volgende. Een prompt activeert een spectrum van kunstmatige probleemposities in het LLM. Dat zijn losjes samenhangende groepen van losjes samenhangende clusters van neuronen die concepten coderen. 

De activatie van een kunstmatige probleempositie is meer of minder sterk. De mate van activatie is de statistische mate van herkenning van de prompt als conceptenpatroon. De prompt activeert ook een spectrum van oplossingsstappen (eveneens groepen van clusters van concepten) en probleemposities die daar het gevolg van zijn enzovoort. Deze zijn allen zwakker geactiveerd. 

De door het LLM gegenereerde tokens liggen in de `contexten' van de statistisch meest passende probleemposities en statistisch meest passende oplossingsstappen. Het tekstgeneratieproces heeft direct effect: het spectrum geactiveerde groepen verschuift door de tekstgeneratie, de samenstelling verandert ook en statistische relevantie van de groepen wijzigt. Sommige probleemposities worden prominenter en andere minder relevant. Als de tekstgeneratie stopt is het neurale netwerk in een nieuwe toestand beland. Er zijn `nieuwe' probleemposities die nu de hoofdrol krijgen in het verwerken van de volgende prompt van de gebruiker.

We vermoeden dat de `software schil' rond het LLM niet klakkeloos het LLM volgt, maar dat er wat geprobeerd wordt om te kijken welke tekstgeneratie-opties de `beste' conversatie oplevert. De schil zal keuzes maken op basis van bepaalde parameters, maar hoe precies, dat doet er niet toe hier. Zo kan het zijn dat er geen noemenswaardige activatie van probleemposities is volgens de schil. Dan reageert de chatbot met een of ander ogenschijnlijk passend zinnetje, bijvoorbeeld een vraag om meer uitleg.

Afhankelijk van de uitkomst van het tekstgeneratieproces, al dan niet met bijsturing of selectie door de schil, reageert de chatbot met een wedervraag, een suggestie, een antwoord of een oplossing voor het gestelde probleem. De conversatie ontrolt zich als een pad door de gesimuleerde metaforische probleempropagatie van de chatbot en tegelijk, maar fundamenteel anders, als een pad door de metaforische probleempropagatie van de gebruiker.\\

\noindent Het verschil tussen conversatie en een prompt is naar ons idee niet scherp. Zowel aan de kant van de chatbot als aan de kant van de gebruiker blijven elementen uit de conversatie doorwerken, ofwel, zijn van invloed op het effect van de laatste prompt. Dat moet ook wel, want anders heeft het antwoord van een gebruiker op een vraag van een chatbot geen effect: het spectrum van geactiveerde probleemposities moet veranderen. Op de wisselwerking tussen conversatie en prompt kan ook expliciet gestuurd worden met \emph{in-context-learning} zoals beschreven in de literatuur, dit dient de chatbot cognitief tot grotere hoogte te brengen \cite{ebac00972-3774896}.

Het is verleidelijk de elkaar opvolgende kunstmatige probleemposities via oplossingsstappen voor te stellen als stappen door het neurale netwerk, van de ene groep neuronen, naar een totaal andere groep neuronen. Dat is niet wat er doorgaans gebeurt denken wij. Als een prompt een spectrum van kunstmatige probleemposities activeert zullen die overlappen, en ze zullen ook overlappen met oplossingstappen en daarmee samenhangende probleemposities. Dat is niet verwonderlijk omdat een oplossingsstap kiezen, over het algemeen, de probleempositie maar gedeeltelijk verandert. Dat is zeker het geval bij gecompliceerde problemen. Als iemand een prompt $ABC$ formuleert, en de chatbot suggereert als oplossingsstap $D$, dan is $ABCD$ de nieuwe probleempositie. Het trainingsproces is een vorm van datacompressie en dat maakt dat deze probleemposities overlappend gecodeerd raken.

\section{Wat voor gesprekspartner is de chatbot?}\label{watvoorp}
Stel dat een LLM een metaforische probleempropagatie codeert, zoals boven
betoogd. Wat voor denkhulp is een chatbot en tot welke denkhulp is een chatbot in staat?

\subsection{Wat voor ding is de chatbot?}
We willen ons eerst bevrijden van de gedachte dat een chatbots iets magisch zou zijn dat zich onttrekt aan het menselijk begrip. 

De Cognitieve taalkunde stelt dat betekenis voortkomt uit een voelend en ervarend lichaam dat zich staande probeert te houden in een fysieke omgeving en een sociaal-culturele situatie. Daar zijn we het mee eens. Een chatbot heeft ergens ook een lichaam, maar dat is niet menselijk. Het datacentrum is geen menselijke sociaal-culturele omgeving. De chatbot ontbeert dus de basis van menselijke primaire onderbewuste concepten. Dit is een bekend argument \cite{Mitchell-A-guide-AI-2019}. Dus wat een chatbot dan ook voor aggregatie is, het kan op zijn best menselijke interacties die het vindt in tekst simuleren. De woorden die het produceert hebben niet de betekenis die ze voor ons hebben. 

Dat wil overigens niet zeggen dat een intelligent systeem een biologisch lichaam behoeft of dat het niet een eigen vorm van begrip zou kunnen hebben \cite{ebac01051-Meincke2018}. Maar dat is een andere discussie \cite{ebac01045-rstb20120111}.  In ons betoog gaat het om de te analyseren basale chatbots en in het verlengde daarvan chatbots in het algemeen. Deze verkeren dus in een beperkt-menselijke woordenwereld en hoe die in verband staat met de wereld die wij ervaren ligt buiten de capaciteit van de chatbots.

Is het dan toch nog denkbaar dat een chatbot onze denkwereld volledig kan simuleren, ook al begrijpt de chatbot niet waar we het over hebben?
Deze gedachte lezen wij terug in een hypothese van blogger janus ``a model [as an LLM] whose objective is text prediction will simulate the causal processes underlying the text creation if optimized sufficiently strongly'' \cite{ebac00943-002404.14082v3}. Deze stelling suggereert, dat als je maar verder optimaliseert, zoals grotere datasets, fijnmaziger en ruimer neuraal netwerk, en een intensiever trainingsregime, een model ontstaat dat de causale werking van ons lichaam en brein kan simuleren. Volgens ons is dat ijdele hoop. Zoals aan het slot van Sectie~\ref{structuur-inhoud-tekst} wordt geconcludeerd, gaan er dimensies en associaties verloren bij de overgang van de mentale menselijke denkwereld naar tekst. Dat is een fundamenteel probleem. En praktisch probleem daarbij is dat mensen niet alles wat er in ze omgaat uitdrukken in tekst. De beschikbare tekst waar we het mee moeten doen voor LLM-training is overwegend van het statische type, zo stelden we. Deze tekst geeft met zekerheid een onvolledig beeld van de menselijke cognitieve mogelijkheden. 

De chatbot is dus op twee manieren beperkt. Het is een simulatie want de chatbot heeft geen menselijk ervarend lichaam; de simulatie is gemankeerd want getraind op principieel onvolledige data. Het blijft een artefact dat naar ons idee ver afstaat van alle claims over Artificial General Intelligence of nog sterkere claims \cite{Kurzweil-Singularity-Nearer}. In Sectie~\ref{opvoerenredeneer} betogen we dat het niet aannemelijk is dat `vollediger' data tot aanmerkelijk `betere' resultaten leidt.

\subsection{Statisch `denken', geen reflectie}
Onze hypothese is dat de trainingstekst voor publiekschatbots een statisch karakter heeft (Sectie~\ref{structuur-inhoud-tekst}). Dat heeft effect op de eruit gedestilleerde kunstmatige metaforische probleempropagatie. Het lijkt zeer aannemelijk dat die daarmee ook een statisch karakter heeft. Wij menen daarmee dat een chatbot een vorm van denken simuleert die eerder verwant is aan Systeem~1-denken dan aan analytisch denken, dus weinig kritisch of analytisch \cite{Kahneman-Fast-slow-2012}. Dit denken gaat van problemen en omstandigheden vlot naar conclusies, of oplossingsstappen, of acties, zonder veel reflectie. Daarmee kunnen de omstandigheden, een belangrijk deel van probleemposities, best omvangrijk zijn, maar de beweringen worden niet kritisch behandeld, ze worden aangenomen. De chatbot dwaalt doorgaans niet af van het probleemfront.

In hoeverre dit `denken' menselijk denken volgt is moeilijk precies te maken want Systeem~1-denken is niet exact gedefinieerd, en er is natuurlijk een `grijs gebied' waarin Systeem~1-denken overgaat in meer analytisch denken. Maar gezien het succes van chatbots lijkt het goed aan te sluiten bij de dagelijkse behoeften van gebruikers. Het lijkt aannemelijk conform Kahneman c.s.\ dat als veel mensen zich in veel activiteiten verlaten op Systeem~1-achtig denken, die dagelijkse behoeften ook kenmerkend zijn in conversaties met chatbots. Laten we dat aannemen. Gebruikers zijn dan op zoek naar snelle antwoorden en accepteren die als deze plausibel overkomen. Een chatbot die op alle slakken zout zou leggen of gaat zeuren over vermeende aannamen die wel eens niet waar zouden kunnen zijn, daar zal de doorsneegebruiker snel op uitgekeken zijn vermoeden we. Nee, de chatbot sluit mooi aan op het overheersende Systeem~1-deel van het denkspectrum. De chatbot levert wat veel mensen zoeken in hun dagelijkse conversaties: bevestiging van hun wereldbeeld. Ze gaan lastige vragen uit de weg als ze menen met een korte klap hun doelen te kunnen bereiken. 

Onze conclusie komt niet als een verrassing voor degenen die kritisch het gedrag van chatbots bestuderen: \begin{quote}\cite{ebac00968-2404.01869v2}: ``It is likely that the apparent success of LLMs in reasoning tasks predominantly reflects their ability to memorize the extensive data they have been trained on.''

\cite{ebac00984-Webster31122025}: ``In terms of the Dual Process Theory of human cognition, LLMs are analogous, at best, to system-1 cognitive processing, and it remains unknown how anything approaching system-2 cognition could be incorporated in a machine.'' \end{quote}

\noindent We zijn het natuurlijk eens met deze stellingen. Het verschil met ons betoog en dat van deze auteurs is dat ons model de structuur van ``LLM reasoning'' respectievelijk ``cognitive processing'' nadrukkelijk structuur geeft door de vraag te stellen naar de motiverende basis van kennis en concepten. Onze kijk is gericht op de kenmerken van de teksten waar LLMs op getraind zijn en legt de nadruk op probleem-oplossingrelaties. Die twee aspecten lijken weinig aandacht te krijgen in de literatuur.

\subsection{Toch een denkhulp}\label{tocheendenkhulp}
Is de chatbot beperkt tot in de trainingsdata aanwezige probleemposities of kan de chatbot meer? Het voorbeeld van Geoffrey Hinton uit Sectie~\ref{chatbots} laat zien dat chatbots rake analogie\"en kunnen vinden. Volgens Hinton een analogie die niet in de trainingsdata zat. Wij vermoeden dat dit een wezenlijke eigenschap is. Dus een chatbot zou een denkhulp kunnen zijn, vermoeden we, maar hoe gaat men te werk en tot welk punt reikt de denkhulp? 

Verschillende onderzoeken tonen aan mensen op idee\"en kunnen komen in samenwerking met chatbots. Mensen die voor een creatieve taak staan presteren beter als ze geholpen worden door een chatbot dan als ze er alleen voor staan of als ze alleen Internet-zoeken tot hun beschikking hebben \cite{ebac01079-ChatGPT-on-creativity}. 

Wat echter blijkt uit vervolgonderzoek naar aanleiding van deze resultaten: aggregeert men over de creatieve prestatie van de groep mensen die men onderzoekt, dan neemt de creativiteit af. Kortom de chatbot stuurt naar de bekende weg, de statistisch voor de hand liggende weg \cite{ebac01057-Rebecca-Winthrop-NY-27mei2026, ebac01078-ChatGPT-decreases-idea-diversity-in-brainstorming, ebac01060-1-s2.0-S294988212500091X-main, ebac01061-sciadv.adn5290}. In de woorden van de titel van een van deze artikelen: `Generative AI enhances individual creativity but reduces the collective diversity of novel content.'

We merken dan ook op dat Hinton de chatbot het woord `overeenkomst' in de mond legt. Het zorgvuldig ontwerpen en optimaliseren van prompts is een actueel onderwerp, zie bijvoorbeeld de Google-pagina over prompt engineering \cite{ebac01080-what-is-prompt-engineering}. Er wordt ook gewerkt aan het beter inzichtelijk te maken welke `denkstappen' en chatbot zet om er zo ook meer sturing aan te kunnen geven, bijvoorbeeld \emph{chain of thought prompting} \cite{ebac01038-NEURIPS2022_8bb0d291,ebac01039-9798331314385,ebac00972-3774896}. Onproblematisch lijkt chain of thought prompting overigens niet \cite{ebac01081-kambhampati2026positionstopanthropomorphizingintermediate}. We noemen hier al deze methoden en hulpmiddelen om de chatbot bij de hand te nemen `gericht prompten'. Door gericht prompten kan met de chatbot gezocht worden naar analogie\"en, metaforen of andere relaties tussen concepten. Die concepten of relaties kunnen voor de gebruiker nieuw of bruikbaar zijn. In de literatuur wordt in dit kader gesproken van \emph{dark knowledge} \cite{ebac00969-2407.15017v4}. Zo'n metafoor plaatst een verschijnsel dat normaal gehanteerd wordt in bepaalde termen opeens in andere termen, en dat kan tot nieuwe bruikbare inzichten of doorbraken leiden. Komt de chatbot als denkpartner met zorgvuldig gericht prompten uit de startblokken? 

Je kan een chatbots vragen wat deze van een bepaalde aanname vindt. Van een kritische analytische denkpartner verwacht je echter dat deze z\'elf nieuwe elementen inbrengt, recapituleert -- `wat hebben we nu' of `waar staan we nu' -- of een vastgelopen analyse revitaliseert. Bij dat laatste kan men denken aan een verborgen en onjuiste aanname blootleggen, of een definitie bijstellen, of met een nieuwe hypothese komen, of teruggaan in de redenering en opnieuw beginnen, of alle denkstappen van een dubieus deelbetoog nog eens nagaan. Dit zijn allemaal mentale strategie\"en die de American Psychological Association vat onder critical, creative, lateral, divergent en convergent thinking. Met slechts het conversatieverslag als geheugen mist de chatbot het mentale instrumentarium om dit soort mentale operaties zelf uit te voeren menen wij. Dus, de chatbot moet bij de hand genomen worden om in het analytische deel van het spectrum wat te betekenen. De onderzoekers Taylor Webb, Keith J. Holyoak and Hongjing Lu bevestigen dat beeld als zij stellen \cite{ebac01008-Holyoak02072024}: ``But there is no reason to suppose that the same system, absent human-generated inputs, would spontaneously develop a disposition to think analogically, as apparently happened at some point in human evolution. Thus, to the extent that LLMs capture the analogical abilities of adult human reasoners, their capacity to do so is fundamentally parasitic on natural human intelligence.'' De chatbot is dus geen volwaardige denkpartner.

Dat wil niet zeggen dat een chatbot geen rol kan spelen in een denkproces, zoals hierboven al aangegeven \cite{ebac01079-ChatGPT-on-creativity}. Maar, naar ons idee vereist dan wel dat de resultaten die de chatbot in zijn rol als respondent oplevert betrouwbaar zijn en van systematische uitvoering getuigen.  Helaas schort het daar ook. 

In de inleidende secties noemden we chatbots al onbetrouwbaar. Het is tijd om uit te leggen wat we onder onbetrouwbaarheid verstaan. We gebruiken deze kwalificatie als noemer voor de volgende eigenschappen: 1. onbetrouwbaar in de zin van de gebruiker om te tuin leiden; 2. onbetrouwbaar in de zin dat de chatbot vooroordelen heeft; 3. onbetrouwbaar in de zin van inconsistent, dus eerst A zeggen en later zonder opgaaf van redenen niet-A; 4. onbetrouwbaar in de zin van niet hetzelfde gedrag vertonen in voor de gebruiker identieke of vergelijkbare situaties; 5. onbetrouwbaar in de zin van fabuleren en hallucineren, dingen uit de duim zuigen. 

Het gebrek aan een lichaam verklaart naar ons idee voor een belangrijk deel de aspecten van onbetrouwbaarheid 3-5, en draagt bij aan 1-2. Problemen van type 4 menen wij terug te zien in werk aan analogiepuzzels en wiskundige problemen \cite{ebac01007-lewis2024evaluatingrobustnessanalogicalreasoning, ebac01009-opielka2025analogicalreasoninginsidelarge, ebac01011-glazer2025frontiermathbenchmarkevaluatingadvanced}. 
Echter 1-2 zijn een effect van de trainingsdata, mensen zijn ook niet altijd even betrouwbaar en veel tekst is meer of minder strategisch, en in sommige gevallen zelfs manipulatief. De concepten die de chatbot hanteert missen een fundament, de chatbot kan niet toetsen of een zin ergens op slaat, of deze te ver gaat in de manipulatie van de gebruiker, de chatbot mist \emph{common sense}. 

De aspecten 3-5 hebben mogelijk voordelen in het analytische denkspectrum. Namelijk om buiten de gebaande patronen mee te denken. Maar de gebruiker moet dan wel heel alert zijn. Deze eigenschappen vormen echter een risico voor gebruikers die in het Systeem~1-deel van het denkspectrum opereren menen we. We komen daar op terug in  Sectie~\ref{gevaarlijkemix}. 

\subsection{Opvoeren van redeneervermogen?}\label{opvoerenredeneer}
Zou het reflectieve vermogen versterkt kunnen worden met een `analytische trainingsset'. Wat zou nu typische analytische tekst zijn? Wij denken aan diepgravend filosofisch onderzoek over een onderwerp dat nog verre van doorsneewetenschap is,  dat systematisch allerlei redeneringen afpelt en weegt en dat tot slot geen harde conclusie trekt, alleen conditionele. Een voorbeeld zou Jan Verplaetses overzicht en kritiek van de vrije wil-discussie kunnen zijn \cite{JanVerplaetse-zvw}. Het zal niet eenvoudig zijn deze analytische trainingsset samen te stellen, denken wij, maar dat terzijde. 

Maar toch, stel dat een analytische trainingsset samengesteld kon worden, wij denken niet dat een chatbot zich daarmee tot analytische denkpartner kan verheffen. Wij vermoeden dat als de gesimuleerde metaforische probleempropagatie gedeeltelijk aspecten van zijn statische karakter verliest door zekerheden in de beweringen te vervangen door reflectieve terugkoppelingen, chatbots sterker zouden kunnen gaan fabuleren. Je kan er dan wellicht meer mee doen als denkpartner, maar de noodzaak tot alerte begeleiding van de chatbot wordt ook groter.

Analytische tekst lost het gebrek aan een in ervaring gegrond conceptueel systeem aan de kant van de chatbot niet op. Zelfs tussen mensen schiet tekst vaak te kort bij het overbrengen van betekenis. Maar wij beschikken wel over een in ervaring gegrond conceptueel systeem. Dat helpt ons om te reconstrueren wat een ander bedoelt. Soms lukt dat alleen door herhaald lezen en teruglezen (\emph{close reading}) om daarmee zelf een mentale structuur te construeren die consistent is met wat je leest. Dan kan je er nog steeds naast zitten, je begrijpt de tekst dan niet echt. Lakoff en Johnson stellen, zoals eerder geciteerd, \cite[p.~3]{Lakoff-Johnson-By-Metaphores}:
``language is an important source of evidence for what [the conceptual system] is like.'' Tekst is daarbij een overvloedig beschikbaar medium. Maar het is wel een problematisch medium.

Onze conclusie is dus dat een chatbot als conversatiepartner niet een analytische denkpartner is en dat deze dat ook niet kan worden met zijn actuele architectuur en door training op tekst alleen. De steeds grotere modellen zullen chatbots-denkpartners niet analytischer maken, bovendien zijn er ook veel andere bezwaren om die kant op te werken \cite{ebac01042-34421883445922}. Yann LeCun, vooraanstaand AI-onderzoeker en Meta’s voormalige\emph{ Chief AI Scientist}, stelt: ``Animals and humans exhibit learning abilities and understandings of the world that are far beyond the capabilities of current AI and machine learning (ML) systems.'' En hij denkt niet dat de actuele AI techniek in staat is dit verschil te overbruggen \cite{ebac01004-LeCun-path_towards_autonomous_mach}. Daarmee zijn we het mee eens. 

\subsection{Concept-based interpretability}\label{gevaarlijkemix}
De metaforische probleempropagatie verbindt menselijk denken, menselijke tekst en de werking van LLMs. Onderzoek in de Interpretability laat zien dat men erin slaagt concepten en eigenschappen te lokaliseren. Als ons model enig nut heeft dan vermoeden wij voor 
Concept-based interpretability (Sectie~\ref{interp}). Ons model suggereert immers een meta-structuur over concepten. We zeggen er meteen bij, we staan te ver buiten dit zeer technische onderzoeksgebied om de waarde van ons model te beoordelen.  

Hoe dan ook, als Concept-based interpretability erin slaagt om de conceptstructuren van LLMs te ontrafelen, ontstaat daarmee misschien de mogelijkheid om bepaalde eigenschappen te sturen, zoals de verschillende aspecten van betrouwbaarheid. Dat klinkt aantrekkelijk want de mix van onbetrouwbaarheid en robuustheid -- de chatbot heeft altijd een vlot lopend antwoord -- is bewezen gevaarlijk voor gebruikers die chatbots te lichtzinnig inzetten of te goed van vertrouwen zijn \cite{ebac00914-nrc-krant-20250920-5725713-je-gevoelens-zijn-echt-chatgpt-adam-zelfmoordplan-EN}. Gloort met Concept-based interpretability de hoop dat het risico op ongelukken valt in te perken? 

Hierboven hebben wij betoogd dat het redeneervermogen niet verbeterd kan worden en dat de robuustheid inherent is. Dan blijft alleen betrouwbaarheid verhogen over om het risico op ongelukken te beperken. We zien een aantal werkwijzen om dit te bereiken: vangrails, correctheidsbewijzen en herontwerp. 

Onder vangrails verstaan we hier software rond een LLM die meekijkt in de conversatie en die ingrijpt als het misgaat. Het is vergelijkbaar met een automatische brandblusser in een keuken met een gasfornuis. Correctheidsbewijzen zijn wiskundig precieze methoden om eigenschappen van software aan te tonen. Herontwerp is een LLM zo inrichten dat het zich altijd goed gedraagt. 

Wij menen dat al deze manieren kampen met een fundamenteel probleem: het is niet goed te specificeren welk gedrag acceptabel is en wat niet. We begrijpen onszelf immers maar zeer ten dele. Vooroordelen en neigingen, \emph{bias}, vormen daarbinnen een apart probleem \cite{ebac01044-TheNooscopeManifested,ebac01042-34421883445922}. Stel dat men er toch in zou slagen een `acceptabel' pakket van gewenste eigenschappen te specificeren, dan is onze verwachting dat het aantonen van die eigenschappen in een implementatie heel lastig zal zijn \cite{Modelling-Analysis-GrooteMousavi}. Twee grote obstakels zijn er dus.

Als men toch beide obstakels voldoende overtuigend overwint, dus als men er redelijk in slaagt het gewenste gedrag te specificeren en aan te tonen dat een systeem zich ernaar gedraagt, dan zal vanuit het voorzorgprincipe de neiging sterk zijn, vermoeden we, om de betrouwbaarheid te maximeren \cite{RoelPieterman2008-EN}. Als die keuze gemaakt wordt, zal dat ten koste gaan van de speelsheid en het associatievermogen van LLMs, vermoeden wij. En wij denken dat de bruikbaarheid van de op deze LLMs gebaseerde chatbots dan sterk afneemt. Wij zijn het niet eens met een bewering uit de literatuur als zouden \emph{hallucination} en \emph{knowlegde conflict} uitdagingen zijn die ``mainly derive
from improper learning data'' \cite{ebac00969-2407.15017v4}.

Het lijkt erop dat men zich voor chatbots, willen ze enig nut hebben als creatieve denkhulp, tevreden moet stellen met een niet-precies gedefinieerde en domein-specifieke ``comfort zone'' \cite{Kleinberg-AI-comfortzone}. En de gebruiker moet er z\'elf alert voor zorgen om in die zone te blijven en zich niet laten meeslepen in illusie dat de chatbot een alwetende innovatie is. Het geeft enige hoop dat de onderzoekers Eric Bigelow en Tomer Ullman in \emph{People Evaluate Agents Based on the Algorithms That Drive Their Behavior} dat mensen, als ze de kans krijgen, zich rekenschap geven van hoe een systeem werkt alvorens de resultaten ervan te beoordelen \cite{ebac00974-opmia26}.

\section{Conclusie}\label{conclusie}
Chatbots zijn computerprogramma's die een gesprekpartner of vraagbaak voor een mens vormen in een geschreven of gesproken conversatie. Hierbij kunnen we denken aan ChatGPT, Grok, Claude, Deepseek, Gemini, Lumo. Het doel van dit artikel is een perspectief te bieden op de aard van chatbots als werkelijke gesprekspartner over problemen in relatie tot hun oplossingen. Wat kunnen chatbots in dat geval wel en wat niet, en hoe valt dat te verklaren?\\ 

\noindent Ons betoog richt zich op basale chatbots in de hoop daarmee uitspraken te doen over de kernfunctionaliteit van chatbots in algemene zin.
Basale chatbots worden verondersteld te bestaan uit een Large Language
Model (LLM) met een eenvoudig interface.

Om het gedrag van deze chatbots in probleem-oplossing gedreven conversaties beter te begrijpen, beschouwen wij de samenhang tussen: het menselijk denken, de menselijke tekstproductie, de tekst-dataset voor LLM-training, de werking van LLMs. Wij introduceren de notie \emph{metaforische probleempropagatie} om over deze samenhang te reflecteren. 

Een metaforische probleempropagatie is een model van de menselijke denkruimte, of een deel ervan dat samenhangt met een bepaalde kwestie, die bestaat uit de combinatie van de probleembegrippen van de Aggregatie-dynamica, het conceptuele systeem van de Cognitieve taalkunde en de voorspellingsdynamiek uit de Predictive processing theory.

Mensen produceren teksten. We vragen ons af wat de eigenschappen van die teksten zijn.  Wij beweren dat de teksten die gebruikt worden om chatbots te trainen slechts gedeeltelijk het menselijk denken representeren. We beweren dat dit komt omdat deze teksten doelgericht zijn, voor specifieke gebruikers zijn en gesloten zijn. We noemen dit \emph{statische} tekst. Deze statische teksten kennen bovendien overeenkomsten omdat mensen in herhaling vallen bij het identificeren en oplossen van problemen, ze zijn minder innovatief dan ze denken (de Innovatie-illusie), en wegens de centrale rol van, onderling sterk verweven, metaforen in de menselijke conceptualisering, het onder woorden brengen van gedachten en de compositie van teksten.
 
Het is bekend dat neurale netwerken patronen kunnen vinden. Onze hypothese is dat het leerproces van een LLM uit de aangeboden teksten de gereduceerde en statische, metaforische probleempropagaties die daarin besloten liggen reconstrueert. Een prompt maakt meerdere probleemposities en oplossingsstappen statistisch meer of minder plausibel. Of het LLM \'e\'en van de oplossingstappen uitwerkt (wat ook een vraag om meer informatie kan zijn) of dat de chatbot ingrijpt met een reactie dat doet er niet zoveel toe. De reactie verandert het landschap van geactiveerde probleemposities en het vervolg erop. In die toestand verandert de volgende prompt van de gebruiker het landschap weer. Een conversatie kan nu gedefinieerd worden als een pad langs de probleemposities in de kunstmatige metaforische probleempropagaties van het LLM en een pad door de metaforische probleempropagaties van de gebruiker.

Een chatbot kan meedenken volgens bekende schema's, feiten opzoeken en tekst genereren volgens bekende redeneerlijnen. Dat staat ver af van analytisch denken. Een chatbot kan desalniettemin verleid worden tot creatieve `denkstappen'. Maar daarvoor moet de chatbot bij de hand genomen worden. Mits goed aangestuurd, kan het analogie\"en vinden in de enorme databank die het tot zijn beschikking heeft, en in zijn ongeremdheid kan het zo relaties leggen tussen systemen, situaties of domeinen die nieuw zijn en wellicht zinvol. We betogen dat om verschillende redenen de chatbottechniek niet kan worden uitgebouwd tot een betrouwbare, zelfstandige, analytische denkpartner.\\

\noindent Onze gevolgtrekkingen sluiten aan bij de scepsis van Yann LeCun en staan haaks op het optimisme van Big Tech. Dat neemt niet weg dat Chatbots er zijn, dat ze
massaal worden toegepast, zowel door individu\"en als organisaties, en dat
het daarmee maatschappelijk en politiek belangrijk is om ze te begrijpen. Een kernvraag voor de discussie is daarbij: voor welk probleem is een chatbot een oplossing? Ons artikel wil een bijdrage leveren aan het antwoord.

\subsection*{Dankwoord}
Dank aan Atze Bosma en Jos van Wamel voor commentaar op eerdere versies van dit artikel.